\documentclass[11pt]{article}

\usepackage{hyperref}       
\usepackage{url}               
\usepackage{nicefrac}       
\usepackage{soul}
\usepackage{color}  
\usepackage{float}
\usepackage{microtype}
\usepackage{graphicx}
\usepackage{booktabs}
\usepackage{epsfig,subfigure}
\usepackage{caption}
\usepackage{algorithm}
\usepackage{algorithmic}
\usepackage{xcolor,cancel}  
\usepackage{amsmath,amsfonts,amssymb}
\usepackage{verbatim}
\usepackage{stfloats}
\newcommand{\define}{\stackrel{\triangle}{=}}

\usepackage[bookmarks=false]{}
\newtheorem{theorem}{Theorem}

\newtheorem{definition}{Definition}
\newtheorem{lemma}{Lemma}

\newtheorem{remark}{Remark}
\newtheorem{observation}{Observation}
\newtheorem{example}{Example}
\newtheorem{experiment}{Experiment}

\begin{document}
\date{}
\title{{\color{black} ForestDSH}: A Universal Hash Design for Discrete Probability Distributions}


\author{ \small Arash Gholami Davoodi, Sean Chang, Hyun Gon Yoo, Anubhav Baweja, Mihir Mongia and  Hosein Mohimani \\
{\small Carnegie Mellon University, Pittsburgh, PA, 15213}\\
{\small \it Email: \{agholami,szchang,hyungony,abaweja,mmongia,hoseinm\}@andrew.cmu.edu}}
\maketitle

\begin{abstract}
In this paper, we consider the problem of classification of $M$ high dimensional queries $y^1,\cdots,y^M\in \mathcal{B}^S$ to  $N$ high dimensional classes  $x^1,\cdots,x^N\in \mathcal{A}^S$ where $\mathcal{A}$ and $\mathcal{B}$ are discrete alphabets and the probabilistic model that relates data to the classes $\mathbb{P}(x,y)$ is known. This problem has applications in various fields including the database search problem in mass spectrometry. The problem is analogous  to the  nearest neighbor search problem, where the goal is to find the data point in a database that is the most similar to a query point. The state of the art method for solving an approximate version of the nearest neighbor search problem in high dimensions is locality sensitive hashing (LSH). LSH is based on designing hash functions that map near points to the same buckets with a probability higher than random (far) points. To solve our high dimensional classification problem, we introduce distribution sensitive hashes that map jointly generated pairs $(x,y)\sim \mathbb{P}$  to the same bucket with probability higher than random pairs $x\sim {\mathbb{P}}^{\mathcal{A}}$ and $y\sim {\mathbb{P}}^{\mathcal{B}}$, where  $ {\mathbb{P}}^{\mathcal{A}}$ and ${\mathbb{P}}^{\mathcal{B}}$ are the marginal probability distributions of $\mathbb{P}$. We design distribution sensitive hashes using a forest of decision trees and we show that the complexity of search grows with $O(N^{\lambda^*({\mathbb{P}})})$ where $\lambda^*({\mathbb{P}})$ is expressed in an analytical form. {\color{black}We further show that the proposed hashes perform faster than state of the art approximate nearest neighbor search methods for a range of probability distributions, in both theory and simulations.} Finally, we apply our method to the spectral library search problem in mass spectrometry, and show that it is an order of magnitude faster than the state of the art methods. 
\end{abstract}

\section{Introduction}
Consider the problem of classifying  a large number of high dimensional data $Y=\{y^1,\cdots,y^M\}\subset\mathcal{B}^S$ into high dimensional classes $X=\{x^1,\cdots,x^N\}\subset\mathcal{A}^S$, given a known joint probability distribution $\mathbb{P}(x,y)$, where $\mathcal{A}$ and $\mathcal{B}$ are discrete alphabets. Given a point $y\in Y$, the goal is to find the class $x\in X$ that maximize $\mathbb{P}(y\mid x)$;
\begin{eqnarray}
{Arg}\max_{x\in{X}}\mathbb{P}(y\mid x), \label{pyx}
\end{eqnarray}
where $\mathbb{P}(y\mid x)$ is factorizable to i.i.d. components, i.e.,\footnote{Here, we assume that $\mathbb{P}(x)$ is also factorizable to i.i.d. components.}
\begin{eqnarray}
\mathbb{P}(y\mid x)=\prod_{s=1}^S  p(y_s\mid x_s).\label{pyx_2}
\end{eqnarray}
{\color{black}In this paper, we refer to $X=\{x^1,\cdots,x^N\}$ as data base points and  $Y=\{y^1,\cdots,y^M\}$ as queries.} This problem has application in various fields, including the clustering of spectra generated by mass spectrometry instruments \cite{frank2011spectral}. Consider the problem where there are billions of data points (mass spectra) and given a query spectrum $y$ the goal is to find the spectrum $x$ that maximize known probability distribution $\mathbb{P}(y\mid x)$ {\cite{Aebersold,MSGF}}. {\color{black}This is similar to the nearest neighbor search problem \cite{dasarathy1977visiting,yianilos1993data}, where instead of minimizing a distance, we maximize a probability distribution, and has applications in machine learning \cite{duda1973pattern}, database querying \cite{guttman1984r} and computer vision \cite{mori2001shape,shakhnarovich2003fast} }.

A naive approach to solve this problem is to compute $\mathbb{P}(y\mid x)$ for each $x\in X$, and find the maximum. The run time for this algorithm is $O(NS)$ which is very slow when the number of classes, $N$, is large. 

In order to address this problem where the number of classes is massive, multiclass classification methods have been established \cite{Bentley,prab,Finkel,jain2016extreme,bhatia2015sparse,yen2016pd,choromanska2015logarithmic,liu2017making,nam2017maximizing,rai2015large,tagami2017annexml,niculescu2017label,zhou2017binary}. 
{\color{black}For example, in \cite{choromanska2015logarithmic} an approach to construct a tree with logarithmic (in terms of number of classes) depth is presented and logarithmic train and test time complexity is achieved. However, these methods are limited to low-dimensional data {\color{black}(when the number of dimensions is less than $10$)}, and their complexity grows exponentially with the dimension. }{\color{black}The methods in \cite{Bentley,prab,Finkel} speed up the prediction by rejecting a significant portion of classes for each query. However, all these methods suffer from the curse of dimensionality, i.e., when the dimension of the data increases, the complexity of these methods increases linearly with the number of classes.\footnote{\color{black}The curse of dimensionality holds for non-deterministic probability distributions. When $p(y\mid x)$ is deterministic, i.e., when it takes only a single value with probability one, there is no curse of dimensionality. In this paper, we are interested in the case of non-deterministic probability distributions.} } {\color{black}Dimension reduction can also be utilized before applying nearest neighbor search algorithms to  high dimensional data   \cite{beyer1999nearest,castelli2000searching,anagnostopoulos2015low,min2005non}.  
Moreover, Structure Preserving Embedding (SPE) which is a low-dimensional embedding algorithm for Euclidean space is also used  as a dimension reduction tool \cite{shaw2009structure}.} {\color{black} Set similarity in nearest neighbor search literature is studied extensively  \cite{christiani2017set,charikar2002similarity}. In \cite{christiani2017set}, the authors develop a data structure that solves the problem of approximate set similarity search under Braun-Blanquet similarity $B(x,y)=\frac{|x\cap y|}{\max(|x|,|y|)}$   in sub-quadratic query time.} 

A similar problem has been investigated in the field of nearest neighbor search. In this problem, given a set of points in a database, the goal is to find the point in the database that is closest to a query point. A popular approach to this problem is locality sensitive hashing \cite{Motwani,Indyk_hd}. This approach solves $\epsilon$-approximate nearest neighbor search problem by designing a family of hashes in a way that near points are hashed to the same bucket with probability much higher than random points.  In $\epsilon$-approximate nearest neighbor search problem, given a query point $y$, the goal is to find $x\in X$ for which $d(x,y)\le (1+\epsilon)d(x',y)$ for all $x'\in X$ and $X$ is the set of all feasible points \cite{Motwani,bawa2005lsh}.    One of the most
popular methods for approximate nearest neighbor search is LSH \cite{Motwani,Indyk_hd,andoni2017optimal,rubinstein2018hardness,shrivastava2014asymmetric,christiani2017set}.  For any metric space $\mathcal{M}=(M,d)$, a family of functions  $h:{\mathcal M}\to S$ is $\mathcal{F}(R,cR,{p}_1,{p}_2)$-sensitive if for  any two points $x,y \in {\mathcal M}$: 
\begin{eqnarray}
&&\mbox{If~} d(x,y) \le R, \mbox{~then~} h^{\mathcal{A}}(x)=h^{\mathcal{B}}(y)\mbox{~with probability at least~} {p}_{1} .\label{aaa}\\
&&\mbox{If~} d(x,y) \ge cR, \mbox{~then~} {h^{\mathcal{A}}(x)}=h^{\mathcal{B}}(y)\mbox{~with probability at most~} {p}_{2} .\label{bbb}
\end{eqnarray}
{\color{black}A family is interesting when $p_1>p_2$.  In the case of hamming distance, i.e., when data are in the form of $d$-dimensional vectors from ${\{0,1\}}^d$, the family of hashes may be defined simply as $H=\cup_{i=1}^d\{h:\{0,1\}^{d}\to \{0,1\}\mid h(x)=x_{i}\}$ where $x_i$ is $i$-th coordinate of $x$. From \eqref{aaa} and \eqref{bbb}, we conclude that $p_1=1-\frac{R}{d}$ and $p_2=1-\frac{cR}{d}$\footnote{Note that $d(x,y)\le R$ is equivalent to $x$ and $y$ are different in at most $R$ coordinates.}.}

In this paper, in addition to going from minimizing a distance to maximizing a probabilistic measure, our approach differs from the classic LSH in the following way. The proposed hashes in this paper are defined to be a subset of integers instead of an integer and our goal is to ensure that\footnote{In fact, to control the complexity, two more conditions are defined in Definition \ref{def_family_ds}. {${\mathbb{P}(x,y)}$ is joint probability distribution while  ${\mathbb{P}^{\mathcal{A}}(x)}$ and ${\mathbb{P}^{\mathcal{B}}(y)}$ are marginal probability distributions of ${\mathbb{P}(x,y)}$. ${\mathbb{Q}(x,y)}$ is also defined as  ${\mathbb{Q}(x,y)}={\mathbb{P}^{\mathcal{A}}(x)}{\mathbb{P}^{\mathcal{B}}(y)}$.}} 
\begin{eqnarray}
Prob(H^{\mathcal{A}}(x)\cap H^{\mathcal{B}}(y)\neq \varnothing\mid (x,y)\sim \mathbb{P})&\ge& \alpha,\\
Prob(H^{\mathcal{A}}(x)\cap H^{\mathcal{B}}(y)\neq \varnothing\mid (x,y)\sim \mathbb{Q})&\le& \beta.
\end{eqnarray}
In words,  these hashes hash the jointly-generated pairs of points to the same buckets with probabilities higher than $\alpha$ while hash random pairs of points to the same buckets with probabilities lower than $\beta$.  Note that, in this case for data points $x$ and $y$, collision happens when we have  $H^{\mathcal{A}}(x)\cap H^{\mathcal{B}}(y)\neq \varnothing$, while in the classic LSH, $x$ and $y$ have collision if $H^{\mathcal{A}}(x)= H^{\mathcal{B}}(y)$. The idea of defining hash collision as the intersection of hash sets has been previously proposed in \cite{christiani2017set}. 

Currently, the locality sensitive hashing approach does not generalize to the cases where the triangle inequality, i.e., $d(x,y)\le d(x,z)+d(z,y)$ does not hold  \cite{charikar2002similarity,Motwani,Indyk_hd,ACM_andoni,nips_andoni,Chakrabarti,Miltersen,noar}. 
 {\color{black}These papers are based on balls on some specific points in the metric space and the notion of balls is well defined only for metric spaces that satisfy triangle inequality.} Recently, high dimensional approximate nearest neighbors in a known probabilistic distribution setting have been investigated \cite{bawa2005lsh,Moshe_Heterogeneous}. However, currently it is not possible to design efficient algorithms based on these methods, due to the large number of parameters involved. 

{\color{black}
The problem of finding high dimensional approximate nearest neighbors  in a known probabilistic setting using a bucketing tree algorithm has been studied previously in \cite{Moshe_Heterogeneous,Moshe_Bucketing}. Dubiner uses a   strategy to hash the data points from an arbitrary joint probability distribution into the leafs of the tree in a way that the paired data collide with a probability higher than the random pairs. {\color{black}Here, paired data refers to pairs coming from joint probability distribution, i.e., $(x,y)\sim \mathbb{P}$  random pairs are the pairs coming from an independent probability distribution, i.e., $x\sim {\mathbb{P}}^{\mathcal{A}}$ and $y\sim {\mathbb{P}}^{\mathcal{B}}$.} However, the algorithm introduced in Dubiner requires solving computationally intractable optimizations, making it impossible to implement (e.g. see equation (126) from \cite{Moshe_Heterogeneous}). However, in this paper, for  the specific case where the distribution {\color{black}for any $s\in\{1,2,\cdots,S\}$ is $
p(x_s=0,y_s=0)=p(x_s=1,y_s=1)=\frac{p}{2},p(x_s=1,y_s=0)=p(x_s=1,y_s=0)=\frac{1-p}{2}$,} our theoretical and practical results are compared to   \cite{Moshe_Heterogeneous} for the range of $0\le p\le 1$.}

In this paper, we propose to solve \eqref{pyx} by defining a family of distribution sensitive hashes satisfying the following property. They hash the jointly-generated pairs of points to the same buckets with probabilities much higher than random pairs. {\color{black}We further design an algorithm to solve \eqref{pyx} in sub-linear time using these families of hashes. Next, a method to find optimal family of hashes is presented to achieve minimum search complexity using multiple decision trees. Note that, these decision trees have the same tree structure while they apply to different permutations $\{1,2,\cdots,S\}\rightarrow\{1,2,\cdots,S\}$ of the data.} This way, we design forest of decision trees where each decision tree captures a very small ratio $\alpha$ of  true pairs for some $\alpha\in \mathbb{R}^+$ and by recruiting $\#bands=O(\frac{1}{\alpha})$ independently permuted decision trees we can reach near perfect recovery of all the true pairs. In this paper, we refer to each decision tree as a band and $\#bands$ is referred to as the number of bands.

{\color{black}The main idea is that we construct decision-tree hashes, in a way that the chance of true pairs hashing to the same leaf nodes in these decision trees is higher than random points (Figure $1$).}  The decision  tree is built in a way that the ratio $\frac{\mathbb{P}(x,y)}{\mathbb{Q}(x,y)}$ is higher than a minimum threshold, while the ratios $\frac{\mathbb{P}(x,y)}{\mathbb{P}^{\mathcal{A}}(x)}$ and   $\frac{\mathbb{P}(x,y)}{\mathbb{P}^{\mathcal{B}}(y)}$ and the number of nodes in the graph are lower than a maximum thresholds, {\color{black}  see Algorithm \ref{algorithm_Tree}. We further determined the optimal tree among many trees that can be constructed in this way. Two theorems are presented here on the complexity of the decision tree built in Algorithm \ref{algorithm_Tree}.
\begin{enumerate}
\item No decision tree exists with the overall search complexity below $O(N^{\lambda^*})$ where $\lambda^*$ is derived analytically from the probability distribution $\mathbb{P}(x,y)$.
\item The decision  tree construction of Algorithm \ref{algorithm_Tree} described in \eqref{ok} results in the overall search with complexity $O(N^{\lambda^*})$.
\end{enumerate}
}

 Our results show that our approach, Forest-wise distribution sensitive hashing ({\color{black}ForestDSH}), provides a universal hash design for arbitrary discrete joint probability distributions, outperforming {\color{black}the existing state of the art approaches in specific range of distributions, in theory and practice.}  Moreover, we applied this method to the problem of clustering spectra generated from mass spectrometry instruments. 

An alternative strategy for solving \eqref{pyx_2} is to reformulate the problem as minimum inner product search problem (MIPS) \cite{shrivastava2014asymmetric} by transferring the data points into a new space. However, as we show in Section 6 the transferred data points are nearly orthogonal to each other making it very slow to find maximum inner product using the existing method \cite{shrivastava2014asymmetric}.        

{\color{black}
Note that, the algorithms presented in this paper are based on the assumption that the true pairs are generated from a known distribution $\mathbb{P}$. The distribution $\mathbb{P}$ can be learned from a training dataset of true pairs.  In practice, training data can be collected by running a brute force search on smaller data sets or portion of the whole data. For example, in case of mass spectrometry search, we collect training data by running brute-force search on small portion of our data \cite{frank2011spectral}.} Note that, we can never perfectly learn the probability distribution that the data is generated from. In Theorem 3, we prove that our algorithm is robust to noise and our experiment confirms this.

{\it Notation:}  The cardinality of a set $A$ is denoted as $|A|$. The sets $\mathbb{N}$ and $\mathbb{R}$  stand for the sets of natural and real numbers, respectively. We use $\mathbb{P}(\cdot)$ and {$\mathbb{Q}(\cdot)$} to denote the probability function $\mbox{Prob}(\cdot)$. $\mathbb{E}(\nu)$ denotes the expected value of the random variable $v$.  Moreover, we use the notation $f(x)=O(g(x))$, if $\limsup_{x\rightarrow\infty}\frac{|f(x)|}{g(x)}<\infty$ and the notation $f(x)=\Omega(g(x))$, if $\limsup_{x\rightarrow\infty}\frac{|f(x)|}{g(x)}>0$. In this paper, $\log x$ is computed to the base $e$.
\section{Definitions}
\begin{figure}[h]
	\centerline{\includegraphics[width=5.33in]{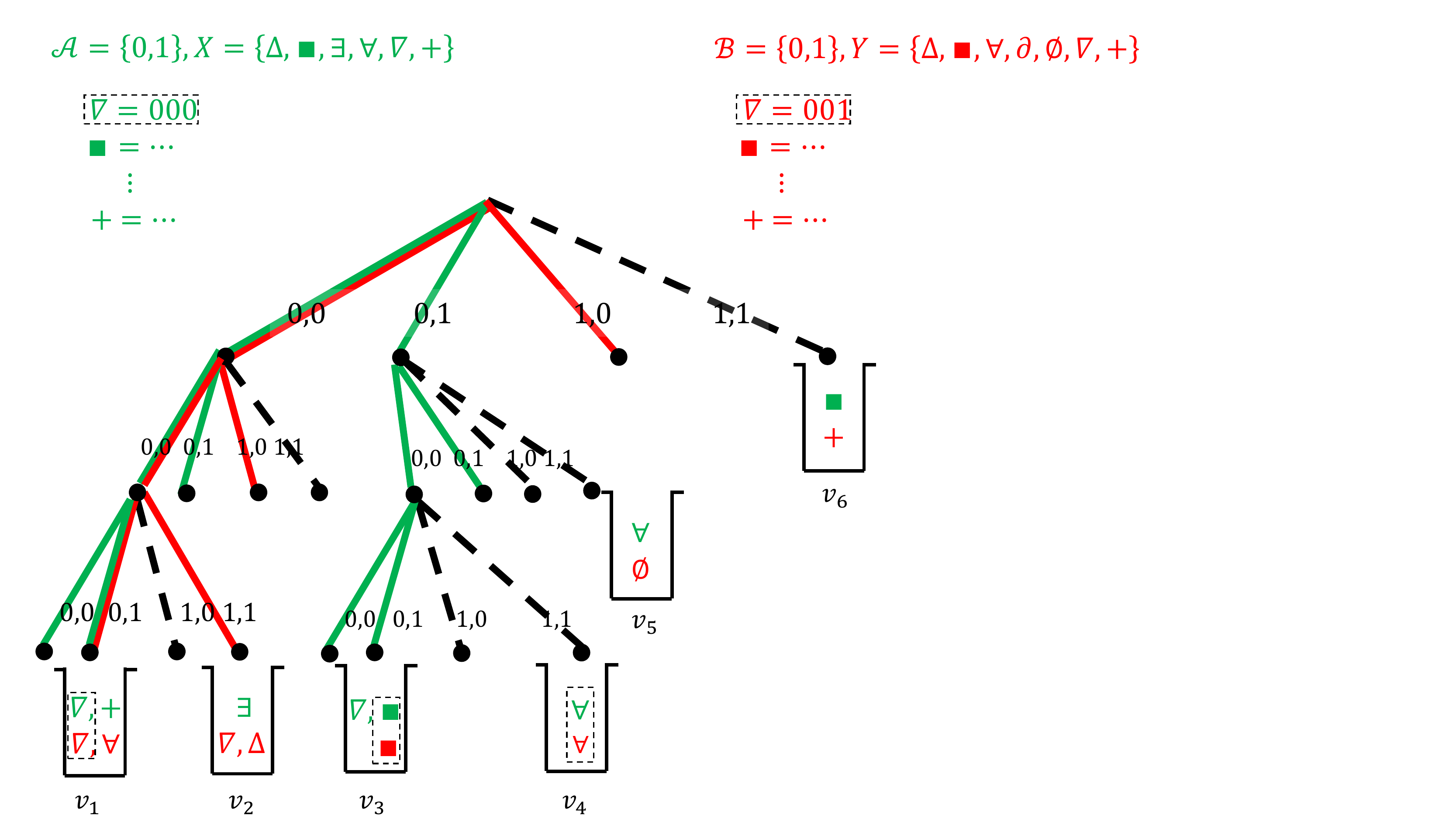}}
\caption{ Distribution sensitive hashing is based on a decision tree $G$ and a set of buckets $V_{Buckets}$ which are a subset of leaf nodes of $G$. Construction of the tree and $V_{Buckets}$ are explained in Algorithm \ref{algorithm_Tree}. Here, $V_{Buckets}$ has six buckets in it. We map data points to buckets by propagating them down the tree, starting from the root using Algorithm \ref{haha}. If we reach a bucket node, we insert the data point. On the other hand, {\color{black}only a subset of leaf nodes are labeled as bucket} and if we reach a leaf node that is not a bucket node, we will simply ignore it. The mapping of {\color{green}$\triangledown=000$} and {\color{red}$\triangledown=001$} are shown in the figure {\color{black}for $S=3$}. Each of the data points {\color{red}$\triangledown=001$} and {\color{green}$\triangledown=000$} are mapped to two buckets. {\color{black}Consider data point {\color{green}$\triangledown=000$}. Mapping this point by propagating down the tree, we end up with the two buckets $v_1$ and $v_3$. Note that, the first character on each of the edges in the paths from root to $v_1$ and $v_3$ is $0$. Similarly mapping  {\color{red}$\triangledown=001$} by propagating down the tree, we end up with the two buckets $v_1$ and $v_2$ as the second alphabet on the edges should be $0$ in all the first two depths and $1$ at the third depth.  }}\label{fig1}
\end{figure}

\begin{figure}[h]
	\centerline{\includegraphics[width=5.33in]{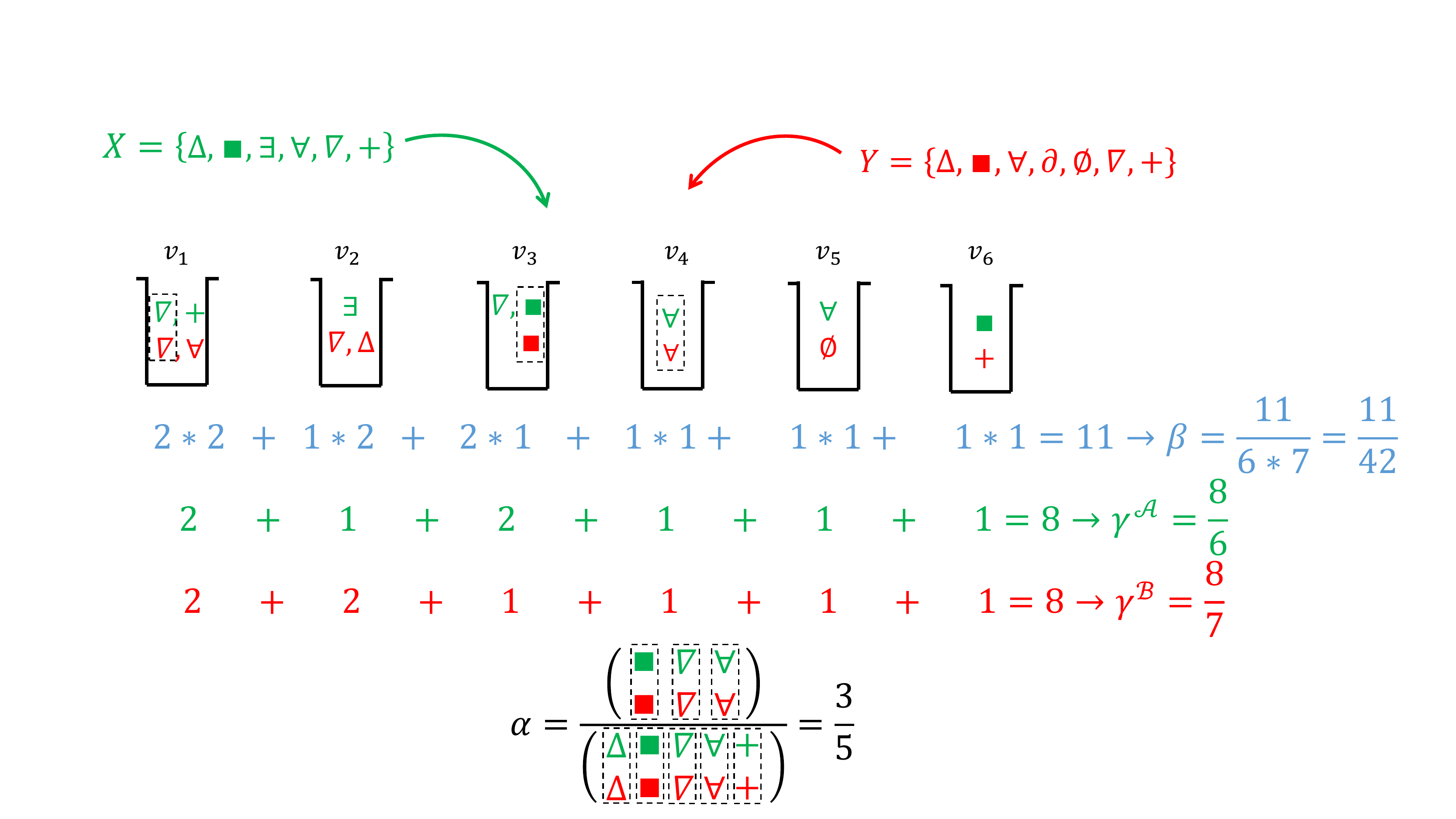}}
\caption{For the data points that are mapped to each bucket, we call them positive (see Algorithm \ref{algorithm_Max}). The rate of positive call for true and random pairs are shown as $\alpha$ and $\beta$. The average number of times $x$ and $y$ data points appear in buckets are also shown as $\gamma^{\mathcal{A}}$ and $\gamma^{\mathcal{B}}$. For formal definition of $\alpha$, $\beta$, $\gamma^{\mathcal{A}}$ and $\gamma^{\mathcal{B}}$, see Definition \ref{def_family_ds}. {\color{black}Here, we derive the empirical probabilities $(\eqref{family1}-\eqref{family2})$ in Definition \ref{def_family_ds}. For example, $\beta$ is derived by counting the number of data points combinations in all the six buckets, i.e., $2*2+1*2+2*1+1*1+1*1+1*1$ over all the possible number of data points which is $6*7$. Here the size of $X$ is six, and the size of $Y$ is 7. Similarly, $\gamma^{\mathcal{A}}$ is derived by counting number of data points from the set $X$ designated to the buckets divided by $|X|$.}}\label{fig2}
\end{figure}

\begin{definition} Define the $S$-dimensional joint probability distribution $\mathbb{P}$ as follows.
\begin{eqnarray}
\mathbb{P}:\mathcal{A}^S\times \mathcal{B}^S\rightarrow [0,1],&& \mathbb{P}(x,y)=\prod_{s=1}^S{p}(x_s,y_s),
\end{eqnarray}
where $\mathcal{A}=\{a_1,a_2,\cdots,a_k\}$, $\mathcal{B}=\{b_1,b_2,\cdots,b_l\}$, $x=(x_1,x_2,\cdots,x_S)\in \mathcal{A}^S$ and $y=(y_1,y_2,\cdots,y_S)\in\mathcal{B}^S$ { where $k,l<\infty$}. On the other hand, we assume that the probability distribution function ${p}(a,b)$ is independent of $s$ and satisfies $\sum_{a\in[k],b\in[l]}p_{a,b}=1$. Similarly, we define the marginal probability distributions $\mathbb{P}^{\mathcal{A}}:\mathcal{A}^S\rightarrow [0,1]$, $\mathbb{P}^{\mathcal{B}}:\mathcal{B}^S\rightarrow [0,1]$ and $\mathbb{Q}:\mathcal{A}^S\times \mathcal{B}^S\rightarrow [0,1]$ as $\mathbb{P}^{\mathcal{A}}(x)=\prod_{s=1}^S{p}^{\mathcal{A}}(x_s)$, $\mathbb{P}^{\mathcal{B}}(y)=\prod_{s=1}^S{p}^{\mathcal{B}}(y_s)$ and $\mathbb{Q}(x,y)=\prod_{s=1}^S{q}(x_s,y_s)$, where
\begin{eqnarray}
{p}^{\mathcal{A}}(x_s)&=&\sum_{j=1}^lp(x_s,b_j),\\
{p}^{\mathcal{B}}(y_s)&=&\sum_{i=1}^kp(a_i,y_s),\\
{q}(x_s,y_s)&=&{p}^{\mathcal{A}}(x_s)\times{p}^{\mathcal{B}}(y_s).
\end{eqnarray}
{We use $p_{ij}$ instead of $p(a_i,b_j)$ for simplicity. Moreover, $p_{i}^{\mathcal{A}}$, $p_j^{\mathcal{B}}$ and $q_{ij}$ are defined as $\sum_{j=1}^lp_{ij}$, $\sum_{i=1}^kp_{ij}$ and $q(a_i,b_j)$, respectively. {\color{black}Finally, we use compact notations $P=[p_{ij}]$ and $Q=[q_{ij}]$ as the $k\times l$ matrices with $p_{ij}$ and $q_{ij}$ in their $i$-th row and $j$-th column, respectively.}}
\end{definition}
We define family of distribution sensitive hashes as follows. 
\begin{definition}\label{def_family_ds}[Family of Distribution Sensitive Hashes] Assume that the four parameters $\alpha$, $\beta$, $\gamma^{\mathcal{A}}$ and $\gamma^{\mathcal{B}}$ along with the probability distributions  $\mathbb{P}$, $\mathbb{P}^{\mathcal{A}}$, $\mathbb{P}^{\mathcal{B}}$, $\mathbb{Q}$ and finite set $V_{Buckets}$ are given where $\mathbb{P}^{\mathcal{A}}$ and $\mathbb{P}^{\mathcal{B}}$  are marginal distributions of $\mathbb{P}$ and $\mathbb{Q}=\mathbb{P}^{\mathcal{A}}\mathbb{P}^{\mathcal{B}}$. A family of hashes $ H_z^{\mathcal{A}}:\mathcal{A}^S\rightarrow 2^{V_{Buckets}}$ and  $ H_z^{\mathcal{B}}:\mathcal{B}^S\rightarrow 2^{V_{Buckets}}$   is called $(\mathbb{P},\alpha,\beta,\gamma^{\mathcal{A}},\gamma^{\mathcal{B}})$-distribution sensitive, if the following hold
\begin{eqnarray}
\alpha&=&\sum_{v\in V_{Buckets}}Prob(v\in H_z^{\mathcal{A}}(x)\cap H_z^{\mathcal{B}}(y)\mid(x,y)\sim \mathbb{P}),\label{family1}\\
\beta&=&\sum_{v\in V_{Buckets}}Prob(v\in H_z^{\mathcal{A}}(x)\cap H_z^{\mathcal{B}}(y)\mid(x,y)\sim \mathbb{Q}),\\
\gamma^{\mathcal{A}}&=&\sum_{v\in V_{Buckets}}Prob(v\in H_z^{\mathcal{A}}(x)\mid x\sim \mathbb{P}^{\mathcal{A}}),\\
\gamma^{\mathcal{B}}&=&\sum_{v\in V_{Buckets}}Prob(v\in H_z^{\mathcal{B}}(y)\mid y\sim \mathbb{P}^{\mathcal{B}}),\label{family2}\\
&&{\mid H_z^{\mathcal{A}}(x)\cap H_z^{\mathcal{B}}(y)\mid} \le 1\label{family5},
\end{eqnarray}
\end{definition}
where $1\le z\le\#bands$ and $\#bands$ represent {the number of bands}, while $ V_{Buckets}$ represent a set of indices for the buckets.  {\color{black}We show how to choose $\#bands$ in \eqref{approx1} in Appendix C and how to select $V_{Buckets}$ in Algorithm 3. } Intuitively, $\alpha$ represents the chance of a true pair falling in the same bucket, while $\beta$ represents the chance of random pairs falling in the same bucket. We will show that $\gamma^{\mathcal{A}}$ and $\gamma^{\mathcal{B}}$ represent the complexity of   computing which buckets the data points fall into. In the next section, we will describe how the families of distribution sensitive hashes can be used to design an efficient solution for \eqref{pyx}.  Note that, in \cite{christiani2017set} definitions similar to Definition  \ref{def_family_ds} have been made in order to speed up set similarity search. {\color{black}How the data points are mapped through a decision tree is sketched in Figure \ref{fig1}. The decision tree and the set of buckets are explained in Algorithm \ref{algorithm_Tree}. Finally, in Figure \ref{fig2}, we show how the rates of positive call for true and random pairs, i.e., $\alpha$ and $\beta$ are derived. The average number of times that data points and queries fall in buckets, i.e., $\gamma^{\mathcal{A}}$ and $\gamma^{\mathcal{B}}$ are also derived and shown in this figure.}
\begin{remark} In classic LSH, we have $|H^{\mathcal{A}}(x)|=1$ and $|H^{\mathcal{B}}(y)|=1$. Therefore, $\gamma^{\mathcal{A}}=\gamma^{\mathcal{B}}=1$. Our approach is more flexible in the following two ways. First, we allow for  $\gamma^{\mathcal{A}}$ and $\gamma^{\mathcal{B}}$ to be larger than one. Moreover, even in the case of $\gamma^{\mathcal{A}}=\gamma^{\mathcal{B}}=1$, our method can optimize over the larger set of possible hashes.
\end{remark}

{\color{black} Here we present a general algorithm on how {\color{black} ForestDSH} algorithm works.
\begin{algorithm}[H]
   \caption{{\color{black} ForestDSH}}
   \label{ForestDSH_algorithmm}
\begin{algorithmic}{\color{black}
   \STATE {\bfseries Inputs:} Probability distribution $\mathbb{P}$, ${X}=\{x^1,\cdots,x^N\}\subset\mathcal{A}^S$, $y\in\mathcal{B}^S$, true positive rate $TP$ and threshold $\Delta$.
\STATE {\bfseries Output:}  Classes $x\in{X}$ satisfying $P(y\mid x)>\Delta$.
\STATE {procedure:}
\STATE {\bf Run Algorithm \ref{numerical_lambda}.}
\STATE ~~~~~~This algorithm takes probability distribution $\mathbb{P}$, $M$ and $N$ as inputs and generates parameters for decision tree design such as  $\mu^*$, $\nu^*$, $\eta^*$, $\lambda^*$, and $\delta$.
\STATE {\bf Run Algorithm \ref{algorithm_Tree}.}
\STATE ~~~~~~Here, the parameters generated in Algorithm \ref{numerical_lambda} are given as an input while it gives  $G=(V,E,f)$ and a subset of {leaf} nodes of the decision tree $V_{Buckets}$ as outputs.
\STATE {\bf Run Algorithm \ref{haha}.}
 \STATE ~~~~~~In this algorithm, list of buckets $V_{Buckets}$, permutations $perm_z,1\le z\le\#bands$, and a set of data points  $X=\{x^1,\cdots,x^N\}$ are given as input while the outputs are   $H_z^{\mathcal{A}}(x)$ for each $x\in X$ and each band $1\le z\le\#bands$. Similarly, $H_z^{\mathcal{B}}(y)$ are derived from this algorithm
\STATE {\bf Run Algorithm \ref{algorithm_Max}.}
\STATE~~~~~~This algorithm takes ${X}=\{x^1,\cdots,x^N\}\subset\mathcal{A}^S$, $y\in\mathcal{B}^S$, $\#bands, V_{Buckets}, H_z^{\mathcal{A}}(x), H_z^{\mathcal{B}}(y)$ and threshold $\Delta$ as an input and gives Classes $x\in X$ satisfying $P(y\mid x)>\Delta$ as an output.}
\end{algorithmic}
\end{algorithm}}

\section{{\color{black}Forest} Distribution Sensitive Hashing Algorithm}
In this section, we assume an oracle has given us a family of distribution sensitive hashes $H_z^{\mathcal{A}},H_z^{\mathcal{B}},\forall z:1\le z\le\#bands$ that satisfies $(\eqref{family1}-\eqref{family5})$. Inspired by LSH method, we present Algorithm \ref{algorithm_Max} for solving \eqref{pyx} using this family.
\begin{algorithm}[H]
   \caption{Solving maximum likelihood classification \eqref{pyx} by {\color{black}ForestDSH}}
   \label{algorithm_Max}
\begin{algorithmic}
   \STATE {\bfseries Inputs:} ${X}=\{x^1,\cdots,x^N\}\subset\mathcal{A}^S$, $y\in\mathcal{B}^S$, {\color{black}$\#bands, V_{Buckets}, H_z^{\mathcal{A}}(x), H_z^{\mathcal{B}}(y)$} and threshold $\Delta$.
\STATE {\bfseries Output:}  Classes $x\in{X}$ satisfying $P(y\mid x)>\Delta$.
\STATE {\bf For} $z\in\{1,2,\cdots,\#bands\}$
\STATE~~~~~~ {\bf For}   $v\in V_{Buckets}$
\STATE~~~~~~~~~~ {\bf For} $\{x\in X \mid v\in H_z^{\mathcal{A}}(x)\}$
\STATE~~~~~~~~~~~~~~ {\color{black}{\bf If} $ v\in H_z^{\mathcal{B}}(y)$}
\STATE~~~~~~~~~~~~~~~~~~~~~~Call $(x,y)$ a positive, compute $\mathbb{P}(y\mid x)$ and report $x$ if $\mathbb{P}(y\mid x)>\Delta$.
\end{algorithmic}
\end{algorithm}
\begin{remark}
Note that, the number of times $\mathbb{P}(y\mid x)$ is computed in the brute force method to solve \eqref{pyx} is  {\color{black}$|X|$. Note that, in the optimization problem \eqref{pyx}, the point $y\in Y$ is given.} The goal of Algorithm \ref{algorithm_Max} is to solve \eqref{pyx} with a much smaller number of comparisons than the brute force.
\end{remark}
\begin{remark}
{\color{black}In the special case when the hashes are coming from a decision tree, we analyze the complexity  of Algorithm \ref{algorithm_Max} in Section \ref{sec_compl}.} We show that the number of positive calls in Algorithm \ref{algorithm_Max} is proportional to $\beta$, while the complexity of computing $|H_z^{\mathcal{A}}(x)|$ and $|H_z^{\mathcal{B}}(y)|$ are proportional to $\gamma^{\mathcal{A}}$ and $\gamma^{\mathcal{B}}$. Moreover, the chance of true pairs being called positive  grows with $\alpha$. Therefore, in the next sections, we attempt to design buckets such that $\alpha$ is maximized, while $\beta$, $\gamma^{\mathcal{A}}$ and $\gamma^{\mathcal{B}}$ are minimized.
\end{remark}
Now, the question is how we can design these families in a way to minimize the complexity, and efficiently map data points to  these families. We investigate these questions in Sections \ref{sec_design} and \ref{sec_map}.
\section{Designing {\color{black}ForestDSH} Using a Decision Tree Structure}\label{sec_design}
In the previous section, we assumed that an oracle has given us a family of distribution sensitive hashes. In this section, we design buckets that satisfy $(\eqref{family1}-\eqref{family5})$ using a forest of decision trees with the same structure.  Here, we focus on the probability distributions  that can be factorized as the product of  i.i.d. components.  

Each of our decision trees recovers  ratio $\alpha$ of true pairs and by recruiting $\#bands=O(\frac{1}{\alpha})$ decision trees we can  recover nearly all true pairs. This can be more efficient than using a single decision tree classifier as achieving near perfect true pair recovery by the single decision tree would require near brute-force complexity. By allowing $\alpha<1$, we can select decision tree that avoid paths with low $\mathbb{P}(x,y)$ resulting in complexities much lower than the brute-force search. {\color{black}Note that $\alpha$ is the true positive rate, e.g. the probability that a true pair $(x,y)$ fall in the same bucket, therefore we always have $\alpha \leq 1$, see Figure \ref{fig2}.}

\begin{figure}[h]
	\centerline{\includegraphics[width=6.53in]{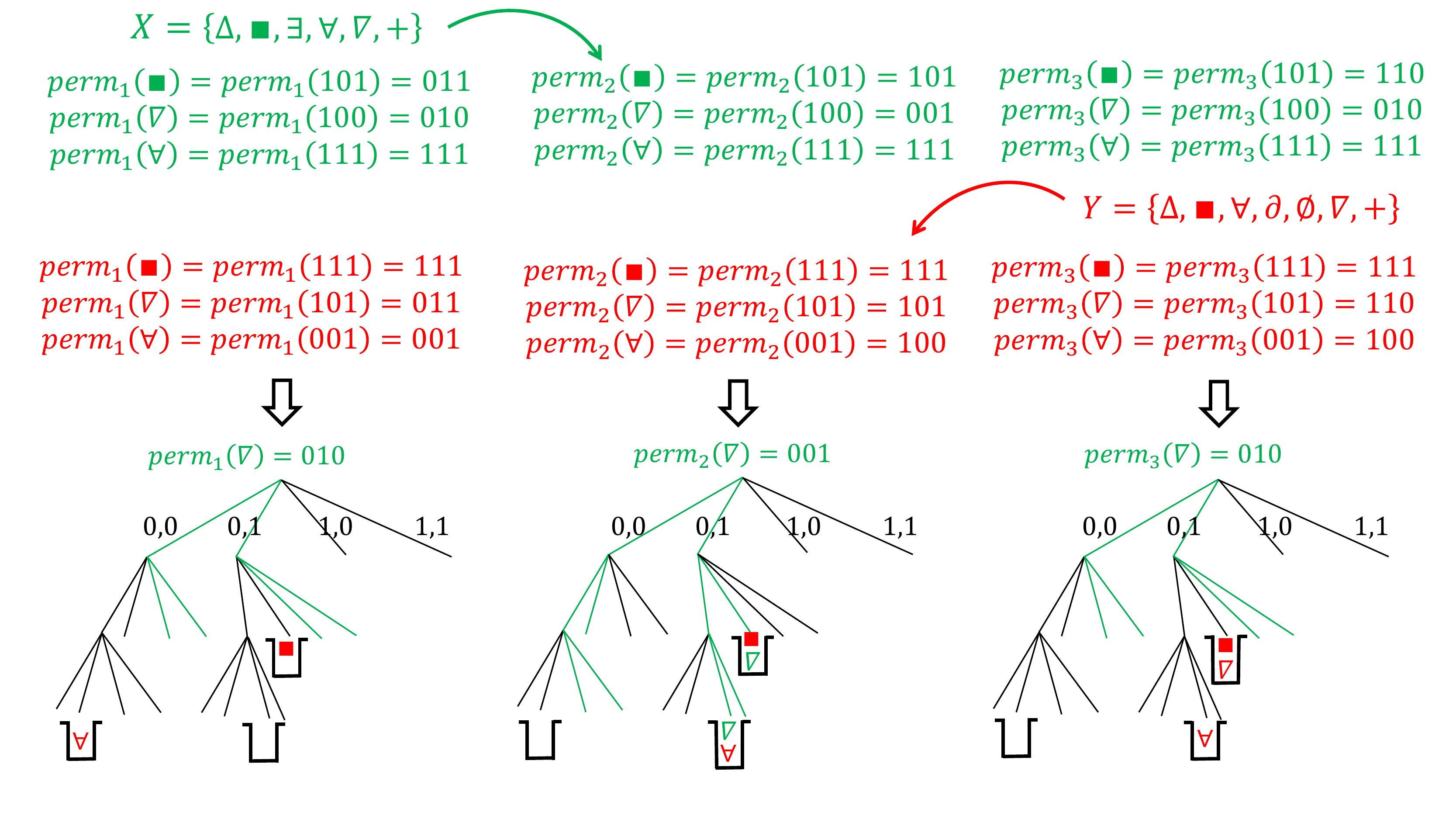}}
\caption{{}With only a single band, many of the true pairs are missed (not called positives). {\color{black}Refer to \eqref{ttt-1m} and \eqref{approx1} in order to see how multiple bands improve true positive rate. While the decision tree are not affected in any band $1\le z\le \#bands$, classes and queries are permuted by $perm_z$  (using the same permutation $perm_z$) and then mapped to the buckets in decision trees.  In this figure, all classes and queries are permuted  through three permutations $perm_1(abc)=bac$, $perm_2(abc)=cba$ and $perm_3(abc)=cab$. We showed how three of classes, i.e., ${\color{green}\blacksquare, \triangledown, \forall}$ and three of queries ${\color{red}\blacksquare, \triangledown, \forall}$ are permuted. How ${\color{green} \triangledown}$ is mapped through these three trees is shown similar to Figure \ref{fig1}. In the end, all the classes and queries designated to the buckets are shown. }Note that, a pair is called positive if they intersect in at least one of the bands.}\label{fig:treekkk}
\end{figure}

Assume that a decision tree $G=(V,E,f)$ is given  where $V$ is the set of nodes, $E$ is the set of edges, $V_l\subset V$ is the set of {leaf} nodes in the decision tree and $f: V/V_l\times\mathcal{A}\times\mathcal{B}\rightarrow V$ is the decision function, see Figure \ref{fig1}. For the two nodes $v_1,v_2 \in V$, $v_1$ is called an ancestor of $v_2$ if $v_1$ falls within the path from $v_2$ to root. In this case, $v_2$ is called a descendant of $v_1$. Furthermore, assume that a subset of leaf nodes $V_{Buckets}\subset V_l$ is given and at depth $s$ in the decision  tree, the decisions depend only  on $x_s$ and $y_s$ where $x=(x_1,\cdots,x_S)$ and $y=(y_1,\cdots,y_S)$ {\color{black}are $S$-dimensional data point and query, respectively.}  We define functions $Seq^{\mathcal{A}}:V\rightarrow \cup_{s=0}^S\mathcal{A}^s$ and $Seq^{\mathcal{B}}:V\rightarrow \cup_{s=0}^S\mathcal{B}^s$ recursively as:
\begin{eqnarray}
Seq^{\mathcal{A}}({root})&=&\varnothing\nonumber\\
Seq^{\mathcal{B}}({root})&=&\varnothing\\
Seq^{\mathcal{A}}{(f(v,a,b))}&=&[ Seq^{\mathcal{A}}(v),a]\\
Seq^{\mathcal{B}}{(f(v,a,b))}&=&[ Seq^{\mathcal{B}}(v),b]
\end{eqnarray}
{\color{black}where $[S, a]$ stands for the concatenation of string
$S$ with character $a$, i.e., for a string $S=s_1,\cdots,s_n$ of length $n$, $[S, a]$ would be $s_1,\cdots,s_n,a$ which is a string of length $n+1$.} Moreover, {\color{black}given permutations  $p_z:\{1,\cdots,S\}\rightarrow\{1,\cdots,S\}, 1\le z\le\#bands$, $x=(x_1,\cdots,x_S)$ and $y=(y_1,\cdots,y_S)$ define $perm_z(x)=(x_{p_z(1)},x_{p_z(2)},\cdots,x_{p_z(S)})$ and $perm_z(y)=(y_{p_z(1)},y_{p_z(2)},\cdots,y_{p_z(S)})$. } Finally, the family of buckets $H_z^{\mathcal{A}}(x)$ and $H_z^{\mathcal{B}}(y)$ are defined as
\begin{eqnarray}
 H_z^{\mathcal{A}}(x)&=&\{v\in V_{Buckets}\mid Seq^{\mathcal{A}}(v) \mbox{~is a prefix of perm}_z(x)\}\label{ujvx},\\
 H_z^{\mathcal{B}}(y)&=&\{v\in V_{Buckets}\mid Seq^{\mathcal{B}}(v) \mbox{~is a prefix of perm}_z(y)\}.\label{ujvy}
\end{eqnarray}
{\color{black}Note that the permutation $perm_z$ is a deterministic function of $z$, i.e., at each band the same permutation is used to randomly permute the data points $x$ and $y$. These permutations are chosen before mapping the data points. In Figure \ref{fig:treekkk}, we show how both $x$ and $y$ data points are first permuted randomly (using the same permutation) and then are mapped to the buckets in decision trees. Note that, here we call a pair positive, if they fall into the same bucket in at least one of the bands.} Now, we show that these hashes are distribution sensitive.

\begin{definition}\label{def_tree_p}
The functions $\Phi:V\rightarrow\mathbb{R}$, ${\Psi}^{\mathcal{A}}:V\rightarrow\mathbb{R}$ and ${\Psi}^{\mathcal{B}}:V\rightarrow\mathbb{R}$ are defined as follows. At root, $\Phi({root})=1$,  ${\Psi}^{\mathcal{A}}({root})=1$ and  ${\Psi}^{\mathcal{B}}({root})=1$, and for {\color{black}$a_i\in\mathcal{A}$, $b_j\in\mathcal{B}$} and $v\in V$, $\Phi(v)$, ${\Psi}^{\mathcal{A}}(v)$, and ${\Psi}^{\mathcal{B}}(v)$  are defined recursively as
\begin{eqnarray}
\Phi(f(v,a_i,b_j))&=&\Phi(v)p_{ij},\forall v\in V\label{def_a},\\
{\Psi}^{\mathcal{A}}(f(v,a_i,b_j))&=&{\Psi}^{\mathcal{A}}(v)p_{i}^{\mathcal{A}},\forall v\in V\label{def_a2},\\
{\Psi}^{\mathcal{B}}(f(v,a_i,b_j))&=&{\Psi}^{\mathcal{B}}(v)p_j^{\mathcal{B}},\forall v\in V\label{def_a3}.
\end{eqnarray}
Moreover, $\Psi:V\rightarrow\mathbb{R}$ is defined as 
\begin{eqnarray}
\Psi(v)&=&{\Psi}^{\mathcal{A}}(v){\Psi}^{\mathcal{B}}(v),\forall v\in V\label{def_b}.
\end{eqnarray}
\end{definition}

\begin{lemma}\label{lem_a0} The following properties hold:
\begin{eqnarray}
\Phi(v)&=&Prob(v\in H_z^{\mathcal{A}}(x)\cap H_z^{\mathcal{B}}(y)\mid(x,y)\sim \mathbb{P})\label{fam1},\\
\Psi(v)&=&Prob(v\in H_z^{\mathcal{A}}(x)\cap H_z^{\mathcal{B}}(y)\mid(x,y)\sim \mathbb{Q})\label{famm1},\\
{\Psi}^{\mathcal{A}}(v)&=&Prob(v\in H_z^{\mathcal{A}}(x)\mid x\sim \mathbb{P}^{\mathcal{A}}),\\
{\Psi}^{\mathcal{B}}(v)&=&Prob(v\in H_z^{\mathcal{B}}(y)\mid y\sim \mathbb{P}^{\mathcal{B}})\label{fam2}.
\end{eqnarray}
\end{lemma}
\begin{remark}Note that the left side of $(\eqref{fam1}-\eqref{fam2})$  is independent of $z$ while it seems like the right side depend on band $z$. The proof of Lemma \ref{lem_a0} in Appendix \ref{app_lem_12} shows that in fact it is independent of band $z$.
\end{remark}

\begin{lemma}\label{lem_a1} For any decision tree $G$, satisfying the condition that  for any pair of buckets $v_1,v_2\in V_{Buckets}$, $v_1$ is not an ancestor or descendant of $v_2$,  $H_z^{\mathcal{A}}(x)$ and $H_z^{\mathcal{B}}(y)$ defined in \eqref{ujvx} and \eqref{ujvy} are $(\mathbb{P},\alpha(G),\beta(G),\gamma^{\mathcal{A}}(G),\gamma^{\mathcal{B}}(G))$-sensitive where
\begin{eqnarray}
\alpha(G)&=&\sum_{v\in V_{Buckets}(G)}\Phi(v)\label{alphdef}\label{alphaav},\\
\beta(G)&=&\sum_{v\in V_{Buckets}(G)}\Psi(v)\label{betdef}\label{betabv},\\
\gamma^{\mathcal{A}}(G)&=&\sum_{v\in V_{Buckets}(G)}{\Psi}^{\mathcal{A}}(v)\label{gammxdef},\\
\gamma^{\mathcal{B}}(G)&=&\sum_{v\in V_{Buckets}(G)}{\Psi}^{\mathcal{B}}(v)\label{gammydef}.
\end{eqnarray}
\end{lemma}
Proofs of Lemmas \ref{lem_a0} and \ref{lem_a1} { are} relegated to Appendix \ref{app_lem_12}. So far, we showed how to design {\color{black} ForestDSH} using a decision tree structure. However, it is not yet clear how to map data points to these buckets. In the next section, we investigate this and provide an algorithm to design optimal decision trees.

\subsection{Mapping Data Points}\label{sec_map}
In the previous sections, we presented Algorithm \ref{algorithm_Max} for solving \eqref{pyx} where we need to compute $H_z^{\mathcal{A}}(x)$ and  $H_z^{\mathcal{B}}(y)$ by mapping data points to the buckets. We did not clarify how this mapping can be done efficiently. We present Algorithm \ref{haha} for mapping data points to the buckets using hash-table search,   see Figure \ref{fig2}.
\begin{algorithm}[H]
   \caption{Mapping data points to the buckets using hash-table search}
\label{haha}   
\begin{algorithmic}
   \STATE ~~{\bfseries Inputs:} List of buckets $V_{Buckets}$, permutations $perm_z,1\le z\le\#bands$, and a set of data points  $X=\{x^1,\cdots,x^N\}$.
\STATE ~~{\bfseries Outputs:}  $H_z^{\mathcal{A}}(x) $ for each $x\in X$ and each band $1\le z\le\#bands$.
\STATE~~Create an empty hash-table.
\STATE ~~{\bf Initialize} $H_z^{\mathcal{A}}(x)=\varnothing$ for all $x\in X$ and $1\le z\le\#bands$.
\STATE~~{\bf For} $v\in V_{Buckets}$
\STATE~~~~~~~Insert $Seq^{\mathcal{A}}(v)$ into the hash-table.
\STATE~~{\bf For} $z=1$ to $\# bands$
\STATE~~~~~~~{\bf For} {$x\in {X}$}
\STATE~~~~~~~~~~~Search  $perm_{z}(x)$ in the hash-table to find all $v\in V_{Buckets}$ for which  $Seq^{\mathcal{A}}(v)$ is a prefix of $perm_{z}(x)$, and insert $v$ into $H_z^{\mathcal{A}}(x)$ {{} (see Figure \ref{fig:treekkk})}.
\end{algorithmic}
\end{algorithm}
\begin{remark} Queries are mapped to buckets using hash-table search similar to data points through Algorithm \ref{haha}.
\end{remark}
Note that we slightly modify the hash-table to search for values that are prefix of a query, rather than being exactly identical. In Section \ref{sec_compl}, we show that the complexity of Algorithms   \ref{algorithm_Max} and \ref{haha} can be formulated as:
\begin{align}
&c_{tree}|V(G)|\nonumber\\
&+\left(\frac{c_{hash}N}{\alpha(G)}+\frac{c_{hash}M}{\alpha(G)}+\frac{c_{insertion}N\gamma^{\mathcal{A}}(G)}{\alpha(G)}+\frac{c_{insertion}M\gamma^{\mathcal{B}}(G)}{\alpha(G)}+\frac{c_{pos}MN\beta(G)}{\alpha(G)}\right){\log \frac{1}{1-TP}},\label{complexity_sum_all2}
\end{align}
where $c_{tree}$, $c_{hash}$, $c_{insertion}$ and $c_{pos}$ are constants not depending on $N$. {\color{black}For the intuition behind \eqref{complexity_sum_all2}, note that} the first term $c_{tree}|V(G)|$ stands for the time required for calculating and storing the tree. The second and third terms, i.e., $\left(\frac{c_{hash}N}{\alpha(G)}+\frac{c_{hash}M}{\alpha(G)}\right){\log \frac{1}{1-TP}}$ denote the time needed for inserting data points to the hash-table. The fourth and fifth terms, i.e., $\left(\frac{c_{insertion}N\gamma^{\mathcal{A}}(G)}{\alpha(G)}+\frac{c_{insertion}M\gamma^{\mathcal{B}}(G)}{\alpha(G)}\right){\log \frac{1}{1-TP}}$ stand for the time required for mapping the data points from the hash-table to buckets. Finally, the last term, i.e., $\left(\frac{c_{pos}MN\beta(G)}{\alpha(G)}\right){\log \frac{1}{1-TP}}$ is the time of brute-force checking within each bucket. 

In order to bound \eqref{complexity_sum_all2} with $O(N^{\lambda})$ for some $\lambda\in\mathbb{R}^+$, it is  necessary and sufficient  to find a tree $G$ that satisfies the following constraints:
\begin{eqnarray}
|V(G)|&=&O(N^{\lambda})\label{order1},\\
\frac{\alpha(G)}{\beta(G)} &=&\Omega(N^{1+\delta-\lambda})\label{order111},\\
\frac{\alpha(G)}{\gamma^{\mathcal{A}}(G)} &=& \Omega(N^{1-\lambda}),\label{order112}\\
\frac{\alpha(G)}{\gamma^{\mathcal{B}}(G)} &=& \Omega(N^{\delta-\lambda}),\\
{\alpha(G)} &=& \Omega(N^{\max(1,\delta)-\lambda})\label{order3},
\end{eqnarray}
where ${\delta}=\frac{\log M}{\log N}$.

{\color{black}\begin{theorem} \label{theorem_main} 
Complexity of Algorithms \ref{algorithm_Max} and \ref{haha} is equal to \eqref{complexity_sum_all2}. Moreover, there exists a decision tree $G$ that is optimal and satisfies $(\eqref{order1}-\eqref{order3})$ for $\lambda$ defined in Definition $4$.  
\end{theorem}
Theorem \ref{theorem_main} is proved in three steps: $(a)$. In Section \ref{sec_compl}, we prove that the complexity is equal to \eqref{complexity_sum_all2}. $(b)$. Theorem \ref{theoremA} shows that the tree $G$ constructed by Algorithm \ref{algorithm_Tree} and presented in Section \ref{sec_con_DSH} satisfies (\eqref{order1}-\eqref{order3}) for $\lambda$ defined in Definition 4. $(c)$. Theorem \ref{theoremB} shows that this is  optimal.}

\section{Complexity Analysis}\label{sec_compl}
In this section, we bound the complexity of Algorithms \ref{algorithm_Max} {\color{black}and \ref{haha} by \eqref{complexity_sum_all2}}. Note that,  the complexity of Algorithms \ref{algorithm_Max} {\color{black}and \ref{haha}} is the summation of the following terms.
\begin{enumerate}
\item{\bf Tree construction complexity.} $c_{tree}|V(G)|$ is the tree construction complexity where $c_{tree}$ is a constant representing per node complexity of  constructing a node and  $|V(G)|$  is the number of nodes in the tree.
\item{\bf Data mapping complexity.} The complexity of this hash-table search grows with 
\begin{eqnarray}
&&c_{hash}(\#bands)|{X}|+c_{insertion}\sum_{z=1}^{\#bands}\sum_{x\in X}|H_z^{\mathcal{A}}(x)|\\
&&+c_{hash}(\#bands)|{Y}|+c_{insertion}\sum_{z=1}^{\#bands}\sum_{y\in Y}|H_z^{\mathcal{B}}(y)|,
\end{eqnarray}
where  $c_{hash}$ and $c_{insertion}$ represent complexity of insertion in the hash-table and insertion in the buckets, respectively. 
\item{\bf Complexity of checking positive calls.} $(\#bands)c_{pos}\sum_{x\in X,y\in Y}|H_z^{\mathcal{A}}(x) \cap H_z^{\mathcal{B}}(y) |\big)$ is the complexity of checking positive calls where $c_{pos}$ is a constant representing the complexity of computing {$\mathbb{P}(y\mid x)$} for a positive. Note that, $c_{tree}$, $c_{hash}$, $c_{insertion}$ and $c_{pos}$ are constants not depending on $N$. From $(\eqref{family1}-\eqref{family5})$, we have
\begin{eqnarray}
&&\mathbb{E}\big(\sum_{z=1}^{\#bands}\sum_{x\in X}|H_z^{\mathcal{A}}(x) |\big)\nonumber\\
&=&{{}\mathbb{E}\big(\sum_{z=1}^{\#bands}\sum_{x\in X,v\in V_{Buckets}(G)}1_{x\in v}\big)}\\
&=&(\#bands)N\sum_{v\in V_{Buckets}(G)}{\Psi}^{\mathcal{A}}(v)=(\#bands)N\gamma^{\mathcal{A}}(G)\\
&&\mathbb{E}\big(\sum_{z=1}^{\#bands}\sum_{y\in Y}|H_z^{\mathcal{B}}(y) |\big)\nonumber\\
&=&\mathbb{E}\big(\sum_{z=1}^{\#bands}\sum_{y\in Y,v\in V_{Buckets}(G)}1_{y\in v}\big)\\
&=&(\#bands)M\sum_{v\in V_{Buckets}(G)}{\Psi}^{\mathcal{B}}(v)=(\#bands)M\gamma^{\mathcal{B}}(G).
\end{eqnarray}
Note that, the total number of collision for random pairs is the sum of number of buckets that they intersect at. Therefore, we conclude that 
\begin{eqnarray}
&&\mathbb{E}\big(\sum_{z=1}^{\#bands}\sum_{x\in X,y\in Y}|H_z^{\mathcal{A}}(x) \cap H_z^{\mathcal{B}}(y) |\big)\nonumber\\
&=&\sum_{z=1}^{\#bands}\sum_{x\in X,y\in Y}Prob\big(|H_z^{\mathcal{A}}(x) \cap H_z^{\mathcal{B}}(y) |=1\big)\label{fd_c}\\
&=&\sum_{z=1}^{\#bands}{\color{black}\sum_{x\in X,y\in Y}\sum_{v\in V_{Buckets}(G)}}Prob\big(v\in H_z^{\mathcal{A}}(x) ,v\in H_z^{\mathcal{B}}(y) \big)\\
&=&(\#bands)MN\sum_{v\in V_{Buckets}(G)}\Psi(v)=(\#bands)MN\beta(G),
\end{eqnarray}
where \eqref{fd_c} is concluded as ${\mid H^{\mathcal{A}}(x)\cap H^{\mathcal{B}}(y)\mid}\le 1$. 
\end{enumerate}
Now, the question is how we can select $\#bands$ such that the true positive rate, defined as the ratio of true pairs that are called positive is high. In each band, the chance of a pair $(x,y)\sim\mathbb{P}$ being called positive is computed as $\alpha(G)=\sum_{v\in V_{Buckets}(G)}\Phi(v)$. Therefore, the overall true positive rate can be computed as:
\begin{eqnarray}
TP&=&Prob((x,y) \mbox{~called positive in Algorithm \ref{algorithm_Max}~}\mid (x,y)\sim\mathbb{P})\label{chagh}\\
&=&1-\prod_{z=1}^{\#bands}\left(1-\sum_{v\in V_{Buckets}} Prob(v\in H_z^{\mathcal{A}}(x)\cap H_z^{\mathcal{B}}(y)\mid(x,y)\sim\mathbb{P})\right)\\
&=&1-{(1-\alpha(G))}^{\#bands}\label{ttt-1m}.
\end{eqnarray}
Using \eqref{ttt-1m}, and the inequality $(1-x)^{\frac{c}{x}}< e^{-c}$, the minimum possible value of $\#bands$ to ensure true positive rate $TP$ can be computed as 
\begin{eqnarray}
\#bands&=&\lceil\frac{\log \frac{1}{1-TP}}{\alpha(G)}\rceil\label{approx1},
\end{eqnarray}
where $\lceil r\rceil$ stands for the smallest integer greater than or equal to $r$. Therefore, the total complexity is equal to \eqref{complexity_sum_all2}.

\section{Constructing optimal decision trees for {\color{black} ForestDSH}}\label{sec_con_DSH}
\begin{figure}[h]
	\centerline{\includegraphics[width=5.13in]{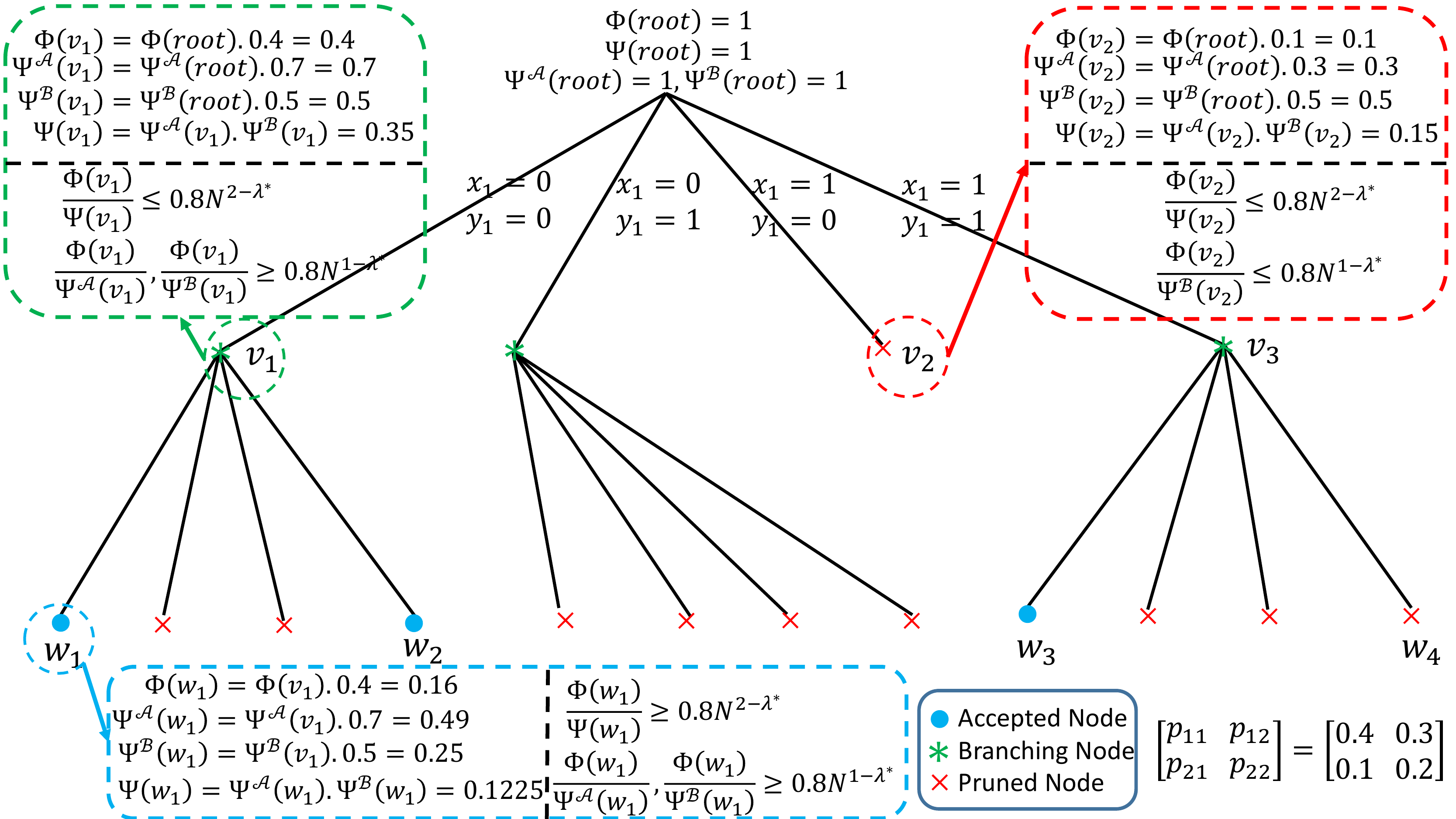}}
\caption{The decision tree and functions $\Phi(v)$, ${\Psi}^{\mathcal{A}}(v)$, ${\Psi}^{\mathcal{B}}(v)$ and $\Psi(v)$ are illustrated   for $\mathcal{A}=\{0,1\},\mathcal{B}=\{0,1\}$, {\color{black} $C_1=C_2=C_3=0.8$   and $N=5$.} For all of the nodes $v_1$, $v_2$ and $w_1$, how the decisions are made are explained in the figure. For instance, consider the node $w_1$. For this node, $\Phi(w_1)$, ${\Psi}^{\mathcal{A}}(w_1)$, ${\Psi}^{\mathcal{B}}(w_1)$ are derived from $(\eqref{def_a}-\eqref{def_a3})$. On the other hand, from \eqref{ok} this node is accepted as a bucket as $\frac{\Phi(w_1)}{\Psi(w_1)}=1.31\ge 0.8N^{2-\lambda^*}=1.26$. Note that, $\lambda^*=1.72$ from \eqref{in0} and Algorithm \ref{numerical_lambda}.}\label{fig:tree}
\end{figure}

In this section, we present an algorithm to design decision trees with complexity $O(N^{\lambda^*})$, where $\lambda^*$ is defined below, and we show that it is the optimal decision tree.
\begin{definition}\label{deflambda}Given probability distributions  $P=[p_{ij}]$ and  $Q=[q_{ij}]$, $1 \leq i \leq k, 1 \leq j \leq l$, and number of queries  and classes  $M$ and $N$  define $\delta=\frac{\log M}{\log N}$   and
\begin{eqnarray}
\mathcal{I}&=& \{(\mu, \nu,\eta) \in \mathbb{R}^3 | \min(\mu,\nu)\ge\eta \geq 0,\nonumber\\
&& \sum_{1 \leq i \leq k, 1 \leq j \leq l}p_{ij} ^ {1+\mu+\nu-\eta} {(p_{i}^{\mathcal{A}})} ^ {-\mu}  {(p_j^{\mathcal{B}})} ^ {-\nu}  =1 \}\label{def_mathI},\\
(\mu^*,\nu^*,\eta^*)&=&Arg\max_{\mathcal{I}}\frac{\max(1,\delta)+\mu+\nu\delta}{1+\mu+\nu-\eta}\label{defmunu},\\
r_{ij}^* &=&{\color{black}p_{ij} ^ {1+\mu^*+\nu^*-\eta^*} {(p_{i}^{\mathcal{A}})} ^ {-\mu^*}  {(p_j^{\mathcal{B}})} ^ {-\nu^*}} \label{rijstarrrr},\\
n^*&=&{\color{black}\frac{(\max(1,\delta)-\lambda^*) \log N}{\sum r_{ij}^*\log \frac{p_{ij}}{r_{ij}^*}}}\label{nstarrr},\\
\lambda^*&=&\frac{\max(1,\delta)+\mu^*+\nu^*\delta}{1+\mu^*+\nu^*-\eta^*}\label{in0}.
\end{eqnarray}
\end{definition}
\begin{remark}For any probability distribution $\mathbb{P}$, the parameters $\mu^*$, $\nu^*$, $\eta^*$ and $\lambda^*$ can be derived numerically from Algorithm \ref{numerical_lambda} in ~Appendix~ \ref{app_numerical}. {\color{black} 
The intuition~ behind ~the~ definition~ of ~ $\mathcal{I}$ and $(\mu^*,\nu^*,\eta^*)$ is that in Lemma \ref{induction_AB} in Section \ref{proof_theoremA} we show that for any decision tree $G$ and the variables $\Phi(v)$, ${\Psi}^{\mathcal{A}}(v)$, ${\Psi}^{\mathcal{B}}(v)$ and $\Psi(v)$ defined in \eqref{def_a}-\eqref{def_b}, we have\\$\sum_{v \in V_{Buckets}(G)}{\big(\Phi(v)\big)}^{1+\mu+\nu-\eta}{\big({\Psi}^{\mathcal{A}}(v)\big)}^ {-\mu +\eta}{\big({\Psi}^{\mathcal{B}}(v)\big)}^ {-\nu +\eta}{\big(\Psi(v)\big)}^ {-\eta } \le 1$ if  $(\mu ,\nu,\eta)\in \mathcal{I}$. Moreover, in proof of Theorem 2 we show that  $(\mu ^ *,\nu ^ *,\eta^*)$ are Lagrangian multipliers in an optimization problem to minimize the search complexity in Algorithm \ref{algorithm_Max} while retaining a nearly perfect recovery. {\color{black}Consider the optimal decision tree and all the nodes $v$ satisfying
\begin{enumerate}
\item Depth of the node $v$ is $n^*$.
\item For any $1\le i\le k,1\le j\le l$, the ratio of times we have $a_i$ at $s$-th position of $x$ and $b_j$ at $s$-th position of $y$ in all $1\le s\le n^*$ is $r_{ij}$.
\end{enumerate}
The node $v$ or one of its ancestors is designated as a bucket by Algorithm \ref{algorithm_Tree}. This is proved in Section \ref{proof_theoremB}.} }\end{remark}
In Algorithm \ref{algorithm_Tree}, we provide an approach for designing decision trees with complexity $O(N^{\lambda^*})$. The algorithm starts with the root, and at each step,  it either accepts a node as a bucket, prunes  a node, or branches a node into $kl$ children based on the following constraints\footnote{If for any node the constraints for accepting as a bucket and pruning hold simultaneously, the algorithm accepts the node as a bucket. }:
\begin{eqnarray}
\left\{\begin{matrix}\frac{\Phi(v)}{\Psi(v)} \ge C_1 {N ^ {1+\delta-\lambda^*}}&: \mbox{Accept bucket},~~~~~~~~~~~~~~~~\\
\frac{\Phi(v)}{{\Psi}^{\mathcal{A}}(v)}\le C_2N^{1-\lambda^*}&: \mbox{{}Prune},~~~~~~~~~~~~~~~~~~~~~~~~~~~\\
\frac{\Phi(v)}{{\Psi}^{\mathcal{B}}(v)}\le C_3N^{\delta-\lambda^*}&: \mbox{{}Prune,}~~~~~~~~~~~~~~~~~~~~~~~~~~~\\
otherwise&: \mbox{Branch into the $kl$ children.} \end{matrix}\right.\label{ok}
\end{eqnarray}
{\color{black}Note that, for the cases where $p_{ij}=0$, we simply remove that branch, therefore, we assume all the $p_{ij}$ are positive real numbers.} In Section \ref{sec_compl}, we prove that in order to bound the complexity in \eqref{complexity_sum_all2} with  $O(N^{\lambda})$ for some $\lambda\in\mathbb{R}^+$, it is  necessary and sufficient  to find a decision tree $G$ that satisfies the constraints $(\eqref{order1}-\eqref{order3})$. 
{\color{black}\begin{remark}\label{remmm}
In Theorem \ref{theoremB}, we prove that the decision  tree construction of Algorithm \ref{algorithm_Tree}  results in a tree with complexity $O(N^{\lambda^*})$ by setting $C_1=C_2=C_3=p_0q_0$ where   $p_{0}$ and $q_{0}$ are defined as $\prod_{i,j}p_{ij}$ and $\min(\prod_{i,j}q_{ij},\prod_{i}{(p_i^{\mathcal{A}})}^{l},\prod_j{(p_j^{\mathcal{B}})}^{k})$. Pessimistic lower bounds $C_1=C_2=C_3=p_0q_0$ are derived in proof of Theorem \ref{theoremB} and in practice  $C_1$, $C_2$ and $C_3$ are chosen in a way that the best complexity is achieved. 
\end{remark}}
\begin{example}\label{exam1}Here, we focus on the case where $\mathcal{A}=\{0,1\},\mathcal{B}=\{0,1\}$, $P=\begin{bmatrix}
0.4 & 0.3  \\ 
0.1 & 0.2 
\end{bmatrix}$,  $Q=\begin{bmatrix}
0.35 & 0.35  \\ 
0.15 & 0.15 
\end{bmatrix}$, $\delta=1$, and $M=N=4$. From Algorithm \ref{numerical_lambda}, we have $\mu^*=12.0791$, $\nu^*=13.4206$, $\eta^*=11.0959$, $\lambda^*=1.7203$. The  decision tree is constructed from Algorithm \ref{algorithm_Tree} and is depicted in Figure \ref{fig:tree}.  The nodes in the tree that are selected as bucket, i.e., satisfying $\frac{\Phi(v)}{\Psi(v)}\ge C_1{N^{1+\delta-\lambda^*}}$, are shown with a green check mark, and the nodes pruned out, satisfying either $\frac{\Phi(v)}{{\Psi}^{\mathcal{A}}(v)}\le C_2N^{1-\lambda^*}$ or $\frac{\Phi(v)}{{\Psi}^{\mathcal{B}}(v)}\le C_3N^{\delta-\lambda^*}$ are shown with a red cross. For the non-leaf (intermediate) nodes, none of the above constraints holds.  The bold edges show the paths in the tree corresponding to the data point $x=(0,0)$. In this case, $x$ falls into a single bucket $w_1$.
\end{example}
\begin{algorithm}[h]
   \caption{Recursive construction of the decision tree}
   \label{algorithm_Tree}
\begin{algorithmic}
   \STATE {\bfseries Inputs:} {\color{black}$C_1$, $C_2$, $C_3$,} $\delta$, $\mathcal{A}$, $\mathcal{B}$, ${P}$,  $M$ and $N$. $\#$ We use Algorithm \ref{numerical_lambda} to derive $\mu^*,\nu^*,\eta^*,\lambda^*$, $\delta$.
\STATE {\bfseries Outputs:}  $G=(V,E,f)$ and a subset of {leaf} nodes of the decision tree $V_{Buckets}$.
\STATE{\bfseries Initialization}: 
\STATE~~~~$Seq^{\mathcal{A}}{(root)}\leftarrow\varnothing$, $Seq^{\mathcal{B}}{(root)}\leftarrow\varnothing$.
\STATE~~~~$\Phi(root)\leftarrow1$,~${\Psi}^{\mathcal{A}}(root)\leftarrow1$,~${\Psi}^{\mathcal{B}}(root)\leftarrow1$,~$\Psi(root)\leftarrow1$.
\STATE~~~~Recursive $TreeConstruction(root)$.
\STATE{\bfseries Procedure}  $TreeConstruction(v)$.
\STATE~~~~{\bf For} $a_i\in\mathcal{A}$
\STATE~~~~~~~~{\bf For} $b_j\in\mathcal{B}$
\STATE~~~~~~~~~~~~Create a new node $w$.~~$\#$ The node is created only if $p_{ij}\neq0$.
\STATE~~~~~~~~~~~~$\Phi(w)\leftarrow \Phi(v)p_{ij}$
\STATE~~~~~~~~~~~~${\Psi}^{\mathcal{A}}(w)\leftarrow {\Psi}^{\mathcal{A}}(v)p_{i}^{\mathcal{A}}$
\STATE~~~~~~~~~~~~${\Psi}^{\mathcal{B}}(w)\leftarrow {\Psi}^{\mathcal{B}}(v)p_j^{\mathcal{B}}$
\STATE~~~~~~~~~~~~$\Psi(w)\leftarrow {\Psi}^{\mathcal{A}}(w){\Psi}^{\mathcal{B}}(w)$
\STATE~~~~~~~~~~~~$f(v,a_i,b_j)\leftarrow w$
\STATE~~~~~~~~~~~~$Seq^{\mathcal{A}}(w)\leftarrow Seq^{\mathcal{A}}(v)$, $Seq^{\mathcal{B}}(w)\leftarrow Seq^{\mathcal{B}}(v)$
\STATE~~~~~~~~~~~~$Seq^{\mathcal{A}}(w).append(a_i)$, $Seq^{\mathcal{B}}(w).append(b_j)$
\STATE~~~~~~~~~~~~{\bf  If} $\frac{\Phi(w)}{\Psi(w)}\ge C_1{N^{1+\delta-\lambda^*}}$ ~~~~~~~~~~~~~~~~~~~~~~~~~~~~~~~~~~~~~~~~~~~~~~~~~~~~~~~~$\#$Accept bucket
\STATE~~~~~~~~~~~~~~~~$V_{Buckets}.insert(w)$
\STATE~~~~~~~~~~~~{\bf Else If}
$\frac{\Phi(w)}{{\Psi}^{\mathcal{A}}(w)}\ge   C_2N^{1-\lambda^*}$ and $\frac{\Phi(w)}{{\Psi}^{\mathcal{B}}(w)}\ge  C_3N^{\delta-\lambda^*}$
~~~~~~~~~~~~~~~~~~$\#$Branch
\STATE~~~~~~~~~~~~~~~~$TreeConstruction(w)$
\STATE~~~~~~~~~~~~{\bf Else}~~~~~~~~~~~~~~~~~~~~~~~~~~~~~~~~~~~~~~~~~~~~~~~~~~~~~~~~~~~~~~~~~~~~~~~~~~~~~~~~$\#$Prune 
\STATE~~~~~~~~~~~~~~~~$f(v,a_i,b_j)\leftarrow null$.
\end{algorithmic}
\end{algorithm}
\begin{theorem}\label{theoremA} No decision tree exists with overall complexity below $O(N^{\lambda^*})$.
\end{theorem}
\begin{theorem}\label{theoremB} The decision  tree construction of Algorithm \ref{algorithm_Tree} described in Remark \ref{remmm}  results in a tree with complexity $O(N^{\lambda^*})$. 
\end{theorem}
In other words, Theorem \ref{theoremB} proves that the tree $G$ constructed by Algorithm \ref{algorithm_Tree} satisfies (\eqref{order1}-\eqref{order3}), and   Theorem \ref{theoremA} shows that this is the optimal decision tree. For proofs of Theorems \ref {theoremA} and \ref{theoremB}, see Sections \ref{proof_theoremA} and \ref{proof_theoremB}. Note that, not only Theorem \ref{theoremB} guarantees that the number of the nodes in our decision tree is bounded by $O(N^{\lambda^*})$ but also it guarantees that the run time for mapping the data points to the decision tree and the number of comparisons that we need to do for the nodes with the collision is bounded by $O(N^{\lambda^*})$, see complexity equation \eqref{complexity_sum_all2} in Section \ref{sec_map}.
\begin{theorem}[Noise Robustness]\label{theorem_noise} Assume that the decision  tree of Algorithm \ref{algorithm_Tree} described in \eqref{ok} is constructed based on distribution  $P=[p_{ij}]$  while the data is generated from unknown $P'=[p'_{ij}]$ satisfying $\frac{p_{ij}}{1+\epsilon}\le p'_{ij}\le{p_{ij}}{(1+\epsilon)}$ for all $1\le i\le k, 1\le j\le l$ and some $\epsilon>0$. Then, it is possible to design an algorithm with complexity $O(N^{\lambda^{*}(p)+3c_d\log(1+\epsilon)})$ where $c_d=  \frac{{(\lambda^*-\min(1,\delta))}}{\log(\max_{i,j}\min(\frac{p_{ij}}{p_i^{\mathcal{A}}},\frac{p_{ij}}{p_j^{\mathcal{B}}}))}$ while maintaining arbitrary high true positive rate $TP$.  
\end{theorem}
In other words, Theorem \ref{theorem_noise} proves that the tree $G$ constructed by Algorithm \ref{algorithm_Tree} is noise robust, i.e., for the case where our understanding from the distribution is not exact, the arbitrary high true positive rate $TP$  can still be maintained while the complexity increases linearly with the noise. For proof of Theorem \ref{theorem_noise}, see Section \ref{proof_theorem_noise}. For a high level intuition behind the relationship between Algorithm \ref{algorithm_Tree} and Theorems \ref{theoremA} and \ref{theoremB}, note that: Every constraint in the decision tree defined by constraints \eqref{ok} directly relates to the constraints $(\eqref{order1}-\eqref{order3})$. Therefore, we expect this decision tree to be optimal. \\

{\color{black}Here, we present an intuition behind proof of Theorem  \ref{theoremA}. For rigorous proof of Theorem \ref{theoremA}, see Section \ref{rigor_theoremA}.
\subsection{Intuition behind proof of Theorem \ref{theoremA}}\label{proof_theoremA}
Note that,  from definition of $\Phi(v)$, ${\Psi}^{\mathcal{A}}(v)$, ${\Psi}^{\mathcal{B}}(v)$ and $\Psi(v)$ in $(\eqref{def_a}-\eqref{def_b})$, we have $\Phi(w_{ij})=p_{ij}\Phi(v)$ where $v$ is the parent of $w_{ij}$. Similar equations hold for ${\Psi}^{\mathcal{A}}(v)$, ${\Psi}^{\mathcal{B}}(v)$ and $\Psi(v)$. On the other hand, from \eqref{def_mathI}, note that for any $(\mu ,\nu ,\eta)\in \mathcal{I}$ we have $\sum_{1 \leq i \leq k, 1 \leq j \leq l}p_{ij} ^ {1+\mu+\nu-\eta} {(p_{i}^{\mathcal{A}})} ^ {-\mu}  {(p_j^{\mathcal{B}})} ^ {-\nu}  =1 $. Therefore, we expect to have similar relation between $\Phi(v)$, ${\Psi}^{\mathcal{A}}(v)$, ${\Psi}^{\mathcal{B}}(v)$, $\Psi(v)$, i.e., 
\begin{eqnarray}
\sum_{v \in V_{Buckets}(G)}{\big(\Phi(v)\big)}^{1+\mu+\nu-\eta}{\big({\Psi}^{\mathcal{A}}(v)\big)}^ {-\mu +\eta}{\big({\Psi}^{\mathcal{B}}(v)\big)}^ {-\nu +\eta}{\big(\Psi(v)\big)}^ {-\eta } \le 1,
\end{eqnarray}
 for any  $(\mu ,\nu ,\eta)\in \mathcal{I}$. On the other hand, from $(\eqref{alphaav}-\eqref{gammydef})$, we have
$\alpha(G)=\sum_{v\in V_{Buckets}(G)}\Phi(v)$. Similar definitions hold for $\beta(G)$, $\gamma^{\mathcal{A}}(G)$ and $\gamma^{\mathcal{B}}(G)$. Therefore, from convexity of the function $f(\theta,\theta_1,\theta_2,\theta_3)={\theta} ^ {1+\rho_1+\rho_2+\rho_3} {\theta_1} ^{-\rho_1} {\theta_2}^ {-\rho_2} {\theta_3} ^{-\rho_3}$ we conclude 
\begin{eqnarray}
1\ge{\big(\alpha(G)\big)}^ {1+\mu^*+\nu^*-\eta^*} {\big(\gamma^{\mathcal{A}}(G)\big)}^{-\mu ^ *+\eta^*} {\big(\gamma^{\mathcal{B}}(G)\big)}^{-\nu ^ *+\eta^*} {\big(\beta(G)\big)}^{-\eta ^ *}\label{hasfasf},
\end{eqnarray}
using the lower bounds on $\frac{\alpha(G)}{\beta(G)}$, $\frac{\alpha(G)}{\gamma^{\mathcal{A}}(G)}$, $\frac{\alpha(G)}{\gamma^{\mathcal{B}}(G)}$ and ${\alpha(G)} $ in $(\eqref{order1}-\eqref{order3})$ and \eqref{hasfasf}, we conclude $\lambda\ge\lambda^*$. } \\

{\color{black}Here, we present an intuition behind proof of Theorem  \ref{theoremB}. For rigorous proof of Theorem \ref{theoremB}, see Appendix \ref{rigor_proof_B}.
\subsection{Intuition behind proof of Theorem \ref{theoremB}}\label{proof_theoremB}\label{proof_lem_tree}
Here, our goal is to prove that the tree construction steps in Algorithm \ref{algorithm_Tree}, i.e., \eqref{ok} result in a tree that satisfies 
\begin{enumerate}
\item  There is at least one node in the tree which is accepted as a bucket.
\item 	$(\eqref{order111}-\eqref{order3})$ holds.
\item  Expected number of nodes in the tree is bounded by $O(N^{\lambda^*})$.
\end{enumerate}
Let us intuitively prove all these three statements one by one.
\begin{enumerate}
\item   Consider a node with the depth equal to $n^*$ given in \eqref{nstarrr}. Moreover, assume that the number of times we have $a_i$ at $s$-th position of $x$ and $b_j$ at $s$-th position of $y$ for all $1\le s\le n^*$ are $n^*r^*_{ij}$. Then, we argue that this  node is not pruned and accepted as a bucket. In order to understand why intuitively it is true, we verify that this node is accepted as a bucket, i.e., we have to verify that $\frac{\Phi(v)}{\Psi(v)} \ge {N ^ {1+\delta-\lambda^*}}{p_0q_0}$. This is true as  
\begin{eqnarray}
\frac{\Phi(v)}{\Psi(v)}&=&\Omega(e^{\sum_{i,j} n^*r^*_{ij}\log \frac{p_{ij}}{q_{ij}}})\label{t-e1}\\
&\ge&\Omega({N ^ {1+\delta-\lambda^*}})\label{t-e2},
\end{eqnarray}
\eqref{t-e1} follows from \eqref{def_a}, \eqref{def_a2} and the definition of node $v$, i.e.,  the number of times we have $a_i$ at $s$-th position of $x$ and $b_j$ at $s$-th position of $y$ are $n^*r^*_{ij}$. \eqref{t-e2} is concluded from the definition of $r^*_{ij}$ in \eqref{rijstarrrr}. For further details, see Section \ref{rigor_proof_B}.
\item 	Let us prove \eqref{order111} as the rest of  $(\eqref{order112}-\eqref{order3})$ follow similarly. From Algorithm \ref{algorithm_Tree}, for all the buckets we have
\begin{eqnarray}
\frac{\Phi(v)}{\Psi(v)}&\ge&{N ^ {1+\delta-\lambda^*}}p_0q_0.
\end{eqnarray}
Note that, at least there is one node that is accepted a a bucket. On the other hand, $\alpha(G)=\sum_{v\in V_{Buckets}(G)}\Phi(v)$ and $\beta(G)=\sum_{v\in V_{Buckets}(G)}\Psi(v)$. Therefore, we conclude that
\begin{eqnarray}
\frac{\alpha(G)}{\beta(G)}&=&\frac{\sum_{v\in V_{Buckets}(G)}\Phi(v)}{\sum_{v\in V_{Buckets}(G)}\Psi(v)}  \geq N^{1+\delta - \lambda^*}p_{0}q_{0}.
\end{eqnarray}
Note that $\frac{\sum_i{a_i}}{\sum_ib_i}\ge c$ if $\frac{a_i}{b_i}\ge c$ and $b_i>0$ for any $i$.
\item  Finally, we prove that number of nodes in the tree is bounded by $O(N^{\lambda^*})$. Note that $\sum_{i,j}p_{ij}=1$, therefore, we expect to have $\sum_{v\in V_l(G)}\Phi(v)=1$. On the other hand, $\Phi(v)$ is expected to be greater than $N^{-\lambda^*}$ for the intermediate nodes from \eqref{ok}. Therefore, it is concluded that $|V_l(G)|$ is at most $N^{\lambda^*}$. This results in $|V(G)|=O(N^{\lambda^*})$ as for any decision tree we have $|V(G)|\le2|V_l(G)|$. For details, see Appendix \ref{rigor_proof_B}.
\end{enumerate}}.

{\color{black}Here, we present an intuition behind proof of Theorem  \ref{theorem_noise}. For rigorous proof of Theorem \ref{theorem_noise}, see Appendix \ref{rigor_proof_C}.
\subsection{Intuition behind proof of Theorem \ref{theorem_noise}}\label{proof_theorem_noise}
Define $\Phi'(v), \alpha'(G), \beta'(G), \#bands'$ for $p'_{ij}$ the same way as $\Phi(v), \alpha(G), \beta(G), \#bands$ for $p_{ij}$.  As $\frac{p_{ij}}{1+\epsilon}\le p'_{ij}\le{p_{ij}}{(1+\epsilon)}$, we expect $\Phi(v)$ to be bounded as
\begin{eqnarray}
 \Phi(v)\frac{1}{{(1+\epsilon)}^d}\le \Phi'(v)\le \Phi(v){(1+\epsilon)}^d.
\end{eqnarray}
Therefore, from  $\eqref{alphaav}$ we have
\begin{eqnarray}
\alpha(G)&=&\sum_{v\in V_{Buckets}(G)}\Phi(v)\\
&\le&\sum_{v\in V_{Buckets}(G)}\Phi'(v) {(1+\epsilon)}^d\\
&\le&\alpha'(G) (1+\epsilon)^d.
\end{eqnarray}
We conclude similar inequalities for  $\gamma^{\mathcal{A}}(G)$ and $\gamma^{\mathcal{B}}(G)$, while for $\beta(G)$ we have 
\begin{eqnarray}
\beta'(G) (1+\epsilon)^{2d}\le \beta(G)\le \beta'(G) \frac{1}{(1+\epsilon)^{2d}}.  
\end{eqnarray}
From \eqref{approx1}, $\#bands$ is inversely  related to $\alpha(G)$. Therefore, $\#bands'$ can be bounded from above by ${(1+\epsilon)}^d\#bands$. As a result, from \eqref{complexity_sum_all2} the total complexity is bounded by ${(1+\epsilon)}^{3d}N^{\lambda^*}$ from above. Assume that $d$ is bounded by $c_d\log N$ for some constant $c_d$. Therefore, we conclude Theorem \ref{theorem_noise}. In order to see why $d\le c_d\log N$, note that for all the leaf nodes, we have
  $\frac{\Phi(v)}{{\Psi}^{\mathcal{A}}(v)}\le{N ^ {1+\delta-\lambda^*}}\max_{i,j}\frac{p_{ij}}{p_i^{\mathcal{A}}}p_0q_0$. On the other hand, $\frac{\Phi(v)}{{\Psi}^{\mathcal{A}}(v)}$ can be bounded by ${\big( \min_{i,j}\frac{p_{ij}}{p_i^{\mathcal{A}}}\big)}^d$ from below. Therefore, we conclude that  $d=c_d\log N$ for some constant $c_d$.}

\section{ Experiments and Observations}\label{sec_examp}
\begin{experiment}\label{exp_dubiii}In this experiment, we compare the complexity of {\color{black} ForestDSH} with the algorithm proposed by Dubiner in \cite{Moshe_Heterogeneous}. {\color{black} Here, we set  $\mathcal{A}=\mathcal{B}=\{0,1\}, S=1000, M=N=1000,000$. For ForestDSH, We use Algorithm \ref{numerical_lambda} to derive $\mu^*,\nu^*,\eta^*,\lambda^*$ and $\delta$ while training $C_1$, $C_2$ and $C_3$ to get the best complexity in Algorithm \ref{algorithm_Tree}.} For Dubiner's algorithm, equation $(126)$ in  \cite{Moshe_Heterogeneous} is  computationally intractable which makes it impossible to compute the complexity for the algorithm presented there for general probability distributions.  {\color{black}However, in the special case where $P(p)=\begin{bmatrix}
\frac{p}{2} & \frac{1-p}{2}  \\ 
\frac{1-p}{2} & \frac{p}{2}
\end{bmatrix}$, $0.5\le p\le 1$ (hamming distance), a solution has been provided for computing the complexity in \cite{Moshe_Heterogeneous}, and we implemented that solution (see Appendix \ref{pseudo_code} for the detail of implementation), and compared it to ForestDSH. Note that currently no source code for the implementation of Dubiner algorithm is available.} Figure \ref{x_dub}, shows that Dubiner algorithm's performance is worse than that of  {\color{black} ForestDSH}. 
\begin{figure}[h]
	\centerline{\includegraphics[width=3.1in]{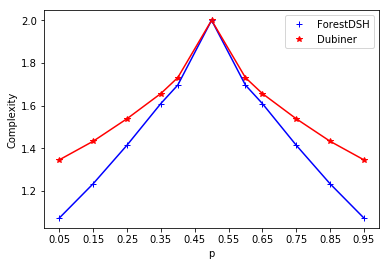}}
\caption{Comparing the practical performances of Dubiner algorithm in \cite{Moshe_Heterogeneous} with {\color{black} ForestDSH} for $S=1000$. {\color{black} ForestDSH} outperforms Dubiner's method for all values of $p$. {\color{black}The pseudo code for the algorithm presented in \cite{Moshe_Heterogeneous} for the case of hamming distance is generated in this paper in Algorithm \ref{d_dub_d} in Appendix \ref{pseudo_code}.} }\label{x_dub}
\end{figure}

\end{experiment}

\begin{observation} {\color{black}In this observation, we reformulate the problem of solving \eqref{pyx_2} to the minimum inner product search problem (MIPS) \cite{shrivastava2014asymmetric}} by transferring the data points from $\mathcal{A}^S$ and $\mathcal{B}^S$ to $R^{klS}$ in a way that $\log\big(\frac{\mathbb{P}(x,y)}{\mathbb{Q}(x,y)}\big)$ is equal to the dot product in this new space.  We transformed $x\in\mathcal{A}^S$ and $y\in \mathcal{B}^S$ to $T(x)\in\mathbb{R}^{klS}$ and $T(y)\in \mathbb{R}^{klS}$  as follows:
\begin{eqnarray}
T(x)&=&\big(f_{s,i,j}\big),1\le s\le S,1\le i\le k,1\le j\le l,\label{g_f1}\\
f_{s,i,j}&=&\left\{\begin{matrix}
\frac{\log\big(\frac{p_{ij}}{q_{ij}}\big)}{\omega_{ij}}& \mbox{if~}x_s=a_i\\
0& \mbox{o.w.}\end{matrix}\right.,\\
T(y)&=&\big(g_{s,i,j}\big),1\le s\le S,1\le i\le k,1\le j\le l,\\
g_{s,i,j}&=&\left\{\begin{matrix}{\omega_{ij}}& \mbox{if~}y_s=b_j\\
0& \mbox{o.w.}\end{matrix}\right..\label{g_f2}
\end{eqnarray}
Then, we have $\log\left(\frac{\mathbb{P}(x,y)}{\mathbb{Q}(x,y)}\right)=<T(x),T(y)>$ where $<.,.>$ stands for the inner product in $\mathbb{R}^{klS}$. In other words, given any $x=(x_1,\cdots,x_S)$, for each $x_s$ we transform it to a $kl\times1$ vector with $l$ non-zero elements. Similarly, given any $y=(y_1,\cdots,y_S)$, each $y_s$ is transformed to a $kl\times1$ vector with $k$ non-zero elements. Therefore, finding pairs of data points with large $\frac{\mathbb{P}(x,y)}{\mathbb{Q}(x,y)}$ is equivalent to finding transformed data points with large dot product. Using this transformation, in Appendix \ref{mips_app} we show that the angle between both the true pairs and false  pairs will be nearly $\frac{\pi}{2}$ for almost all the probability distributions ($\frac{S_0}{M^2}\approx0$, using the notation from \cite{shrivastava2014asymmetric}). It is well known that MIPS performs poorly in detection of pairs that are nearly orthogonal  \cite{shrivastava2014asymmetric}. Therefore, solving \eqref{pyx_2} by transforming it to a MIPS problem and using existing approaches fails. {{}Note that, \cite{shrivastava2014asymmetric} is based on data independent hashes for Euclidean distance that are not the state of the art. Recently, data dependent hashes have been introduced for Euclidean distance that improve on their data independent counterparts \cite{andoni2017optimal,rubinstein2018hardness}, while currently there is no data dependent strategy for maximum inner product search. {\color{black}Therefore, MIPS is currently unable to solve \eqref{pyx} efficiently.}  }
\end{observation}

\begin{experiment}\label{exp_a}  In this experiment, we compared the complexity for the three algorithms LSH-hamming, MinHash and ForestDSH for a range of probability distributions. We benchmark the three methods using matrices ${P}(t)={P}_1(1-t)+{P}_2t$ where $0\le t\le1$, ${P}_1=\begin{bmatrix}
0.345 & 0  \\ 
0.31 & 0.345 
\end{bmatrix}$,  ${P}_2=\begin{bmatrix}
0.019625 & 0  \\ 
0.036875 & 0.9435 
\end{bmatrix}$, and $\delta=1$, i.e., $M=N$. {\color{black}The selection of ${P}_1$ was such that the complexity of MinHash minus the complexity of LSH-hamming was maximized. ${P}_2$ was selected such that the complexity of  LSH-hamming minus the complexity of MinHash was maximized.}  Fig. \ref{fig:complexity_p(x)} $(a)$ shows the theoretical complexities of MinHash, LSH-hamming and ForestDSH for each matrix. See Appendix \ref{LSH_MinHash}, for the details on the derivation of complexities for MinHash, LSH-hamming and ForestDSH. For instance, for ${P}_1=\begin{bmatrix}
0.345 & 0  \\ 
0.31 & 0.345 
\end{bmatrix}$,  the theoretical per query complexities of MinHash, LSH-hamming and ForestDSH are equal to $0.5207$, $0.4672$ and $0.4384$, respectively. We further consider $N$ data points of dimension $S$, $\{x^1,\cdots,x^N\}$ and $\{y^1,\cdots,y^N\}$ where each $(x^i,y^i)$ is generated from ${P}(t)$, and $x^i$ is independent from $y^j$ for $i\neq j$ $(N=2000,S=2000)$.  Then, we used ForestDSH, LSH-hamming and MinHash to find the matched pairs. In each case, we tuned {\color{black}$\#rows$\footnote{\color{black}In MinHash and LSH-Hamming, we start with $\#bands\times \#rows$ randomly selected hashes from the family of distribution sensitive hashes, where $\#rows$ is the number of rows and $\#bands$ is the number of bands. We recall a pair $(x,y)$, if $x$ and $y$ are hashed to the same value in all $\#rows$ rows and in at least one of the $\#bands$ bands.}} and $\#bands$ to achieve $99\%$ true positive (recall) rate. Total simulation time for each of the three methods is plotted for each probability distribution in Figure \ref{fig:complexity_p(x)} $(b)$. The simulation times in Figure \ref{fig:complexity_p(x)} $(b)$ are consistent with the theoretical guarantees in Figure \ref{fig:complexity_p(x)} $(a)$.  Figure \ref{fig:complexity_p(x)} parts $(b)$ and $(c)$ show that for sparse matrices, $(t\approx1)$, MinHash and ForestDSH outperform LSH-hamming. In denser cases, $(t\approx0)$, LSH-hamming and ForestDSH outperform MinHash. For $(t\le0.4)$, ForestDSH outperforms both MinHash and LSH-hamming. {\color{black} Note that, for the case when the data is coming from sparse distribution, MinHash beats ForestDSH in practice (as opposed to Theory). This is because for sparse data the total complexity tends to its minimum, i.e., $O(N)$. ForestDSH is inefficient compared to MinHash for small number of $N$ and when the data is sparse as the constant terms and $log N$ terms which play a significant role in this case are not optimized in ForestDSH.} In Figure \ref{compare_VNG}, we further plotted  $V(G(N))$, $\frac{\alpha(G(N))}{\beta(G(N))}$, $\frac{\alpha(G(N))}{\gamma^{\mathcal{A}}(G(N))}$ and $\frac{\alpha(G(N))}{\gamma^{\mathcal{B}}(G(N))}$ as a function of $N$ for trees $G(N)$ constructed by Algorithm \ref{algorithm_Tree}  for ${P}(t=0.25)$, where $M=N$. As predicted by Theorem \ref{theoremB}, we observed that these quantities grow/decay proportional to $N^{\lambda^*}$, $N^{1+\delta-\lambda^*}$, $N^{1-\lambda^*}$ and $N^{\delta-\lambda^*}$, respectively. {\color{black}In Figure \ref{fig_b_TP_tt}, true positive rate and total complexity are plotted in terms of $\#bands$ for the case of $N=20,000$, $S=10,000$ and the probability distribution ${P}_2$. From \eqref{ttt-1m}, i.e., $TP=1-{(1-\alpha(G))}^{\#bands}$, we expect true positive rate to increase while $\#bands$ increases (see Figure \ref{fig_b_TP_tt} $(a)$). On the other hand, from  \eqref{approx1} and \eqref{complexity_sum_all2}, total complexity $(N^{\lambda^*})$ is expected to linearly increase while $\#bands$ increases (see Figure \ref{fig_b_TP_tt} $(b)$).}
\begin{figure}[H]
\centering
\centerline{\includegraphics[width=6in]{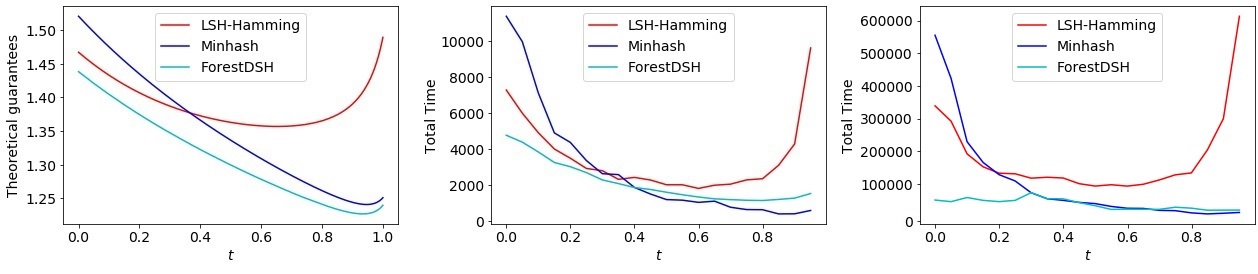}}
\begin{minipage}[c]{\textwidth}
~~~~~~~~~~~~~~~~~~~~~~~~~$(a)$~~~~~~~~~~~~~~~~~~~~~~~~~~~~~~~~~~~~$(b)$~~~~~~~~~~~~~~~~~~~~~~~~~~~~~~~~~~~~~$(c)$
\end{minipage}\\
\caption{{\color{black} Total (opposed to per query)} complexities of LSH-hamming, MinHash, and ForestDSH  are plotted for all the probability distribution matrices ${P}(t)={P}_1(1-t)+{P}_2t$  where $0\le t\le1$. $(a)$ Theoretical guarantees, $(b)$ Simulation time for $N=2000$ and  $S=2000$, $(c)$ Simulation time for $N=20000$ and  $S=2000$. {\color{black}Note that ForestDSH performs faster than MinHash if the data is not sparse. While for N=2,000 MinHash is superior to ForestDSH on sparse data, when N=20,000, ForestDSH is performing the same as MinHash. This is mainly due to the fact that constant and $\log N$ terms vanish compared to $N$ as $N$ grows.} } \label{fig:complexity_p(x)}
\end{figure}
\begin{figure}[H]
\centerline{\includegraphics[width=5.7in]{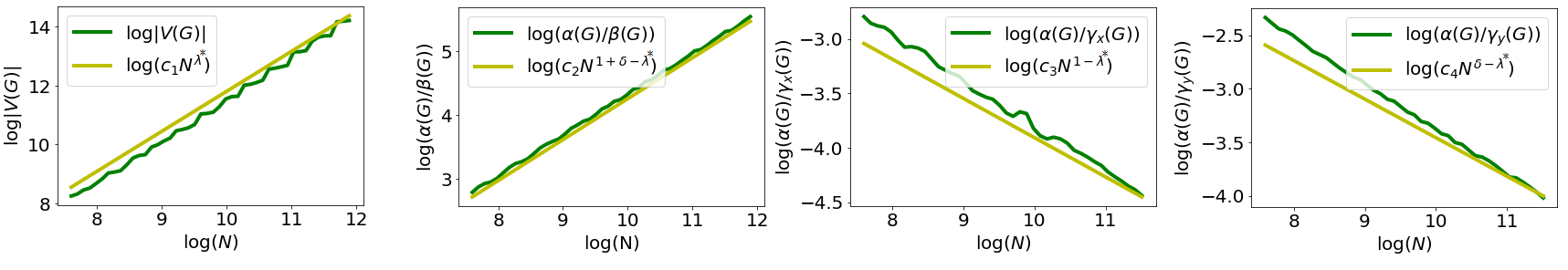}}
\caption{\color{black}In this figure, $V(G(N))$, $\frac{\alpha(G(N))}{\beta(G(N))}$, $\frac{\alpha(G(N))}{\gamma^{\mathcal{A}}(G(N))}$ and $\frac{\alpha(G(N))}{\gamma^{\mathcal{B}}(G(N))}$ are depicted as a function of $N$ and constants {\color{black}$c_1$, $c_2$, $c_3$ and $c_4$ not depending on $N$}. The parameters are chosen from Experiment \ref{exp_a}. This figure shows how these functions grow/decay proportional to $N^{\lambda^*}$, $N^{1+\delta-\lambda^*}$, $N^{1-\lambda^*}$ and $N^{\delta-\lambda^*}$, respectively. (See proof of Theorem \ref{theoremB} in Section \ref{proof_theoremB} for the justifications. This confirms (\eqref{order1}-\eqref{order3}) for trees $G(N)$ constructed by Algorithm \ref{algorithm_Tree}). }\label{compare_VNG}
\end{figure} 
\begin{figure}[H]
\centerline{\includegraphics[width=6.3in]{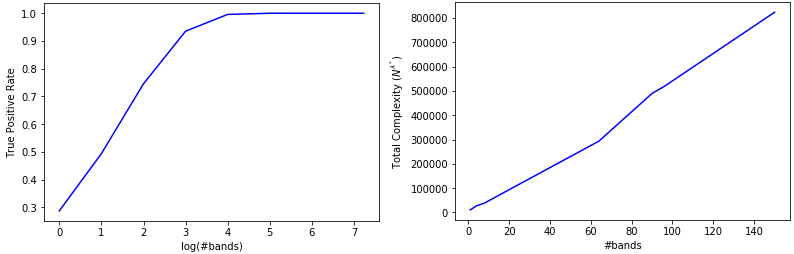}}
\begin{minipage}[c]{0.54\textwidth}
~~~~~~~~~~~~~~~~~~~~~~~~~~~~~~~~~~~~~~~~~~~~~~~~~~~$(a)$~~~~~~~~~~~~~~~~~~~~~~~~~~~~~~~~~~~~~~~~~~~~~~~~~~~~~~~~~~~~~~~~~~~~~~~~~~~~~~~~~~~~~~~~~~$(b)$
\end{minipage}
\caption{{\color{black} $(a)$ True positive rate and $(b)$ total complexity $(N^{\lambda^*})$  are plotted for the probability distribution matrix ${P}_2(t)$, $N=2000$ and  $S=2000$ in terms of $\#bands$. }}
\label{fig_b_TP_tt} 
\end{figure}
\end{experiment}

\begin{experiment}\label{exp_mass}In this experiment, we examine the robustness of our algorithm from Theorem \ref{theorem_noise} for  matrices considered in Experiment \ref{exp_a} while $M=N=300,000$. For each ${P}(t)$, we generated random matrices ${P}'(t)$ within $1+\epsilon$ of ${P}(t)$, see Theorem \ref{theorem_noise}. We derive the practical complexity and theoretical complexity, i.e., $\lambda^{*}$ from Definition \ref{deflambda}. Moreover, for each of these matrices, we derive the worst case complexity in the case of $\epsilon=0.03$. This is sketched in Figure \ref{noise_fig}.
\begin{figure}[H]
\centerline{\includegraphics[width=3.1in]{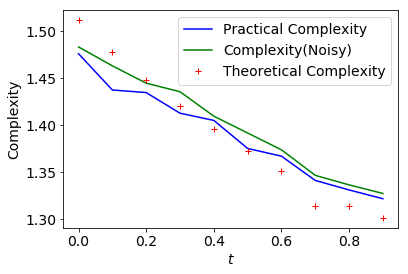}}
\caption{\color{black}In this figure, practical, theoretical and noisy complexity are compared for ${P}(t)={P}_1(1-t)+{P}_2t$ given in Experiment \ref{exp_mass}. Here, $M=N=300,000$ and the noisy complexity is derived from Theorem \ref{theorem_noise} with $\epsilon=0.03$.}
\label{noise_fig}
\end{figure}
\end{experiment}

\begin{figure}[H]
\centerline{\includegraphics[width=5.651in]{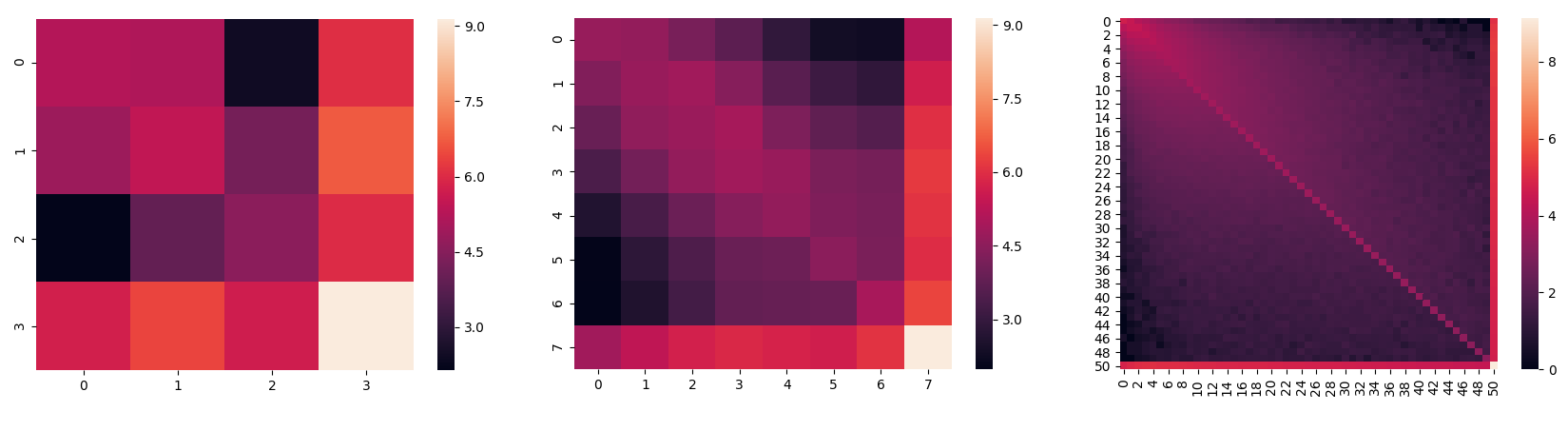}}
~~~~~~~~~~~~~~~~~~~~~~~$(a)$~~~~~~~~~~~~~~~~~~~~~~~~~~~~~~~~~~~$(b)$~~~~~~~~~~~~~~~~~~~~~~~~~~~~~~~~~~~~$(c)$
\caption{Mass spectrometry joint probability distribution in the case of $(a)$ $\log_4 Rank$, $(b)$ $\log_2 Rank$, and $(c)$ no $\log Rank$ filter. {\color{black}For further details on mass spectrometry joint probability distributions, see Appendix \ref{app_mass}.} }
\label{mass_5151}
\end{figure}

\begin{experiment}\label{exp_mass}
In this experiment, we used the mass spectral data from human microbiome \cite{mcdonald2018american}, extracted using MSCluster \cite{frank2011spectral}. In this problem, there are $N$ spectra $X={x^1,\cdots,x^N}$, and a query spectrum $y$. Our goal is to find the spectra $x^i$ that maximize a probabilistic model $\mathbb{P}(y|x^i)$. $\mathbb{P}(y|x)$ is learned assuming that it can be factorized to i.i.d. components. To learn $p(y\mid x)$,  each mass spectra is sorted based on the peak intensities, and in order to reduce the number of parameters we need to learn,  instead of the peak ranks we use log of the peak ranks \footnote{$\log_b Rank$ of a peak is defined as the {\color{black}$\log_b$} of its $rank$.  For any natural number $n$, {\color{black}$\log_b Rank=n$ for  the peaks at rank $\{b^{n-1},\cdots,b^{n}-1\}$, e.g., $\log_2 Rank(m)=3$ for $ m \in\{4,5,6,7\}$.} Joint probability distribution of logRanks for the data from \cite{frank2011spectral} is shown in Figures \ref{mass_5151} {\color{black}$(a)$ ($4\times4$ data matrix is obtained using $\log_4 Rank$), $(b)$ ($8\times8$ data matrix is obtained using $\log_2 Rank$) and $(c)$ ($51\times51$ data matrix is obtained not using any $\log_b Rank$)}.}. Using these data, we learn the joint probability distribution $p(\log Rank(y_s)=i\mid\log Rank(x_s)=j)$.  {\color{black}Among 90753 data points, we used 70753 for training, and 20000 for the test.} The joint probability distribution of matching pairs are shown in Figures \ref{mass_5151}, before and after logRank transformation.
 
After learning this probabilistic model, we applied ForestDSH method to the test mass spectra (after $\log Rank$ transformation), and we were able to speed up the search of  $20000$ mass spectra nine times, in {\color{black}comparison} to brute force search {\color{black}while maintaining true positive  rate of $90\%$.  For ForestDSH method, the total runtime is $705,197ms$, while for brute force, the total runtime is 6,022,373ms (eight times slower than ForestDSH), and for LSH, the runtime for achieving $90\%$ TP rate is 5,495,518ms (seven times slower than ForestDSH).} 
The amount of memory used peaks at 220MB.
\end{experiment}

\begin{experiment}
{\color{black} In this experiment, we considered a set of $50,000$ images ($25,000$ pairs of images) from CIFAR-10 database \cite{krizhevsky2009learning}. Each pair of images consists of a grey-scale $32*32$ pixels image and a noisy version of the image constructed by adding independent Gaussian noise $\mathcal{N}(\mu=0.5,\sigma=0.039)$ and discretizing pixels to binary. {\color{black}ForestDSH} is able to detect true pairs with the success rate of $99\%$ in 1,970 seconds while brute force approach detects true pairs in 19,123 seconds (nine times slower than ForestDSH) and LSH detects true pairs within 15,080 seconds (seven times slower than ForestDSH).} 
\end{experiment}
\begin{figure}[H]
\centering
\begin{minipage}{\textwidth}
\centering
\includegraphics[width=.18\linewidth]{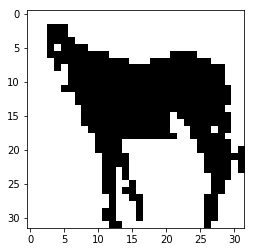}
\includegraphics[width=.18\linewidth]{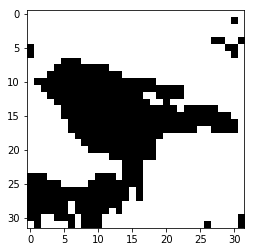}
\includegraphics[width=.18\linewidth]{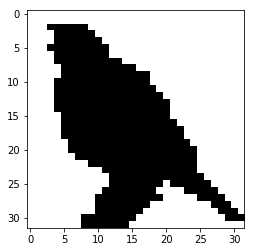}
\includegraphics[width=.18\linewidth]{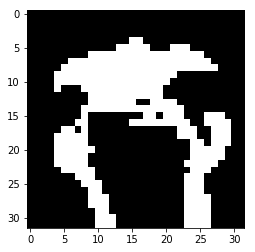}
\includegraphics[width=.18\linewidth]{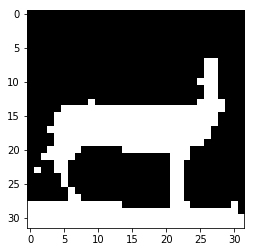}
\end{minipage}
\begin{minipage}{\textwidth}
~~~~~~~~ $\downarrow+\mathcal{N}(\mu,\sigma)$~~~~~~~~~$\downarrow+\mathcal{N}(\mu,\sigma)$~~~~~~~~~~$\downarrow+\mathcal{N}(\mu,\sigma)$ ~~~~~~~~$\downarrow+\mathcal{N}(\mu,\sigma)$~~~~~~~~~$\downarrow+\mathcal{N}(\mu,\sigma)$
\end{minipage}
\begin{minipage}{\textwidth}
\centering
\includegraphics[width=.18\linewidth]{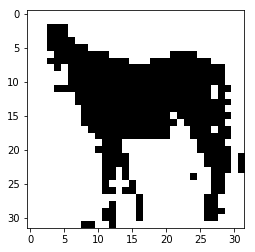}
\includegraphics[width=.18\linewidth]{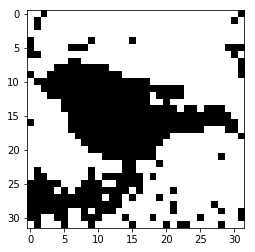}
\includegraphics[width=.18\linewidth]{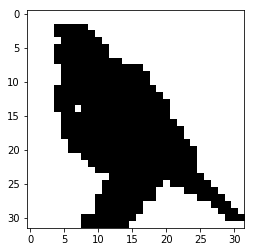}
\includegraphics[width=.18\linewidth]{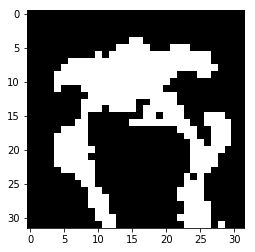}
\includegraphics[width=.18\linewidth]{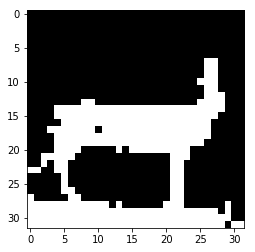}
\end{minipage}
\caption{\color{black} In this figure, five grey-scale $32*32$ pixels images along with their noisy versions are shown. Here,  ForestDSH is able to detect true pairs with the success rate of $99\%$ while being nine times faster than brute force and seven times faster than LSH.}
\end{figure}

\subsection{Codes}
{\color{black}For the codes, see https://github.com/mohimanilab/ForestDSH.}

\section{Conclusion}
ForestDSH algorithm proposed in this paper is comprehensive and efficient in the sense that for a wide range of probability distributions it enables us to capture the difference between pairs coming from a joint probability distributions and independent pairs. This algorithm is built upon a family of distribution sensitive hashes and is designed using a decision tree structure which is constructed recursively.  Moreover, we prove that the decision tree introduced here has a complexity of $O(N^{\lambda^*})$ and there is no decision tree with overall complexity below it. We prove that this algorithm {\color{black}outperforms existing state of the art approaches in specific range of distributions, in theory and practice and} enabled speeding up the spectral library search in mass spectrometry by a factor of nine. {\color{black}Finally, we should note that ForestDSH has some limitations which are discussed as follows.
ForestDSH assumes that the probability distribution function is factorizable to i.i.d. components. Generalizing ForestDSH to the case of general probability distribution function is an open problem and our results here open a path towards solving the general probability mass function problem. Moreover, the special case of factorizable models are simple but crucial models that have been widely used in computational biology{\color{black}, e.g., see \cite{MSGF}}, and other areas of data sciences. In the future, we will address the more general problem of Markov chain models. Note that ForestDSH performs faster than MinHash if the data is not sparse. While for a small number of datapoints MinHash is superior to ForestDSH on sparse data, for a larger number of datapoints ForestDSH is performing the same as MinHash. This is mainly due to the fact that constant and logarithmic terms vanish in ratio as the number of datapoints  grows. }


\appendix
\section{Proof of Lemmas \ref{lem_a0} and \ref{lem_a1}}\label{app_lem_12}
\subsection{ Proof of Lemma \ref{lem_a0}.} Lemma \ref{lem_a0} is proved by induction on the depth of the node $v$. For example, consider $v$ and its children $w_{ij}=f(v,a_i,b_j)$. From \eqref{def_a}, we have $\Phi(w_{ij})=\Phi(v)p_{ij}$. Therefore, \eqref{fam1} is proved by induction as follows. Assume \eqref{fam1} holds for any node with depth less than $d$. Consider the node $w_{ij}$ which is a children of {\color{black}$v$, i.e., $w_{ij}=f(v,a_i,b_j)$ and $depth(w_{ij})=d$.   }
\begin{eqnarray}
\Phi(w_{ij})&=&\Phi(v)p_{ij}\\
&=& Prob(v\in H_z^{\mathcal{A}}(x)\cap H_z^{\mathcal{B}}(y)\mid(x,y)\sim \mathbb{P})p_{ij}\label{tp1}\\
&=& Prob(Seq^{\mathcal{A}}(v) \mbox{~is a prefix of perm}_z(x)\nonumber\\
&&, Seq^{\mathcal{B}}(v) \mbox{~is a prefix of perm}_z(y)\mid(x,y)\sim \mathbb{P})p_{ij}\label{tp2}\\
&=& Prob(Seq^{\mathcal{A}}(v) \mbox{~is a prefix of perm}_z(x)\nonumber\\
&&, Seq^{\mathcal{B}}(v) \mbox{~is a prefix of perm}_z(y),a_i={({perm_z}(x))}_d,b_j={\color{black}{({perm_z}(y))}_d}\mid(x,y)\sim \mathbb{P})\nonumber\\
\label{tp3}\\
&=& Prob(w_{ij}\in H_z^{\mathcal{A}}(x)\cap H_z^{\mathcal{B}}(y)\mid(x,y)\sim \mathbb{P})\label{tp4}.
\end{eqnarray}
See Section 4 for the Definition of $perm_z$. {\color{black}Note that ${({perm_z}(x))}_d$ stands for $d$-th entry of the vector ${({perm_z}(x))}$.}  \eqref{tp1} follows from induction assumption for the nodes with depth less than $d$, \eqref{tp3} is a result of {{}i.i.d. assumption \eqref{pyx_2}, i.e., $\mathbb{P}(y\mid x)=\prod_{s=1}^S  p(y_s\mid x_s)$ and the defnition of $H_z^{\mathcal{A}}(x)$ in \eqref{ujvx}}. $(\eqref{famm1}-\eqref{fam2})$ follow similarly.

\subsection{Proof of Lemma \ref{lem_a1}.} 
Using $(\eqref{fam1}-\eqref{fam2})$, constraints  $(\eqref{family1}-\eqref{family2})$ hold for $\alpha=\alpha(G)$, $\beta=\beta(G)$, $\gamma^{\mathcal{A}}=\gamma^{\mathcal{A}}(G)$ and $\gamma^{\mathcal{B}}=\gamma^{\mathcal{B}}(G)$. Since no two buckets are ancestor/descendant of each other, we have
\begin{eqnarray}
{\mid H^{\mathcal{A}}(x)\cap H^{\mathcal{B}}(y)\mid} \le 1.
\end{eqnarray}
Therefore, \eqref{family5} holds. This completes the proof that  $H_z^{\mathcal{A}}(x)$ and $H_z^{\mathcal{B}}(y)$ defined in \eqref{ujvx} and \eqref{ujvy} are\\ $(\alpha(G),\beta(G),\gamma^{\mathcal{A}}(G),\gamma^{\mathcal{B}}(G))$-sensitive.

\section{Deriving $\mu^*$, $\nu^*$, $\eta^*$, $\lambda^*$, $p_0$, $q_0$ and $\delta$ for $\mathbb{P}$, $M$ and $N$}\label{app_numerical}
In this section, an algorithm for deriving $\mu^*$, $\nu^*$, $\eta^*$, $\lambda^*$, $p_0$, $q_0$ and $\delta$ for $\mathbb{P}$, $M$ and $N$ for the probability distribution $\mathbb{P}$ and $\delta$ is presented for a given probability distribution $\mathbb{P}$.
\begin{algorithm}[H]
   \caption{Deriving $\mu^*$, $\nu^*$, $\eta^*$, $\lambda^*$, $p_0$, $q_0$ and $\delta$ for $\mathbb{P}$, $M$ and $N$}
   \label{numerical_lambda}
\begin{algorithmic}
   \STATE {\bfseries Inputs:}  The probability distribution $\mathbb{P}$, ${list}_\mu$, ${list}_\nu$, ${list}_\eta$, threshold $T$, $M$ and $N$.
\STATE {\bfseries Outputs:}  $\mu^*$, $\nu^*$, $\eta^*$, $\lambda^*$, $p_0$, $q_0$ and $\delta$.
\STATE{\bfseries Procedure} 
\STATE~~~~$\delta\leftarrow\frac{\log M}{\log N}$
\STATE~~~~$\lambda^*=0$
\STATE~~~~{\bf For} $\mu\in{list}_\mu$ ~~~$\#$ For some set of ${list}_\mu$, e.g., $\{0,0.1,0.2,\cdots,10\}$.
\STATE~~~~~~~~{\bf For} $\nu\in{list}_\nu$~~~$\#$ For some set of ${list}_\nu$, e.g., $\{0,0.1,0.2,\cdots,10\}$.
\STATE~~~~~~~~~~~~{\bf For} $\eta\in{list}_\eta$~~~$\#$ For some set of ${list}_\eta$, e.g., $\{0,0.1,0.2,\cdots,10\}$.
\STATE~~~~~~~~~~~~~~~~{\bf If} $|\sum_{i,j}p_{ij} ^ {1+\mu+\nu-\eta} {(p_{i}^{\mathcal{A}})} ^ {-\mu}  {(p_j^{\mathcal{B}})} ^ {-\nu}-1|\le T$ and $\frac{\max(1,\delta)+\mu+\delta\nu}{1+\mu+\nu-\eta}> \lambda^*$. $\#$ For some small threshold $T$, e.g., $0.001$
\STATE~~~~~~~~~~~~~~~~~~~~$\mu^*\leftarrow\mu$, $\nu^*\leftarrow\nu$, $\eta^*\leftarrow\eta$, $\lambda^*\leftarrow\frac{\max(1,\delta)+\mu+\delta\nu}{1+\mu+\nu-\eta}$
\STATE~~~~\#$p_0$, $q_0$ are computed as follows.
\STATE~~~~$p_0\leftarrow1$, $q_0\leftarrow1$, $p_0^{\mathcal{A}}\leftarrow1$, $p_0^{\mathcal{B}}\leftarrow1$
\STATE~~~~{\bf For} $a_i\in\mathcal{A}$
\STATE~~~~~~~~{\bf For} $b_j\in\mathcal{B}$
\STATE~~~~~~~~~~~~$p_0\leftarrow p_0p_{ij}$
\STATE~~~~~~~~~~~~$q_0\leftarrow q_0q_{ij}$
\STATE~~~~~~~~~~~~$p_0^{\mathcal{A}}\leftarrow p_0^{\mathcal{A}}p_{i}^{\mathcal{A}}$
\STATE~~~~~~~~~~~~$p_0^{\mathcal{B}}\leftarrow p_0^{\mathcal{B}}p_j^{\mathcal{B}}$
\STATE~~~~$q_0\leftarrow\min(q_0,p_0^{\mathcal{A}},p_0^{\mathcal{B}})$.
\end{algorithmic}
\end{algorithm}
\begin{remark} In Algorithm \ref{numerical_lambda}, the parameters $\mu^*$, $\nu^*$, $\eta^*$ and $\lambda^*$ could be derived from newton method too. 
\end{remark}

\section{Proof of Theorem \ref{theoremA}}\label{rigor_theoremA}
In order to prove Theorem \ref{theoremA}, we first state the following two lemmas.
\begin{lemma}\label{lem_con_f}
The function $f(\theta,\theta_1,\theta_2,\theta_3)={\theta} ^ {1+\rho_1+\rho_2+\rho_3} {\theta_1} ^{-\rho_1} {\theta_2}^ {-\rho_2} {\theta_3} ^{-\rho_3}$ is a convex function on the region $(\theta,\theta_1,\theta_2,\theta_3)\in \mathbb{R}^{4+}$ where $(\rho_1,\rho_2,\rho_3)\in \mathbb{R}^{3+}$. \footnote{For any natural number $n$, $\mathbb{R}^{n+}$ denotes as the set of all $n$-tuples non-negative real numbers.}
\end{lemma}
\begin{lemma}\label{induction_AB} $\sum_{v \in V_{Buckets}(G)}{\big(\Phi(v)\big)}^{1+\mu+\nu-\eta}{\big({\Psi}^{\mathcal{A}}(v)\big)}^ {-\mu +\eta}{\big({\Psi}^{\mathcal{B}}(v)\big)}^ {-\nu +\eta}{\big(\Psi(v)\big)}^ {-\eta } \le 1$ for any  $(\mu ,\nu ,\eta)\in \mathcal{I}$.
\end{lemma}
Proof of Lemma \ref{lem_con_f} and Lemma \ref{induction_AB}, are relegated to Appendices \ref{Convexity} and \ref{tt+e}, respectively. Consider $(\mu ^ *,\nu ^ *,\eta^*)$ that satisfy \eqref{defmunu}. For any decision tree satisfying $(\eqref{order1}-\eqref{order3})$, we have: 
\begin{eqnarray}
&&{(\frac{\sum_{v \in V_{Buckets}(G)}{\Phi(v)}}{|V_{Buckets}(G)|})}^ {1+\mu^*+\nu^*-\eta^*} {(\frac{\sum_{v \in V_{Buckets}(G)}{{\Psi}^{\mathcal{A}}(v)}}{|V_{Buckets}(G)|})}^{-\mu ^ *+\eta^*}\nonumber\\
&&\times {(\frac{\sum_{v \in V_{Buckets}(G)}{{\Psi}^{\mathcal{B}}(v)}}{|V_{Buckets}(G)|})}^{-\nu ^ *+\eta^*} {(\frac{\sum_{v \in V_{Buckets}(G)}{\Psi(v)}}{|V_{Buckets}(G)|})}^{-\eta ^ *}  \nonumber\\
&\leq& \frac{ \sum_{v \in V_{Buckets}(G)}{\big(\Phi(v)\big)}^{1+\mu^*+\nu^*-\eta^*}{\big({\Psi}^{\mathcal{A}}(v)\big)}^ {-\mu ^ *+\eta^*}{\big({\Psi}^{\mathcal{B}}(v)\big)}^ {-\nu ^ *+\eta^*}{\big(\Psi(v)\big)}^ {-\eta ^ *}}{|V_{Buckets}(G)|}\label{nnn1+} \\
&\le& \frac{1}{|V_{Buckets}(G)|}\label{nnn1},
\end{eqnarray}
where \eqref{nnn1+} holds to the convexity of $f(\theta,\theta_1,\theta_2,\theta_3)={\theta} ^ {1+\rho_1+\rho_2+\rho_3} {\theta_1} ^{-\rho_1} {\theta_2}^ {-\rho_2} {\theta_3} ^{-\rho_3}$ in Lemma  \ref{lem_con_f}  and \eqref{nnn1} follows from Lemma \ref{induction_AB}. Therefore, we have 
\begin{eqnarray}
&&{\big({\sum_{v \in V_{Buckets}(G)}{\Phi(v)}}\big)}^ {1+\mu^*+\nu^*-\eta^*} {\big({\sum_{v \in V_{Buckets}(G)}{{\Psi}^{\mathcal{A}}(v)}}\big)}^{-\mu ^ *+\eta^*} \nonumber\\
&&\times{\big({\sum_{v \in V_{Buckets}(G)}{{\Psi}^{\mathcal{B}}(v)}}\big)}^{-\nu ^ *+\eta^*} {\big({\sum_{v \in V_{Buckets}(G)}{\Psi(v)}}\big)}^{-\eta ^ *} \nonumber\\
& \leq& 1\label{hhh+1}.
\end{eqnarray}
On the other hand, using \eqref{nnn1} and the definitions of $\alpha(G)$, $\beta(G)$, $\gamma^{\mathcal{A}}(G)$ and $\gamma^{\mathcal{B}}(G)$ in $(\eqref{alphdef}-\eqref{gammydef})$, we have
\begin{eqnarray}
{\big(\alpha(G)\big)}^ {1+\mu^*+\nu^*-\eta^*} {\big(\gamma^{\mathcal{A}}(G)\big)}^{-\mu ^ *+\eta^*} {\big(\gamma^{\mathcal{B}}(G)\big)}^{-\nu ^ *+\eta^*} {\big(\beta(G)\big)}^{-\eta ^ *}
& \leq& 1\label{hhh+1+1}.
\end{eqnarray}
Therefore, from $(\eqref{order1}-\eqref{order3})$ and \eqref{hhh+1+1} we have
%
%
%
\begin{eqnarray}
&&1\nonumber\\
&\ge&{\big(\alpha(G)\big)}^ {1+\mu^*+\nu^*-\eta^*} {\big(\gamma^{\mathcal{A}}(G)\big)}^{-\mu ^ *+\eta^*} {\big(\gamma^{\mathcal{B}}(G)\big)}^{-\nu ^ *+\eta^*} {\big(\beta(G)\big)}^{-\eta ^ *} \nonumber\\
&=& \alpha(G){\left(\frac{\alpha(G)}{{\gamma^{\mathcal{A}}(G)}}\right)}^{\mu^*-\eta^*} {\left(\frac{\alpha(G)}{{\gamma^{\mathcal{B}}(G)}}\right)}^{\nu^*-\eta^*} {\left(\frac{\alpha(G)}{\beta(G)}\right)}^{\eta^*}\\
&\ge& \left(N^{\max(1,\delta)-\lambda}\right){\left(N^{1-\lambda}\right)}^{\mu^*-\eta^*} {\left(N^{\delta-\lambda}\right)}^{\nu^*-\eta^*} {\left(N^{1+\delta-\lambda}\right)}^{\eta^*}\\
&=&N^{\max(1,\delta)+\mu^*+\delta\nu^*-(1+\mu^*+\nu^*-\eta^*)\lambda}.
\end{eqnarray}
Therefore, we conclude Theorem \ref{theoremA} as follows
\begin{eqnarray}
\lambda&\ge&\lambda^*=\frac{\max(1,\delta)+\mu^*+\delta\nu^*}{1+\mu^*+\nu^*-\eta^*}.
\end{eqnarray}
\subsection{Proof of Lemma \ref{lem_con_f}}\label{Convexity}
The Hessian matrix for $f(\theta,\theta_1,\theta_2,\theta_3)$ is represented as 
\begin{eqnarray}
&&H(\theta,\theta_1,\theta_2,\theta_3)\nonumber\\
&=&f(\theta,\theta_1,\theta_2,\theta_3)\nonumber\\
&&\times\begin{bmatrix}
\frac{(1+\rho_1+\rho_2+\rho_3)(\rho_1+\rho_2+\rho_3)}{\theta^2}&\frac{-\rho_1(1+\rho_1+\rho_2+\rho_3)}{\theta\theta_1}&\frac{-\rho_2(1+\rho_1+\rho_2+\rho_3)}{\theta\theta_2}&\frac{-\rho_3(1+\rho_1+\rho_2+\rho_3)}{\theta\theta_3}\\
\frac{-\rho_1(1+\rho_1+\rho_2+\rho_3)}{\theta\theta_1}&\frac{\rho_1(\rho_1+1)}{\theta_1^2}&\frac{\rho_1\rho_2}{\theta_1\theta_2}&\frac{\rho_1\rho_3}{\theta_1\theta_3}\\
\frac{-\rho_2(1+\rho_1+\rho_2+\rho_3)}{\theta\theta_2}&\frac{\rho_1\rho_2}{\theta_1\theta_2}&\frac{\rho_2(\rho_2+1)}{\theta_2^2}&\frac{\rho_2\rho_3}{\theta_2\theta_3}\\
\frac{-\rho_3(1+\rho_1+\rho_2+\rho_3)}{\theta\theta_3}&\frac{\rho_1\rho_3}{\theta_1\theta_3}&\frac{\rho_2\rho_3}{\theta_2\theta_3}&\frac{\rho_3(\rho_3+1)}{\theta_3^2}
\end{bmatrix}\nonumber\\
&=&f(\theta,\theta_1,\theta_2,\theta_3)\nonumber\\
&&\times\begin{bmatrix}
W^2+\frac{\rho_1+\rho_2+\rho_3}{\theta^2}&WW_1-\frac{\rho_1}{\theta\theta_1}&WW_2-\frac{\rho_2}{\theta\theta_2}&WW_3-\frac{\rho_3}{\theta\theta_3}\\
W_1W-\frac{\rho_1}{\theta\theta_1}&W_1^2+\frac{\rho_1}{\theta_1^2}&W_1W_2&W_1W_3\\
W_2W-\frac{\rho_2}{\theta\theta_2}&W_2W_1&W_2^2+\frac{\rho_2}{\theta_2^2}&W_2W_3\\
W_3W-\frac{\rho_3}{\theta\theta_3}&W_3W_1&W_3W_2&W_3^2+\frac{\rho_3}{\theta_3^2}
\end{bmatrix},
\end{eqnarray}
where $W=\frac{\rho_1+\rho_2+\rho_3}{\theta}$, $W_i=\frac{\rho_i}{\theta_i}$ for any $i\in\{1,2,3\}$. In order to show that the function $f(\theta,\theta_1,\theta_2,\theta_3)$ is a convex function it is necessary and sufficient to prove that $H(\theta,\theta_1,\theta_2,\theta_3)$ is positive semidefinite on $\mathbb{R}^{4+}$ On the other hand, for positive semidefinite matrices we have
\begin{enumerate}
\item{}For any non-negative scalar $a$ and positive semidefinite matrix $M$, $aM$ is positive semidefinite.
\item{}For positive semidefinite matrices $M_1$ and $M_2$,  $M_1+M_2$ is positive semidefinite.
\end{enumerate}
As $f(\theta,\theta_1,\theta_2,\theta_3)>0$ for any $\theta,\theta_1,\theta_2,\theta_3$, it is sufficient to prove that $\frac{H(\theta,\theta_1,\theta_2,\theta_3)}{f(\theta,\theta_1,\theta_2,\theta_3)}$ is positive semidefinite. Define, $M_1=\begin{bmatrix}
W^2&WW_1&WW_2&WW_3\\
W_1W&W_1^2&W_1W_2&W_1W_3\\
W_2W&W_2W_1&W_2^2&W_2W_3\\
W_3W&W_3W_1&W_3W_2&W_3^2
\end{bmatrix}$ and $M_2=\begin{bmatrix}
\frac{\rho_1+\rho_2+\rho_3}{\theta^2}&-\frac{\rho_1}{\theta\theta_1}&-\frac{\rho_2}{\theta\theta_2}&-\frac{\rho_3}{\theta\theta_3}\\
-\frac{\rho_1}{\theta\theta_1}&\frac{\rho_1}{\theta_1^2}&0&0\\
-\frac{\rho_2}{\theta\theta_2}&0&\frac{\rho_2}{\theta_2^2}&0\\
-\frac{\rho_3}{\theta\theta_3}&0&0&\frac{\rho_3}{\theta_3^2}
\end{bmatrix}$. {\color{black}Therefore, we have $M_1+M_2=\frac{H(\theta,\theta_1,\theta_2,\theta_3)}{f(\theta,\theta_1,\theta_2,\theta_3)}$. In order to prove that $f(\theta,\theta_1,\theta_2,\theta_3)$ is positive semidefinite, it is sufficient to prove that $M_1$ and $M_2$ are positive semidefinites.} The matrices $M_1$ and $M_2$ are positive semidefinite as for any non-zero vector $z=\begin{bmatrix}a&b&c&d\end{bmatrix}$, we have $zM_1z^T\ge0$ and $zM_2z^T\ge0$, i.e.,
\begin{eqnarray}
&&\begin{bmatrix}a&b&c&d\end{bmatrix}\begin{bmatrix}
W^2&WW_1&WW_2&WW_3\\
W_1W&W_1^2&W_1W_2&W_1W_3\\
W_2W&W_2W_1&W_2^2&W_2W_3\\
W_3W&W_3W_1&W_3W_2&W_3^2
\end{bmatrix}\begin{bmatrix}a\\b\\c\\d\end{bmatrix}\nonumber\\
&=&(Wa+W_1b+W_2c+W_3d)^2\\
&\ge&0,\\
&&\begin{bmatrix}a&b&c&d\end{bmatrix}\begin{bmatrix}
\frac{\rho_1+\rho_2+\rho_3}{\theta^2}&-\frac{\rho_1}{\theta\theta_1}&-\frac{\rho_2}{\theta\theta_2}&\frac{\rho_3}{\theta\theta_3}\\
-\frac{\rho_1}{\theta\theta_1}&\frac{\rho_1}{\theta_1^2}&0&0\\
-\frac{\rho_2}{\theta\theta_2}&0&\frac{\rho_2}{\theta_2^2}&0\\
-\frac{\rho_3}{\theta\theta_3}&0&0&\frac{\rho_3}{\theta_3^2}
\end{bmatrix}\begin{bmatrix}a\\b\\c\\d\end{bmatrix}\nonumber\\
&=&\rho_1{\big(\frac{a}{\theta}-\frac{b}{\theta_1}\big)}^2+\rho_2{\big(\frac{a}{\theta}-\frac{c}{\theta_2}\big)}^2\nonumber\\
&&+\rho_3{\big(\frac{a}{\theta}-\frac{d}{\theta_3}\big)}^2\\
&\ge&0\label{semi+},
\end{eqnarray} 
where \eqref{semi+} is concluded as $\rho_1,\rho_2,\rho_3\ge0$.

\subsection{Proof of Lemma \ref{induction_AB}}\label{tt+e}
First of all, note that $\Psi(v)={\Psi}^{\mathcal{A}}(v){\Psi}^{\mathcal{B}}(v)$. Let us define
\begin{eqnarray}
D(v)=\sum_{v \in V_{Buckets}(G)}\Phi(v) ^ {1+\mu +\nu -\eta }{\big({\Psi}^{\mathcal{A}}(v)\big)}^ {-\mu }{\big({\Psi}^{\mathcal{B}}(v)\big)}^ {-\nu }.
\end{eqnarray}
We show
\begin{eqnarray}
\sum_{v\in V_{Buckets}(G)}D(v)\le 1 \label{ttyl1},
\end{eqnarray}
by induction on the number of nodes in the tree. If the tree has only one node, i.e., {\it root}, then \eqref{ttyl1} is holds as $\Phi({root})=1$, ${\Psi}^{\mathcal{A}}({root})=1$ and ${\Psi}^{\mathcal{B}}({root})=1$  from the definition of $\Phi(v)$, ${\Psi}^{\mathcal{A}}(v)$ and ${\Psi}^{\mathcal{B}}(v)$ in $(\eqref{def_a}-\eqref{def_a3})$. Assume that \eqref{ttyl1} holds for any decision tree with $|G|<Z$. Our goal is to prove that \eqref{ttyl1} holds for a decision tree with $|G|=Z$.  Assume $v_{11}$ is the node with maximum length in $G$ and consider a tree $G'$ constructed by removing $v_{11}$ and all its siblings $v_{ij},1\le i\le k,1\le j\le l$ belonging to the same parent $v$.  In other words, for the tree $G'$ we have\footnote{Note that,  $V_b(G')$ satisfies bucket-list property, e.g., there is no bucket in the tree that is ancestor of another bucket.}
\begin{eqnarray}
V(G')&=&V(G)/\{v_{11},\cdots,v_{kl}\},\\
V_{Buckets}(G')&=&\left(V_{Buckets}(G)\cup\{v\}\right)/\{v_{11},\cdots,v_{kl}\}\label{manbi}.
\end{eqnarray}
\eqref{manbi} is true as for the gragh $G'$, the node $v$ in now a leaf node while the nodes $v_{11},\cdots,v_{kl}$ are removed. Then, we have
\begin{eqnarray}
&&\sum_{v\in V_{Buckets}(G)}D(v)\nonumber\\
&\le&\sum_{v\in V_{Buckets}(G')}D(v)-\Phi(v) ^ {1+\mu +\nu -\eta }{\big({\Psi}^{\mathcal{A}}(v)\big)}^ {-\mu }{\big({\Psi}^{\mathcal{B}}(v)\big)}^ {-\nu }\nonumber\\
&&+\sum_{i,j}\Phi(v_{ij}) ^ {1+\mu +\nu -\eta }{{\Psi}^{\mathcal{A}}(v_{ij})}^ {-\mu }{{\Psi}^{\mathcal{B}}(v_{ij})}^ {-\nu }\label{ttyl2}\\
&=&\sum_{\in V_{Buckets}(G')}D(v)-\Phi(v) ^ {1+\mu +\nu -\eta }{\big({\Psi}^{\mathcal{A}}(v)\big)}^ {-\mu }{\big({\Psi}^{\mathcal{B}}(v)\big)}^ {-\nu }\nonumber\\
&&+\sum_{i,j}\left(\Phi(v) ^ {1+\mu +\nu -\eta }{\big({\Psi}^{\mathcal{A}}(v)\big)}^ {-\mu +\eta }{\big({\Psi}^{\mathcal{B}}(v)\big)}^ {-\nu }\right.\nonumber\\
&&\left.\times{p_{ij}} ^ {1+\mu +\nu -\eta }{(p_{i}^{\mathcal{A}})}^ {-\mu }{(p_{i}^{\mathcal{B}})}^ {-\nu }\right)\label{ttyl3}\\
&=&\sum_{v\in V_{Buckets}(G')}D(v)-\Phi(v) ^ {1+\mu +\nu -\eta }{\big({\Psi}^{\mathcal{A}}(v)\big)}^ {-\mu }{\big({\Psi}^{\mathcal{B}}(v)\big)}^ {-\nu }\nonumber\\
&&\times\left(1-\sum_{i,j}{p_{ij}} ^ {1+\mu +\nu -\eta }{(p_i^{\mathcal{A}})}^ {-\mu }{(p_j^{\mathcal{B}})}^ {-\nu }\right)\\
&=&\sum_{v\in V_{Buckets}(G')}D(v)\label{ttyl4}\\
&=&1,
\end{eqnarray}
where \eqref{ttyl2} holds from the definition of tree $G'$, \eqref{ttyl3} follows from the recursive definition of $\Phi(v)$, ${\Psi}^{\mathcal{A}}(v)$ and ${\Psi}^{\mathcal{B}}(v)$ in $(\eqref{def_a}-\eqref{def_a3})$ and \eqref{ttyl4} holds, note that from the definition of $\mu$, $\nu$ and $\eta$ in \eqref{def_mathI}, i.e., 
\begin{eqnarray}
\sum_{i,j}{p_{ij}} ^ {1+\mu +\nu -\eta }{(p_i^{\mathcal{A}})}^ {-\mu }{(p_j^{\mathcal{B}})}^ {-\nu }=1\label{b+b}.
\end{eqnarray} 
Therefore, we conclude that
\begin{eqnarray}
\sum_{v\in V_{Buckets}(G)}D(v)&\le&1.
\end{eqnarray}
Note that, the inequality \eqref{ttyl2} becomes an equality only in cases where the tree is homogeneous and none of the children are pruned.

\section{Proof of Theorem \ref{theoremB}}\label{rigor_proof_B}
In order to prove Theorem \ref{theoremB}, we first present the following lemma.
\begin{lemma}\label{lem_optim} Given a fixed  $N\in\mathbb{N}$ and probability distribution $\mathbb{P}$, consider the following region $R_{\lambda}$
\begin{eqnarray}
\mathcal{R}(\lambda,r_{ij},n)=\left\{\lambda,r_{ij},n\mbox{~s.t.~}~\lambda\ge0,
\sum_{i,j}{r_{ij}} \right.&=& 1, r_{ij} \geq 0,r_{ij} \in \mathbb{R}^+,n \in \mathbb{R}^+,\label{l0}\\
\sum_{i,j}{r_{ij}\log p_{ij}}-\sum_{i,j}{r_{ij}\log r_{ij}} &\ge& \frac{(\max(1,\delta)-\lambda) \log N}{n},\label{l00}\\
\sum_{i,j}{r_{ij}\log p_{ij}}-\sum_{i,j}{r_{ij}\log p_{i}^{\mathcal{A}}} &\ge& \frac{(1-\lambda) \log N}{n}\label{l00},\\
\sum_{i,j}{r_{ij}\log p_{ij}}-\sum_{i,j}{r_{ij}\log p_j^{\mathcal{B}}} &\ge& \frac{(\delta-\lambda) \log N}{n},\label{l00}\\
\sum_{i,j}{r_{ij}\log p_{ij}}-\sum_{i,j}{r_{ij}\log q_{ij}} &\geq&\left. \frac{(1+\delta-\lambda) \log N}{n} \right\}\label{l---1}
\end{eqnarray}
\footnote{Recall that, for simplicity we use the notation $\sum_{i,j}$ and $\prod_{i,j}$ instead of $\sum_{1\le i\le k,1\le j\le l}$ and $\prod_{1\le i\le k,1\le j\le l}$, respectively.} Then, $(\lambda^*,r_{ij}^*,n^*)$ defined in Definition \ref{deflambda} is a member of $\mathcal{R}(\lambda,r_{ij},n)$.
\end{lemma}
The proof of Lemma \ref{lem_optim} is relegated to Appendix \ref{app_optimization_problem}. Let us prove that the following tree construction steps in Algorithm \ref{algorithm_Tree} result in a tree that satisfies $(\eqref{order1}-\eqref{order3})$.

\begin{eqnarray}
\left\{\begin{matrix}\frac{\Phi(v)}{\Psi(v)} \ge {N ^ {1+\delta-\lambda^*}}{p_0q_0}&: {accept~bucket},~~~~~~~~~~~~~~~~~~~\\
\frac{\Phi(v)}{{\Psi}^{\mathcal{A}}(v)}\le N^{1-\lambda^*}{p_0q_0}&: prune,~~~~~~~~~~~~~~~~~~~~~~~~~~~~~\\
\frac{\Phi(v)}{{\Psi}^{\mathcal{B}}(v)}\le N^{\delta-\lambda^*}{p_0q_0}&: prune,~~~~~~~~~~~~~~~~~~~~~~~~~~~~~\\
otherwise&: {branch~into~the~kl~children.} \end{matrix}\right.{}
\end{eqnarray}
Consider the set of $r^*_{ij} =p_{ij} ^ {1+\mu^*+\nu^*-\eta^*} {(p_{i}^{\mathcal{A}})} ^ {-\mu^*}  {(p_j^{\mathcal{B}})} ^ {-\nu^*}$ and $n^* =\frac{(\max(1,\delta)-\lambda^*) \log N}{\sum{r_{ij}^*\log \frac{p_{ij}}{r_{ij}^*}} }$. Note that we assume $p_{ij}$ and $q_{ij}$ are non-zero\footnote{Note that in the cases where $q_{ij}$ is zero, then from the definition of $q_{ij}$, $p_{ij}$ would also be equal to zero. Therefore, we will ignore those branches during the tree construction.}. Consider $n_{ij} = \lceil n^*r_{ij}^* \rceil$ if $r^*_{ij} > \frac{1}{2}$ and $n_{ij} = \lfloor n^*r^*_{ij} \rfloor$ if $r^*_{ij} \le \frac{1}{2}$. Therefore, we have $n^*-kl<\sum_{ij}n_{ij}\le n^*$. For any $v\in V(G)$, define the set ${S}_{ij}(v)$ as follows
\begin{eqnarray}
{S}_{ij}(v)\define \{s\mid 1\le s\le depth(v),{Seq^{\mathcal{A}}_s}(v)=a_i \& {Seq^{\mathcal{B}}_s}(v)=b_j\},
\end{eqnarray}
where $depth(v)$ is the depth of node $v$ in the tree, $Seq^{\mathcal{A}}_s(v)$ and $Seq^{\mathcal{B}}_s(v)$ stand for position $s$ in the strings $Seq^{\mathcal{A}}(v)$ and $Seq^{\mathcal{B}}$(v), respectively. Now, consider a node $v^*$ in the graph that satisfies the following constraints:
\begin{eqnarray}
|{S}_{ij}(v)|&=&n_{ij}, \forall 1\le i\le k,1\le j\le l\label{3star}.
\end{eqnarray}
The number of nodes $v$ that satisfy this constraint {\color{black}is} $\binom{n^*}{n_{11},\cdots,n_{kl}}$. Moreover, define
\begin{eqnarray}
|V_{n_{11},\cdots,n_{kl}}|&\define&\{v\in V(G)\mid |{S}_{ij}(v)|=n_{ij}, \forall 1\le i\le k,1\le j\le l\}.
\end{eqnarray}
\subsection{Node $v^*$ or one of its ancestors is designated as a bucket by Algorithm \ref{algorithm_Tree}}
 Here, we prove that {\color{black}the node $v$ or one of its ancestors is designated as a bucket} by Algorithm \ref{algorithm_Tree}. In order to show this, we need to prove that:
\begin{eqnarray}
\frac{\Phi(v^*)}{\Psi(v^*)} &\ge& e ^ {\sum n^*r^*_{ij}\log p_{ij}}e ^ {-\sum n^*r^*_{ij}\log q_{ij}}{p_{0}}{q_{0}}  \geq N^{1+\delta - \lambda^*}{p_{0}}{q_{0}}\label{qmmm1}\label{i-i=1},\\
\frac{\Phi(v^*)}{{\Psi}^{\mathcal{A}}(v^*)}  &\ge& e ^ {\sum n^*r^*_{ij}\log p_{ij}}e ^ {-\sum n^*r^*_{ij}\log p_i^{\mathcal{A}}}{p_{0}}{q_{0}}  \geq N^{1 - \lambda^*}{p_{0}}{q_{0}}, \\
\frac{\Phi(v^*)}{{\Psi}^{\mathcal{B}}(v^*)} &\ge& e ^ {\sum n^*r^*_{ij}\log p_{ij}}e ^ {-\sum n^*r^*_{ij}\log p_j^{\mathcal{B}}}{p_{0}}{q_{0}}  \geq N^{\delta - \lambda^*}{p_{0}}{q_{0}}, \label{qttttt2}\label{iii2}
\end{eqnarray}
where   $p_{0}$ and $q_{0}$ are defined as $\prod_{i,j}p_{ij}$ and $\min(\prod_{i,j}q_{ij},\prod_{i}{(p_i^{\mathcal{A}})}^{l},\prod_j{(p_j^{\mathcal{B}})}^{k})$. Note that, $\Phi(v^*)$, $\Psi(v^*)$, ${\Psi}^{\mathcal{A}}(v^*)$ and ${\Psi}^{\mathcal{B}}(v^*)$ {\color{black}are computed as follows}
\begin{eqnarray}
\Phi(v^*)&=& \prod_{i,j} p_{ij}^{n_{ij}} = e ^ {\sum_{i,j} n_{ij}\log p_{ij}}\ge  e ^ {\sum_{i,j} (n^*r_{ij}^*+1)\log p_{ij}} \nonumber\\
&\geq& e ^ {\sum_{i,j} n^*r_{ij}^*\log p_{ij}}(\prod_{i,j}p_{ij}) =e ^ {\sum_{i,j} n^*r^*_{ij}\log p_{ij}}p_{0}\label{..1},\\
\Psi(v^*) &=& \prod_{i,j} q_{ij}^{n_{ij}} = e ^ {\sum_{i,j} n_{ij}\log q_{ij}} \nonumber\\
&\le&  e ^ {\sum_{i,j} (n^*r_{ij}^*-1)\log q_{ij}}  \leq \frac{e ^ {\sum_{i,j} n^*r_{ij}^*\log q_{ij}}}{\prod_{i,j}q_{ij}} = \frac{e ^ {\sum_{i,j} n^*r_{ij}^*\log q_{ij}}}{q_{0}}\label{..2},\\
{\Psi}^{\mathcal{A}}(v^*) &=& \prod_{i,j} {(p_i^{\mathcal{A}})}^{n_{ij}} = e ^ {\sum_{i,j} n_{ij}\log p_i^{\mathcal{A}}} \nonumber\\
&\le&  e ^ {\sum_{i,j} (n^*r_{ij}^*-1)\log {(p_i^{\mathcal{A}})}}  \leq \frac{e ^ {\sum_{i,j} n^*r_{ij}^*\log p_i^{\mathcal{A}}}}{\prod_{i,j}p_i^{\mathcal{A}}} = \frac{e ^ {\sum_{i,j} n^*r_{ij}^*\log p_i^{\mathcal{A}}}}{q_{0}}\label{..3},\\
{\Psi}^{\mathcal{B}}(v^*) &=& \prod_{i,j} {(p_j^{\mathcal{B}})}^{n_{ij}} = e ^ {\sum_{i,j} n_{ij}\log p_j^{\mathcal{B}}} \nonumber\\
&\le&  e ^ {\sum_{i,j} (n^*r_{ij}^*-1)\log {p_j^{\mathcal{B}}}}  \leq \frac{e ^ {\sum_{i,j} n^*r_{ij}^*\log p_j^{\mathcal{B}}}}{\prod_{i,j}{p_j^{\mathcal{B}}}} = \frac{e ^ {\sum_{i,j} n^*r_{ij}^*\log p_j^{\mathcal{B}}}}{q_{0}}\label{..4}.
\end{eqnarray}
Therefore, from Lemma \ref{lem_optim} and $(\eqref{..1}-\eqref{..4})$ we conclude $(\eqref{qmmm1}-\eqref{qttttt2})$.  This means $v^*$  or {\color{black}one of its ancestors is an accepted bucket}. 

\subsection{Proof of bounds (\eqref{order111}-\eqref{order3})}
First of all, we derive a lower bound on $\alpha(G)$ as follows.
\begin{eqnarray}
\alpha(G)&=&\sum_{v\in V_{Buckets}(G)}\Phi(v)\nonumber\\
&\ge&\sum_{v\in V_{n_{11},\cdots,n_{kl}}}\Phi(v)\\
&\ge&|V_{n_{11},\cdots,n_{kl}}|\Phi(v) \\
&\ge&|V_{n_{11},\cdots,n_{kl}}|e ^ {\sum_{i,j} n^*r^*_{ij}\log p_{ij}}p_{0}\label{hj1},
\end{eqnarray}
where $V_{n_{11},\cdots,n_{kl}}$ is the set of nodes that satisfies \eqref{3star}. $|V_{n_{11},\cdots,n_{kl}}|$ is lower bounded as
\begin{eqnarray}
|V_{n_{11},\cdots,n_{kl}}|&=&\binom{n^*}{n_{11},\cdots,n_{kl}}\nonumber\\
&\ge& \frac{n^*!}{(kl)!\prod_{i,j}{n_{ij}!}} \geq \frac{(\frac{n^*}{e})^n\sqrt{2\pi n^*}}{(kl)!\prod_{i,j} (\frac{n_{ij}}{e})^{n_{ij}}\sqrt{2\pi n_{ij}}  e} \label{kkk1}\\
&\ge& \prod_{i,j}{(\frac{n_{ij}}{n^*}) ^ {-n_{ij}}} {(n^*)} ^ {\frac{1 - kl}{2}}  \frac{ (2\pi) ^ {\frac{1 - kl}{2}}e ^ {-kl}}{(kl)!}\\
&=&c e ^ {-\sum_{i,j} n_{ij}\log (\frac{n_{ij}}{n^*})}   {(n^*)} ^ {\frac{1 - kl}{2}} \\
&\geq&c e ^ {-n^*\sum_{i,j} r^*_{ij}\log r^*_{ij}},  \label{in1}
\end{eqnarray}
for some constant $c=\frac{ (2\pi) ^ {\frac{1 - kl}{2}}e ^ {-kl}}{(kl)!}   {(n^*)} ^ {\frac{1 - kl}{2}}$ depending on $n$, $k$ and $l$. \eqref{kkk1} is true as for any natural number $m$ we have $\sqrt{2\pi m}{(\frac{m}{e})}^m\le m!< \sqrt{2\pi m}{(\frac{m}{e})}^me$. \eqref{in1} follows as $a\log \frac{1}{a}$ is an increasing function for $0\le x\le 0.5$, and a decreasing function for $0.5\le x\le1$.  Therefore, from \eqref{hj1} and \eqref{in1}, ${\alpha(G)} \geq N^{\max(1,\delta)-\lambda^*}$ is concluded. Similarly, $(\eqref{order1}-\eqref{order3})$ are proved as follows.
\begin{eqnarray}
\frac{\alpha(G)}{\beta(G)}&=&\frac{\sum_{v\in V_{Buckets}(G)}\Phi(v)}{\sum_{v\in V_{Buckets}(G)}\Psi(v)}  \geq N^{1+\delta - \lambda^*}p_{0}q_{0}\label{mmm1}, \\
\frac{\alpha(G)}{\gamma^{\mathcal{A}}(G)}&=& \frac{\sum_{v\in V_{Buckets}(G)}\Phi(v)}{\sum_{v\in V_{Buckets}(G)}{\Psi}^{\mathcal{A}}(v)} \geq N^{1 - \lambda^*}p_{0}q_{0}, \\
\frac{\alpha(G)}{\gamma^{\mathcal{B}}(G)}&=&\frac{\sum_{v\in V_{Buckets}(G)}\Phi(v)}{\sum_{v\in V_{Buckets}(G)}{\Psi}^{\mathcal{B}}(v)}    \geq N^{\delta - \lambda^*}p_{0}q_{0} \label{ttttt2},
\end{eqnarray}
where \eqref{mmm1} and \eqref{ttttt2} is concluded from  $(\eqref{i-i=1}-\eqref{iii2})$ and the fact that $\frac{\sum_i{a_i}}{\sum_ib_i}\ge c$ if $\frac{a_i}{b_i}\ge c$ and $b_i>0$ for any $i$.

\subsection{Bounding number of nodes in the tree, i.e., \eqref{order1}}
The number of nodes in the decision tree defined in \eqref{ok}, is bounded by $O(N^{\lambda^*})$ using the following three lemmas.
\begin{lemma}\label{lem_numnodes_}
For any leaf node $v$ of the decision tree defined in \eqref{ok}, we have
\begin{eqnarray}
{\Phi(v)}&\ge &{N ^ {-\lambda^*}}\min_{i,j} p_{ij}{p_0q_0}\label{v1_5}.
\end{eqnarray}
\end{lemma}
\begin{lemma}\label{lemma_sum_av}For any tree $G$, the summation of $\Phi(v)$ over all the leaf nodes is equal to one, i.e., $\sum_{v\in V_{l}}\Phi(v)=1$.
\end{lemma}
\begin{lemma}\label{lemma_num_nodess} The number of nodes in the decision tree defined in \eqref{ok} is at most two times of the number of leaf nodes.
\end{lemma}
For proof of Lemmas \ref{lem_numnodes_}, \ref{lemma_sum_av} and \ref{lemma_num_nodess}, see Appendices \ref{proof_lem_numnodes_}, \ref{app_lemma_sum_av} and \ref{app_lemma_num_nodess}. Therefore, we have
\begin{eqnarray}
 && 1\nonumber\\
&=&\sum_{v \in V_1(G)}{\Phi(v)}\label{tttt7}\\
&\ge&\sum_{v \in V(G)}N^{-\lambda^*}\min_{i,j} p_{ij}{p_0q_0}\label{jjo}\\
&=&|V_l(G)|N^{-\lambda^*}\min_{i,j} p_{ij}{p_0q_0}\label{chie}\\
&\ge&\frac{|V(G)|}{2}N^{-\lambda^*}\min_{i,j} p_{ij}{p_0q_0}\label{chie2},
\end{eqnarray}
where \eqref{tttt7} follows from Lemma \ref{lemma_sum_av}, \eqref{jjo} is true from \eqref{v1_5} and \eqref{chie2} is concluded from Lemma \ref{lemma_num_nodess}. Therefore, we conclude that {{}$|V(G)|=O( N^{\lambda^*})$}.

\subsection{Proof of Lemma \ref{lem_optim}}\label{app_optimization_problem}
Consider the optimization problem of finding the member of  $(\lambda,r_{ij},n)\in \mathcal{R}(\lambda,r_{ij},n)$ with minimum $\lambda$. This optimization problem  is a convex optimization problem\footnote{Note that $f_1(n)=\frac{1}{n}$, $f_2(r)=r\log r$  and  $f_3(r)=ar$ are convex functions. Therefore, as the sum of the convex functions is a convex function, the optimization problem $(\eqref{l0}-\eqref{l---1})$ is in the form of convex optimization problem
\begin{eqnarray}
\underset{\lambda}{\mbox{Minimize~}}~f(x),&&\\
{\mbox{subject to~}}~\{g_i(x)&\le&0, i\in\{1,\cdots,m\},\\
h_j(x)&=&0, j\in\{1,\cdots,p\}\},
\end{eqnarray}
where $x\in\mathbb{R}^n$, $f(x)$ and $g_i(x)$ are convex functions and $h_j(x)$ is affine functions.}. Therefore, writing the KKT conditions, we have
\begin{eqnarray}
&&F(r_{ij},n,\lambda)\nonumber\\
&=&\lambda + \sum_{i,j}\mu_{1ij}( -r_{ij}) +\mu_2\big(\frac{(\max(1,\delta)-\lambda) \log N}{n}-\sum_{i,j}{r_{ij}\log p_{ij}}+\sum_{i,j}{r_{ij}\log r_{ij}}\big)\nonumber\\
&&+\mu_3\big(\frac{(1-\lambda) \log N}{n}-\sum_{i,j}{r_{ij}\log p_{ij}}+\sum_{i,j}{r_{ij}\log p_i^{\mathcal{A}}}\big)\nonumber\\
&&+\mu_4\big(\frac{(\delta-\lambda) \log N}{n}-\sum_{i,j}{r_{ij}\log p_{ij}}+\sum_{i,j}{r_{ij}\log p_j^{\mathcal{B}}}\big)\nonumber\\
&&+\mu_5\big( \frac{(1+\delta-\lambda) \log N}{n}-\sum_{i,j}{r_{ij}\log p_{ij}}+\sum_{i,j}{r_{ij}\log q_{ij}} \big)+ \mu_6\big(\sum_{i,j}{r_{ij}}-1\big),
\end{eqnarray}
where
\begin{eqnarray}
\mu_2,\mu_3,\mu_4,\mu_5,r_{ij}\ge0,  \mu_{1ij}r_{ij}=0, \sum_{i,j}{r_{ij}}-1=0\label{kl+},\\
\frac{(\max(1,\delta)-\lambda) \log N}{n}-\sum_{i,j}{r_{ij}\log p_{ij}}+\sum_{i,j}{r_{ij}\log r_{ij}} \leq0,\\
 \mu_2\left( \frac{(\max(1,\delta)-\lambda) \log N}{n}-\sum_{i,j}{r_{ij}\log p_{ij}}+\sum_{i,j}{r_{ij}\log r_{ij}} \right)=0,\label{kl1}\\
\frac{(1-\lambda) \log N}{n}-\sum_{i,j}{r_{ij}\log p_{ij}}+\sum_{i,j}{r_{ij}\log p_{i}^{\mathcal{A}}}\leq 0\label{tr-cv},\\
\mu_3\left(\frac{(1-\lambda) \log N}{n}-\sum_{i,j}{r_{ij}\log p_{ij}}+\sum_{i,j}{r_{ij}\log p_{i}^{\mathcal{A}}}\right)=0,\label{kl2}\\
\frac{(\delta-\lambda) \log N}{n}-\sum_{i,j}{r_{ij}\log p_{ij}}+\sum_{i,j}{r_{ij}\log p_j^{\mathcal{B}}}\le0,\\
\mu_4\left( \frac{(\delta-\lambda) \log N}{n}-\sum_{i,j}{r_{ij}\log p_{ij}}+\sum_{i,j}{r_{ij}\log p_j^{\mathcal{B}}}\right)=0\label{kl+2},\\
 \frac{(1+\delta-\lambda) \log N}{n}-\sum_{i,j}{r_{ij}\log p_{ij}}+\sum_{i,j}{r_{ij}\log q_{ij}}\le0,\\
\mu_5\left( \frac{(1+\delta-\lambda) \log N}{n}-\sum_{i,j}{r_{ij}\log p_{ij}}+\sum_{i,j}{r_{ij}\log q_{ij}}\right)=0\label{kl+4}.
\end{eqnarray}
From \eqref{kl+}, $\mu_{1ij}$ is zero if $r_{ij}$ is a non-zero number. Therefore, we only keep $i$ and $j$ where $r_{ij}\neq0$ and $\mu_{1ij}=0$.
\begin{eqnarray}
&&\frac{d F(r_{ij},n,\lambda)}{dr_{ij}} = 0 \rightarrow \mu_2 + \mu_2\log r_{ij}+ \mu_3\log p_{i}^{\mathcal{A}}+ \mu_4\log p_j^{\mathcal{B}}+ \mu_5\log q_{ij}+ \mu_6\nonumber\\
&& -(\mu_2+\mu_3+\mu_4+\mu_5)\log p_{ij}= 0 \label{in-1}\\
&&\rightarrow r_{ij} ^ {\mu_2} = p_{ij} ^ {\mu_2+\mu_3+\mu_4+\mu_5} {(p_i^{\mathcal{A}})} ^ {-\mu_3}{( p_j^{\mathcal{B}})} ^ {-\mu_4}q_{ij} ^ {-\mu_5} e ^ {-\mu_2  - \mu_6}.
\end{eqnarray}
Consider the following two cases.
\begin{enumerate}
\item{$\mu_2=0$.} In this case, all the constraints are affine functions and therefore we have a linear programming problem and the feasible set of this linear programming problem is a polyhedron. From \eqref{l0}, the polyhedron is bounded, i.e., $0\le r_{ij}\le M$ for some constant $M$. Assume that the polyhedron is nonempty, otherwise the solution is $\infty$. Moreover, a nonempty bounded polyhedron cannot contain a line, thus it must have a basic feasible solution and the optimal solutions are restricted to the corner points. 
\item{$\mu_2\neq0$.} As $r_{ij}\neq0$ and $\mu_{1ij}=0$, we have
\begin{eqnarray}
&&\frac{d F(r_{ij},n,\lambda)}{dr_{ij}} = 0 \rightarrow \mu_2 + \mu_2\log r_{ij}+ \mu_3\log p_{i}^{\mathcal{A}}+ \mu_4\log p_j^{\mathcal{B}}+ \mu_5\log q_{ij}+ \mu_6\nonumber\\
&& -(\mu_2+\mu_3+\mu_4+\mu_5)\log p_{ij}= 0 \label{...}\\
&&\rightarrow r_{ij} ^ {\mu_2} = p_{ij} ^ {\mu_2+\mu_3+\mu_4+\mu_5} {(p_i^{\mathcal{A}})} ^ {-\mu_3}{(p_j^{\mathcal{B}})} ^ {-\mu_4}q_{ij} ^ {-\mu_5} e ^ {-\mu_2  - \mu_6}
 \label{goodd}\\
&&\rightarrow r_{ij} =cp_{ij} ^ {\frac{\mu_2+\mu_3+\mu_4+\mu_5}{\mu_2}} {(p_i^{\mathcal{A}})} ^ {-\frac{\mu_3}{\mu_2}}  {(p_j^{\mathcal{B}})} ^ {-\frac{\mu_4}{\mu_2}}  q_{ij} ^ {-\frac{\mu_5}{\mu_2}},  \label{good}\\
&&\frac{d F(r_{ij},n,\lambda)}{dn} = 0 \nonumber\\
&&\rightarrow-\left(\mu_2(\max(1,\delta)-\lambda) +\mu_3(1-\lambda) +\mu_4(\delta-\lambda)\right.\nonumber\\
&&\left.+\mu_5(1+\delta-\lambda) \right)\frac{\log N}{n^2}=0\label{j1}\\
&&\rightarrow\mu_2(\max(1,\delta)-\lambda) +\mu_3(1-\lambda) \nonumber\\
&&+\mu_4(\delta-\lambda)+\mu_5(1+\delta-\lambda)=0\label{tr+4}\\
&&\rightarrow\lambda=\frac{\mu_2\max(1,\delta)+\mu_3+\mu_4\delta+\mu_5(1+\delta)}{\mu_2+\mu_3+\mu_4+\mu_5}\label{gdgdd}.
\end{eqnarray}
Summing \eqref{kl1}, \eqref{kl2}, \eqref{kl+2} and \eqref{kl+4} and using \eqref{good}, we have
\begin{eqnarray}
&&\left(\mu_2(\max(1,\delta)-\lambda) +\mu_3(1-\lambda) \right.\left.+\mu_4(\delta-\lambda)+\mu_5(1+\delta-\lambda) \right)\frac{\log N}{n}\nonumber\\
&=&\mu_2\log c.\label{good+.}
\end{eqnarray}
From \eqref{gdgdd} and $\mu_2\neq0$, we conclude that $c=1$. Moreover, we have $\lambda=\frac{\max(1,\delta)+\mu+\nu\delta}{1+\mu+\nu+\eta}$, where $\mu$, $\nu$ and $\eta$ are defined as: 
\begin{eqnarray}
\mu = \frac{\mu_3+\mu_5}{\mu_2},\label{j4}
\\\nu = \frac{\mu_4+\mu_5}{\mu_2},\label{j5}
\\\eta = \frac{\mu_2'-\mu_5}{\mu_2}.\label{j555}
\end{eqnarray}
Assume that $\eta>0$, therefore $r_{ij}<p_{ij}$ as $p_{ij}\le\min(q_i,q_j)$. On the other hand, $\sum_{i,j}{r}_{ij}=\sum p_{ij}=1$ which contradicts our assumption that $\eta>0$. Thus, $\eta\le0$.   Define $(\mu^*,\nu^*,\eta^*)$ as follows
\begin{eqnarray}
(\mu^*,\nu^*,\eta^*)&=&Arg\max_{\min(\mu,\nu)\ge\eta\ge0,\sum_{i,j}p_{ij} ^ {1+\mu+\nu-\eta} {(p_{i}^{\mathcal{A}})} ^ {-\mu}  {(p_j^{\mathcal{B}})} ^ {-\nu}   = 1}\nonumber\\
&&\frac{\max(1,\delta)+\mu+\nu\delta}{1+\mu+\nu-\eta}\label{local1}.
\end{eqnarray}
Therefore, from \eqref{kl1}, \eqref{good} and \eqref{gdgdd} we conclude Lemma \ref{lem_optim} as follows 
\begin{eqnarray}
r_{ij}^* &=&p_{ij} ^ {1+\mu^*+\nu^*-\eta^*} {(p_{i}^{\mathcal{A}})} ^ {-\mu^*}  {(p_j^{\mathcal{B}})} ^ {-\nu^*}, \\
\lambda^*&=&\frac{\max(1,\delta)+\mu^*+\nu^*\delta}{1+\mu^*+\nu^*-\eta^*},\\
n^*&=&\frac{(\max(1,\delta)-\lambda^*) \log N}{\sum r_{ij}^*\log \frac{p_{ij}}{r_{ij}^*}}.\label{local2}
\end{eqnarray}
\end{enumerate}

\subsection{Proof of Lemma \ref{lem_numnodes_}}\label{proof_lem_numnodes_}
In order to prove Lemma \ref{lem_numnodes_}, consider a leaf node $v_0$ and its parent $v_1$. From \eqref{ok}, for the node $v_1$ we have
\begin{eqnarray}
\frac{\Phi(v_1)}{\Psi(v_1)} &\le& {N ^ {1+\delta-\lambda^*}}{p_0q_0}\label{v1_1},\\
\frac{\Phi(v_1)}{{\Psi}^{\mathcal{A}}(v_1)}&\ge &N^{1-\lambda^*}{p_0q_0}\label{v1_2},\\
\frac{\Phi(v_1)}{{\Psi}^{\mathcal{B}}(v_1)}&\ge &N^{\delta-\lambda^*}{p_0q_0}\label{v1_3}.
\end{eqnarray}
$(\eqref{v1_1}-\eqref{v1_3})$ follow from \eqref{ok} and the fact that the node  $v_1$ is niether pruned nor accepted as it is not a leaf node. Therefore, from $\eqref{v1_2}\times\eqref{v1_3}/\eqref{v1_1}$, we conclude that,
\begin{eqnarray}
{\Phi(v_1)}&\ge &{N ^ {-\lambda^*}}{p_0q_0}\label{v1_4}.
\end{eqnarray}
Therefore, from definition of $\Phi(v)$, for the leaf node $v_0$ we have
\begin{eqnarray}
{\Phi(v_0)}&\ge &{N ^ {-\lambda^*}}\min_{i,j} p_{ij}{p_0q_0}.\label{v1_5}
\end{eqnarray}
\eqref{v1_5} is true for all the leaf nodes as $(\eqref{v1_1}-\eqref{v1_5})$ were derived independent of the choice of the leaf node.

\subsection{Proof of Lemma \ref{lemma_sum_av}}\label{app_lemma_sum_av} 
The proof is straightforward based on proof of Lemma \ref{induction_AB}. $\sum_{v\in V_{l}}\Phi(v)=1$ is proved by the induction on depth of the tree. For the tree $G$ with $depth(G)=1$, $\sum_{v\in V_{l}}\Phi(v)=1$ is trivial as for the children $v_{ij}$ of the root we have $\Phi(v_{ij})=p_{ij}$ and $\sum_{ij}p_{ij}=1$. Assume that $\sum_{v\in V_{l}}\Phi(v)=1$ is true for all the trees $G$ with $depth(G)\le depth$. Our goal is to prove that  $\sum_{v\in V_{l}}\Phi(v)=1$ is true for all the trees $G$ with $depth(G)= depth+1$. Consider a tree $G$ with $depth(G)= depth+1$ and the tree $G'$ obtained by removing all the nodes at depth $ depth+1$.
\begin{eqnarray}
&&\sum_{v\in V_{l}(G)}\Phi(v)\nonumber\\
&=&\sum_{v\in V_{l}(G),depth(v)=depth+1}\Phi(v)-\sum_{w\in V(G), w \mbox{ is a parent of a leaf node},depth(w)=depth}\Phi(w)\nonumber\\
&&+\sum_{v\in V_{l}(G')}\Phi(v)\label{tr_e0}\\
&=&\sum_{v\in V_{l}(G')}\Phi(v)\label{tr_e1}\\
&=&1\label{tr_e2}.
\end{eqnarray}
\eqref{tr_e0} is a result of definition of $G'$, i.e., $G'$ is  obtained from $G$ by removing all the nodes at depth $ depth+1$. \eqref{tr_e1} is true as for any node $w$ and its children $v_{ij}$ we have $\Phi(w)=\sum_{ij} \Phi(v_{ij})$ which is a result of the fact that $\Phi(v_{ij})=\Phi(w)p_{ij}$. \eqref{tr_e2} is concluded from the induction assumption, i.e.,  $\sum_{v\in V_{l}}\Phi(v)=1$ is true for all the trees $G$ with $depth(G)\le depth$.

\subsection{Proof of Lemma \ref{lemma_num_nodess}}\label{app_lemma_num_nodess} For any decision tree which at each node either it is pruned, accepted or branched into $kl$ children, number of nodes in the tree is at most two times number of leaf nodes, i.e., $|V(G)|\le 2|V_l(G)|$. This is true by induction on the depth of the tree. For a tree $G$ with $depth(G)=1$, we have $|V_l(G)|=kl$ and $|V(G)|=kl+1$. Therefore, Lemma \ref{lemma_num_nodess} is true in this case. Assume that $|V(G)|\le 2|V_l(G)|$ is true for all the trees $G$ with $depth(G)\le depth$. Our goal is to prove that  $|V(G)|\le 2|V_l(G)|$ is true for all the trees $G$ with $depth(G)= depth+1$. Consider a tree $G$ with  $depth(G)=depth+1$. Consider the tree $G'$ obtained by removing all the nodes $v$ where $depth(v)=depth+1$. Assume there are $klr$ of them (each intermediate node has $kl$ children). Therefore, we have $|V(G)|=|V(G')|+klr$, $|V_l(G)|=|V_l(G')|+(kl-1)r$ and $|V(G')|\le 2|V_l(G')|$. This results in  $|V(G)|\le 2|V_l(G)|$.

\section{Proof of Theorem \ref{theorem_noise}}\label{rigor_proof_C}
First of all, note that from $\frac{p_{ij}}{1+\epsilon}\le p'_{ij}\le{p_{ij}}{(1+\epsilon)}$, we conclude that  $\frac{p_{i}^{\mathcal{A}}}{1+\epsilon}\le {p'}_{i}^{\mathcal{A}}\le{{p}_{i}^{\mathcal{A}}}{(1+\epsilon)}$, $\frac{p_j^{\mathcal{B}}}{1+\epsilon}\le {p'}_{j}^{\mathcal{B}}\le{p_j^{\mathcal{B}}}{(1+\epsilon)}$ and $\frac{q_{ij}}{{{(1+\epsilon)}^2}}\le q'_{ij}\le{q_{ij}}{{{(1+\epsilon)}^2}}$. Assume that $depth(G)=d$. Define the variables $\Phi(v)$, ${\Psi}^{\mathcal{A}}(v)$, ${\Psi}^{\mathcal{B}}(v)$, $\Psi(v)$, $\alpha(G)$, $\gamma^{\mathcal{A}}(G)$, $\gamma^{\mathcal{B}}(G)$, $\beta(G)$ and $TP$ for the tree with the distribution $p'$ as  $\Phi'(v)$, $\Psi'^{\mathcal{A}}(v)$, $\Psi'^{\mathcal{B}}(v)$, $\Psi'(v)$, $\alpha'(G)$, $\gamma'^{\mathcal{A}}(G)$, $\gamma'^{\mathcal{B}}(G)$, $\beta'(G)$ and $TP'$. Therefore, from $\eqref{alphaav}-\eqref{gammydef}$ we have
\begin{eqnarray}
\alpha(G)&=&\sum_{v\in V_{Buckets}(G)}\Phi(v)\label{alphdefeee}\\
&\le&\sum_{v\in V_{Buckets}(G)}\Phi'(v) (1+\epsilon)^d\label{ter_e}\\
&\le&\alpha'(G) (1+\epsilon)^d,
\end{eqnarray}
\eqref{ter_e} follows from $\Phi(v)\le \Phi'(v)(1+\epsilon)^d$ which is a result of the definition of $\Phi(v)$ in \eqref{def_a2}, i.e., $\Phi(f(v,a_i,b_j))=\Phi(v)p_{ij},\forall v\in V$. Similarly, we have
\begin{eqnarray}
\frac{\alpha(G)}{(1+\epsilon)^d}\le&\alpha'(G)&\le \alpha(G) (1+\epsilon)^d\label{ty_yt1},\\
\frac{\gamma^{\mathcal{A}}(G)}{(1+\epsilon)^d}\le&\gamma'^{\mathcal{A}}(G)&\le \gamma^{\mathcal{A}}(G) (1+\epsilon)^d,\\
\frac{\gamma^{\mathcal{B}}(G)}{(1+\epsilon)^d}\le&\gamma'^{\mathcal{B}}(G)&\le \gamma^{\mathcal{B}}(G) (1+\epsilon)^d,\\
\frac{\beta(G)}{(1+\epsilon)^{2d}}\le&\beta'(G)&\le \beta(G) (1+\epsilon)^{2d}\label{ty_yt2}.
\end{eqnarray}
On the other hand, from $\eqref{chagh}-\eqref{approx1}$, we have 
\begin{eqnarray}
TP&=&1-{(1-\alpha'(G))}^{\#bands}\ge 1-{(1-\frac{\alpha(G)}{(1+\epsilon)^d})}^{\#bands},
\end{eqnarray}
similar to \eqref{ttt-1m}, and the inequality $(1-x)^{\frac{c}{x}}< e^{-c}$, the minimum possible value of $\#bands$ to ensure true positive rate $TP$ can be computed as 
\begin{eqnarray}
\#bands&=&\lceil\frac{\log \frac{1}{1-TP}}{\frac{\alpha(G)}{(1+\epsilon)^d}}\rceil\label{approx1s}.
\end{eqnarray}
Thus, the  total complexity is computed as
\begin{eqnarray}
&&c_{tree}|V(G)|+\left(\frac{c_{hash}N}{\alpha(G)}+\frac{c_{hash}M}{\alpha(G)}+\frac{c_{insertion}N\gamma'^{\mathcal{A}}(G)}{\alpha(G)}+\frac{c_{insertion}M\gamma'^{\mathcal{B}}(G)}{\alpha(G)}\right.\nonumber\\
&&\left.+\frac{c_{pos}MN\beta'(G)}{\alpha(G)}\right).{{(1+\epsilon)^d}\log \frac{1}{1-TP}}\\
&\le&{(1+\epsilon)}^{3d} N^{\lambda^{*}}\label{ty_yt}\\
&\le&N^{\lambda^{*}+3c_d\log(1+\epsilon)}, \label{ty_yt3}
\end{eqnarray}
where \eqref{ty_yt} follows from $\eqref{ty_yt1}-\eqref{ty_yt2}$ and the fact that the total complexity for distribution $p'$ is $O(N^{\lambda^{*}(p')})$. Finally, \eqref{ty_yt3} is obtained from the following lemma in which we prove that the depth of the tree is bounded by $c_d\log N$ where $c_d$ is a constant depending on the distribution.
\begin{lemma}\label{lem_depthh} For the decision tree $G$, $depth(G)=c_d\log N$ where $c_d=\min\big(\frac{{(\lambda^*-1)}}{\log(\max_{i,j}\frac{p_{ij}}{p_i^{\mathcal{A}}})},\frac{{(\lambda^*-\delta)}}{\log(\max_{i,j}\frac{p_{ij}}{p_j^{\mathcal{B}}})}\big)$.
\end{lemma}
For proof of Lemma \ref {lem_depthh}, see Appendix \ref{proof_lem_depthh}.
\subsection{Proof of Lemma \ref{lem_depthh}}\label{proof_lem_depthh}
From \eqref{ok}, for the decision tree $G$ we have
\begin{eqnarray}
\left\{\begin{matrix}\frac{\Phi(v)}{\Psi(v)} \ge {N ^ {1+\delta-\lambda^*}}{p_0q_0}&: \mbox{Accept bucket},~~~~~~~~~~~~~~~~\\
\frac{\Phi(v)}{{\Psi}^{\mathcal{A}}(v)}\le N^{1-\lambda^*}{p_0q_0}&: \mbox{{}Prune},~~~~~~~~~~~~~~~~~~~~~~~~~~~\\
\frac{\Phi(v)}{{\Psi}^{\mathcal{B}}(v)}\le N^{\delta-\lambda^*}{p_0q_0}&: \mbox{{}Prune},~~~~~~~~~~~~~~~~~~~~~~~~~~~\\
otherwise&: \mbox{Branch into the $kl$ children.} \end{matrix}\right.\label{okk}
\end{eqnarray}
Our goal here is to prove that for any pruned node $v$ and any accepted node $v$, $depth(v)\le c_d\log N$ where $c_d$ is a constant depending on the distribution. Consider a pruned or accepted node $v$. For its parent $w$, we have
\begin{eqnarray}
N^{1-\lambda^*}{p_0q_0}
&<&\frac{\Phi(v)}{{\Psi}^{\mathcal{A}}(v)}.
\end{eqnarray}
Therefore, we conclude that
\begin{eqnarray}
d&\le& \frac{{(\lambda^*-1)}\log N}{\log(\max_{i,j}\frac{p_{ij}}{p_i^{\mathcal{A}}})}\label{g_q}.
\end{eqnarray}
Similarly, we have
\begin{eqnarray}
N^{\delta-\lambda^*}{p_0q_0}
&<&\frac{\Phi(v)}{{\Psi}^{\mathcal{B}}(v)}\\
\rightarrow d&\le& \frac{{(\lambda^*-\delta)}\log N}{\log(\max_{i,j}\frac{p_{ij}}{p_j^{\mathcal{B}}})}\label{g_q22}.
\end{eqnarray}
Thus, $c_d$ is defined as $ \min\big(\frac{{(\lambda^*-1)}}{\log(\max_{i,j}\frac{p_{ij}}{p_i^{\mathcal{A}}})},\frac{{(\lambda^*-\delta)}}{\log(\max_{i,j}\frac{p_{ij}}{p_j^{\mathcal{B}}})}\big)$. \eqref{g_q} and \eqref{g_q22} are true as $p_0,q_0\le1$.

\section{{Pseudo Code }}\label{pseudo_code}
{\color{black}Here, we present the pseudo code for the algorithm for \cite{Moshe_Heterogeneous} for the case of hamming distance (see Experiment \ref{exp_dubiii}).

\begin{algorithm}[H]
   \caption{Algorithm for \cite{Moshe_Heterogeneous} for the case of hamming distance}\label{d_dub_d}
\begin{algorithmic}
  {\color{black} \STATE {\bfseries Inputs:} Probability $p$, number of data points $N$, and dimension $S$.
\STATE {\bfseries Output:} {{}Complexity of dubiner algorithm.}
\STATE {\bf For} $d\in\{1,2,\cdots,S\}$
\STATE~~~~~~~~~~~ Generate data points $x\in{\{0,1\}}^S$ and buckets $b\in {\{0,1\}}^S$ from Bernoulli($\frac{1}{2}$)
\STATE~~~~~~~~~~~ $P_1(d) \leftarrow$ ratio of data points $x$ falling within distance $d$ of $b$.
\STATE~~~~~~~~~~~ Generate data points $x,y\in{\{0,1\}}^S$ by $\begin{bmatrix}0.5-p&p\\
p&0.5-p\end{bmatrix}$ and buckets $b\in {\{0,1\}}^S$ from Bernoulli($\frac{1}{2}$)
\STATE~~~~~~~~~~~ $P_2(d) \leftarrow$ ratio of pairs of data points $(x,y)$ both falling within distance $d$ of $b$.
\STATE {\bf End For}
\STATE Select $d_0$ such that $P_1(d_0)\approx \frac{1}{N}$.
\STATE {\bf Return} $\frac{\log(P_2(d_0))}{\log(P_1(d_0))}$.}
\end{algorithmic}
\end{algorithm}}

%
%
%
%
%

\section{Further Discussion on MIPS}\label{mips_app}

In order to use MIPS to solve this problem, i.e., \eqref{pyx_2}, we need to derive optimal weights $\omega_{ij}$ to minimize the norm $M^2$ in \cite{shrivastava2014asymmetric}. The term $M$ stands for the radius of the space which is computed as follows: $M^2=\mathbb{E}{\big(||x||\big)}^2+\mathbb{E}{\big(||y||\big)}^2$. Therefore,  from \eqref{g_f1}-\eqref{g_f2} we conclude that $M^2=\sum_{ij}\left( p_j^{\mathcal{B}}\omega_{ij}^2+\frac{p_i^{\mathcal{A}}{\log^2(\frac{p_{ij}}{q_{ij}})}}{\omega_{ij}^2}\right)$ which results in optimal $\omega_{ij}={\big(\frac{p_i^{\mathcal{A}}}{p_j^{\mathcal{B}}}\big)}^{0.25}{\big(\mid\log\frac{p_{ij}}{q_{ij}}\mid\big)}^{0.5}$. On the other hand, for $(x,y)\sim Q(x,y)$ we have 
\begin{eqnarray}
\mathbb{E}\big(||x||||y||\big)\ge \mathbb{E}(<T(x),T(y)>) =S\sum_{ij}q_{ij}{\mid\log(\frac{p_{ij}}{q_{ij}})\mid}.
\end{eqnarray}
In order to have nearly one true positive rate and sub-quadratic complexity we need $S_0\le S d_{KL}(p_{ij}||q_{ij})$ and $cS_0\ge -S d_{KL}(q_{ij}||p_{ij})$ where  $d_{KL}$ stands for kullback leibler divergence.  Moreover, we should have $M^2\ge S\sum_{ij}\sqrt{q_{ij}}|\log(\frac{p_{ij}}{q_{ij}})|$. Setting $c=0$, $S_0$ and $M$ as above, the complexity will be more than $1.9$ for any $2\times2$ probability distribution matrix. The reason is that the transferred data points are nearly orthogonal to each other and this makes it very slow to find maximum inner product using the existing method \cite{shrivastava2014asymmetric}.   

\section{Complexities of MinHash, LSH-hamming and ForestDSH}\label{LSH_MinHash}
In this section, we derive the complexities of MinHash, LSH-hamming and ForestDSH for   ${P}_1=\begin{bmatrix}
0.345 & 0  \\ 
0.31 & 0.345 
\end{bmatrix}$ for instance. Complexities are computed for the other probability distributions similarly. 
\subsection{ Complexity of MinHash}
For MinHash the query complexity is 
\begin{eqnarray}
N^{\min(mh_1,mh_2,mh_3,mh_4)}\label{Eq_y},
\end{eqnarray}
where $mh_1={\frac{\log\frac{p_{00}}{1-p_{11}}}{\log\frac{q_{00}}{1-q_{11}}}}$, $mh_2={\frac{\log\frac{p_{01}}{1-p_{10}}}{\log\frac{q_{01}}{1-q_{10}}}}$, $mh_3={\frac{\log\frac{p_{10}}{1-p_{01}}}{\log\frac{q_{10}}{1-q_{01}}}}$ and $mh_4={\frac{\log\frac{p_{11}}{1-p_{00}}}{\log\frac{q_{11}}{1-q_{00}}}}$. Similarly for   ${P}_1$, the per query complexity  is derived and is equal to $0.5207$.
\subsection{ Complexity of  LSH-hamming}
In the case of LSH-hamming, the query complexity  is
\begin{eqnarray}
O(N^{\min(\frac{\log(p_{00}+p_{11})}{\log(q_{00}+q_{11})},\frac{\log(p_{01}+p_{10})}{\log(q_{01}+q_{10})})})\label{Eq_x},
\end{eqnarray}
 and the storage required for the algorithm is $O(N^{1+\min(\frac{\log(p_{00}+p_{11})}{\log(q_{00}+q_{11})},\frac{\log(p_{01}+p_{10})}{\log(q_{01}+q_{10})})})$. Similarly for   ${P}_1$, the per query complexity  is derived and is equal to $0.4672$. 
\subsection{ Complexity of  ForestDSH}
From Definition \ref{deflambda}, we derive $\lambda^*$  as follows
\begin{eqnarray}
(\mu^*,\nu^*,\eta^*)&=&Arg\max_{\min(\mu,\nu)\ge\eta>0,\sum_{i,j}p_{ij} ^ {1+\mu+\nu-\eta} {(p_{i}^{\mathcal{A}})} ^ {-\mu}  {(p_j^{\mathcal{B}})} ^ {-\nu}  = 1}\frac{1+\mu+\nu}{1+\mu+\nu-\eta}\\
&=&( 4.6611,4.6611, 3.1462)\\
\lambda^*&=&\frac{1+\mu^*+\nu^*}{1+\mu^*+\nu^*-\eta^*}\\
&=&1.4384.
\end{eqnarray}
Note that $\delta=1$ and the  per query complexity  is equal to $0.4384$.

\section{Joint Probability Distributions Learned on Mass Spectrometry Data}\label{app_mass}
The mass spectrometry data for experiment \ref{exp_mass}, is shown in Figures \ref{mass_5151} $(a)$, $(b)$ and $(c)$ in case of $\log Rank$ at base $4$ (a $4\times4$ matrix), $\log Rank$ at base $2$ (an $8\times8$ matrix), and no $\log Rank$ transformation (a $51\times51$ matrix). For the mass spectrometry data shown in Figure \ref{mass_5151} $(a)$, the probability distribution $P_{4\times4}$ is given in \eqref{mxm1}. Note that, in the case of  LSH-hamming the query complexity for these $4\times4$, $8\times8$ and $51\times51$  matrices are $0.901$, $0.890$ and $0.905$, respectively. Similarly,  per query complexity for  MinHash for these $4\times4$, $8\times8$ and $51\times51$  matrices are $0.4425$, $0.376$ and $0.386$, respectively. 

For the mass spectrometry data shown in Figure \ref{mass_5151} $a$ and $(b)$, the probability distribution $p(x,y)$ is represented as
\begin{eqnarray}
P_{4\times4}&=&\begin{bmatrix}
0.000125&  5.008081*{10}^{-5}&  9.689274* {10}^{-8}&  0.000404\\
5.008082* {10}^{-5}&  0.000209&  6.205379* {10}^{-6}&  0.001921\\
 9.689274* {10}^{-8}&  6.205379* {10}^{-6}&  2.688879* {10}^{-5}&  0.000355\\
 0.000404&  0.001921&  0.000355&  0.994165
\end{bmatrix}\label{mxm1},\\
P_{8\times8}&=&\left[\begin{matrix}
3.458* {10}^{-5}&
  1.442* {10}^{-5}&
  5.434* {10}^{-6}&
  1.723* {10}^{-6}\\
1.442* {10}^{-5}&
  3.708* {10}^{-5}&
  2.550* {10}^{-5}&
  8.706* {10}^{-6}\\
5.434* {10}^{-6}&
  2.550* {10}^{-5}&
  3.907* {10}^{-5}&
  2.948* {10}^{-5}\\
 1.723* {10}^{-6}&
  8.706* {10}^{-6}&
  2.948* {10}^{-5}&
  4.867* {10}^{-5}\\
2.921* {10}^{-7}&
  1.561* {10}^{-6}&
  6.442* {10}^{-6}&
  1.813* {10}^{-5}&
\\
7.496* {10}^{-8}&
  4.809* {10}^{-7}&
  2.008* {10}^{-6}&
  6.098* {10}^{-6}\\
6.718* {10}^{-8}&
  2.680* {10}^{-7}&
  1.251* {10}^{-6}&
  4.531* {10}^{-6}\\
5.023* {10}^{-5}&
  1.574* {10}^{-4}&
  3.671* {10}^{-4}&
  5.539* {10}^{-4}
\end{matrix}\right.\\
&&\left.\begin{matrix}&
  2.920* {10}^{-7}&
  7.496* {10}^{-8}&
  6.718* {10}^{-8}&
  5.023* {10}^{-5}\\&
  1.561* {10}^{-6}&
  4.809* {10}^{-7}&
  2.680* {10}^{-7}&
  1.575* {10}^{-4}\\&
  6.442* {10}^{-6}&
  2.008* {10}^{-6}&
  1.251* {10}^{-6}&
  3.672* {10}^{-4}\\
 &
  1.813* {10}^{-5}&
  6.098* {10}^{-6}&
  4.532* {10}^{-6}&
  5.539* {10}^{-4}\\&
  2.887* {10}^{-5}&
  6.892* {10}^{-6}&
  5.309* {10}^{-6}&
  4.138* {10}^{-4}\\&
  6.892* {10}^{-6}&
  2.123* {10}^{-5}&
  5.826* {10}^{-6}&
  3.246* {10}^{-4}\\&
  5.309* {10}^{-6}&
  5.826* {10}^{-6}&
  6.411* {10}^{-5}&
  8.364* {10}^{-4}\\&
  4.138* {10}^{-4}&
  3.246* {10}^{-4}&
  8.364* {10}^{-4}&
  0.994
\end{matrix}\right].
\end{eqnarray}
From \eqref{in0}, for $P_{4\times4}$,  $(\mu^*,\nu^*,\eta^*,\lambda^*)$ are derived as
\begin{eqnarray}
\mu^*&=& 1.151016\label{mxm2},\\
\nu^*&=& 1.151016,\\
\eta^*&=& 0.813168,\\
\lambda^*&=& 1.326723.\label{mxm3}
\end{eqnarray}
Similarly, for $P_{8\times8}$, we have
\begin{eqnarray}
\mu^*&=& 0.871147,\\
\nu^*&=& 0.871147,\\
\eta^*&=& 0.624426,\\
\lambda^*&=& 1.294837.
\end{eqnarray}
For the mass spectrometry data shown in Figure \ref{mass_5151} $(c)$, $(\mu^*,\nu^*,\eta^*,\lambda^*)$ are derived as
\begin{eqnarray}
\mu^*&=& 0.901208,\\
\nu^*&=& 0.901208,\\
\eta^*&=& 0.615797,\\
\lambda^*&=& 1.281621.
\end{eqnarray}


\begin{thebibliography}{10}
\providecommand{\url}[1]{#1}
\csname url@samestyle\endcsname
\providecommand{\newblock}{\relax}
\providecommand{\bibinfo}[2]{#2}
\providecommand{\BIBentrySTDinterwordspacing}{\spaceskip=0pt\relax}
\providecommand{\BIBentryALTinterwordstretchfactor}{4}
\providecommand{\BIBentryALTinterwordspacing}{\spaceskip=\fontdimen2\font plus
\BIBentryALTinterwordstretchfactor\fontdimen3\font minus
  \fontdimen4\font\relax}
\providecommand{\BIBforeignlanguage}[2]{{%
\expandafter\ifx\csname l@#1\endcsname\relax
\typeout{** WARNING: IEEEtran.bst: No hyphenation pattern has been}%
\typeout{** loaded for the language `#1'. Using the pattern for}%
\typeout{** the default language instead.}%
\else
\language=\csname l@#1\endcsname
\fi
#2}}
\providecommand{\BIBdecl}{\relax}
\BIBdecl

\bibitem{frank2011spectral}
A.~M. Frank, M.~E. Monroe, A.~R. Shah, J.~J. Carver, N.~Bandeira, R.~J. Moore,
  G.~A. Anderson, R.~D. Smith, and P.~A. Pevzner, ``Spectral archives:
  extending spectral libraries to analyze both identified and unidentified
  spectra,'' \emph{Nature methods}, vol.~8, no.~7, p. 587, 2011.

\bibitem{Aebersold}
R.~Aebersold and M.~Mann, ``Mass spectrometry-based proteomics,''
  \emph{Nature}, vol. 422, no. 6928, p. 198, 2003.

\bibitem{MSGF}
S.~Kim and P.~A. Pevzner, ``MS-GF+ makes progress towards a universal database
  search tool for proteomics,'' \emph{Nature communications}, vol.~5, p. 5277,
  2014.

\bibitem{dasarathy1977visiting}
B.~V. Dasarathy and B.~V. Sheela, ``Visiting nearest neighbors-a survery of
  nearest neighbor pattern classification techniques,'' in \emph{Proceedings of
  the international conference on cybernetics and society}, pp. 630--636, 1977.

\bibitem{yianilos1993data}
P.~N. Yianilos, ``Data structures and algorithms for nearest neighbor search in
  general metric spaces,'' in \emph{Soda}, vol.~93, no. 194, pp. 311--21, 1993.

\bibitem{duda1973pattern}
R.~O. Duda, P.~E. Hart \emph{et~al.}, \emph{Pattern classification and scene
  analysis}.\hskip 1em plus 0.5em minus 0.4em\relax Wiley New York,
  vol.~3, 1973.

\bibitem{guttman1984r}
A.~Guttman, ``R-trees: A dynamic index structure for spatial searching,'' in
  \emph{Proceedings of the 1984 ACM SIGMOD international conference on
  Management of data}, pp. 47--57, 1984.

\bibitem{mori2001shape}
G.~Mori, S.~Belongie, and J.~Malik, ``Shape contexts enable efficient retrieval
  of similar shapes,'' in \emph{Proceedings of the 2001 IEEE Computer Society
  Conference on Computer Vision and Pattern Recognition. CVPR 2001},
  vol.~1, pp. I--I,  2001.



\bibitem{shakhnarovich2003fast}
G.~Shakhnarovich, P.~Viola, and T.~Darrell, ``Fast pose estimation with
  parameter-sensitive hashing,'' in IEEE International Conference on Computer Vision-Volume 2, p. 750, 2003.



\bibitem{Bentley}
J.~L. Bentley, ``Multidimensional binary search trees used for associative
  searching,'' \emph{Communications of the ACM}, vol.~18, no.~9, pp. 509--517,
  1975.

\bibitem{prab}
Y.~Prabhu and M.~Varma, ``Fastxml: A fast, accurate and stable tree-classifier
  for extreme multi-label learning,'' in \emph{Proceedings of the 20th ACM
  SIGKDD international conference on Knowledge discovery and data mining}, pp. 263--272, 2014.

\bibitem{Finkel}
J.~Friedman, J.~Bentley, and R.~Finkel, ``An algorithm for finding best matches
  in logarithmic time,'' \emph{ACM Trans. Math. Software, 3(SLAC-PUB-1549-REV.
  2}, pp. 209--226, 1976.

\bibitem{jain2016extreme}
H.~Jain, Y.~Prabhu, and M.~Varma, ``Extreme multi-label loss functions for
  recommendation, tagging, ranking \& other missing label applications,'' in
  \emph{Proceedings of the 22nd ACM SIGKDD International Conference on
  Knowledge Discovery and Data Mining}.\hskip 1em plus 0.5em minus 0.4em\relax
  ACM, pp. 935--944, 2016.

\bibitem{bhatia2015sparse}
K.~Bhatia, H.~Jain, P.~Kar, M.~Varma, and P.~Jain, ``Sparse local embeddings
  for extreme multi-label classification,'' in \emph{Advances in neural
  information processing systems}, pp. 730--738, 2015.

\bibitem{yen2016pd}
I.~E.-H. Yen, X.~Huang, P.~Ravikumar, K.~Zhong, and I.~Dhillon, ``Pd-sparse: A
  primal and dual sparse approach to extreme multiclass and multilabel
  classification,'' in \emph{International Conference on Machine Learning}, pp. 3069--3077, 2016.

\bibitem{choromanska2015logarithmic}
A.~E. Choromanska and J.~Langford, ``Logarithmic time online multiclass
  prediction,'' in \emph{Advances in Neural Information Processing Systems}, pp. 55--63, 2015.

\bibitem{liu2017making}
W.~Liu and I.~W. Tsang, ``Making decision trees feasible in ultrahigh feature
  and label dimensions,'' \emph{The Journal of Machine Learning Research},
  vol.~18, no.~1, pp. 2814--2849, 2017.

\bibitem{nam2017maximizing}
J.~Nam, E.~L. Menc{\'\i}a, H.~J. Kim, and J.~F{\"u}rnkranz, ``Maximizing subset
  accuracy with recurrent neural networks in multi-label classification,'' in
  \emph{Advances in neural information processing systems}, pp.
  5413--5423, 2017.

\bibitem{rai2015large}
P.~Rai, C.~Hu, R.~Henao, and L.~Carin, ``Large-scale bayesian multi-label
  learning via topic-based label embeddings,'' in \emph{Advances in Neural
  Information Processing Systems}, pp. 3222--3230, 2015.

\bibitem{tagami2017annexml}
Y.~Tagami, ``Annexml: Approximate nearest neighbor search for extreme
  multi-label classification,'' in \emph{Proceedings of the 23rd ACM SIGKDD
  international conference on knowledge discovery and data mining}.\hskip 1em
  plus 0.5em minus 0.4em\relax ACM, pp. 455--464, 2017.

\bibitem{niculescu2017label}
A.~Niculescu-Mizil and E.~Abbasnejad, ``Label filters for large scale
  multilabel classification,'' in \emph{Artificial Intelligence and
  Statistics}, pp. 1448--1457, 2017.

\bibitem{zhou2017binary}
W.-J. Zhou, Y.~Yu, and M.-L. Zhang, ``Binary linear compression for multi-label
  classification.'' in \emph{IJCAI}, pp. 3546--3552, 2017.

\bibitem{beyer1999nearest}
K.~Beyer, J.~Goldstein, R.~Ramakrishnan, and U.~Shaft, ``When is “nearest
  neighbor” meaningful?'' in \emph{International conference on database
  theory}.\hskip 1em plus 0.5em minus 0.4em\relax Springer, pp. 217--235, 1999.

\bibitem{castelli2000searching}
V.~Castelli, C.-S. Li, and A.~Thomasian, ``Searching multidimensional indexes
  using associated clustering and dimension reduction information,'' Oct.~17, {U}{S} Patent 6,134,541, 2000.

\bibitem{anagnostopoulos2015low}
E.~Anagnostopoulos, I.~Z. Emiris, and I.~Psarros, ``Low-quality dimension
  reduction and high-dimensional approximate nearest neighbor,'' in \emph{31st
  International Symposium on Computational Geometry (SoCG 2015)}.\hskip 1em
  plus 0.5em minus 0.4em\relax Schloss Dagstuhl-Leibniz-Zentrum fuer
  Informatik, 2015.

\bibitem{min2005non}
R.~Min, \emph{A non-linear dimensionality reduction method for improving
  nearest neighbour classification}.\hskip 1em plus 0.5em minus 0.4em\relax
  University of Toronto, 2005.

\bibitem{shaw2009structure}
B.~Shaw and T.~Jebara, ``Structure preserving embedding,'' in \emph{Proceedings
  of the 26th Annual International Conference on Machine Learning}, pp. 937--944, 2009.

\bibitem{christiani2017set}
T.~Christiani and R.~Pagh, ``Set similarity search beyond minhash,'' in
  \emph{Proceedings of the 49th Annual ACM SIGACT Symposium on Theory of
  Computing}.\hskip 1em plus 0.5em minus 0.4em\relax ACM, pp. 1094--1107, 2017.

\bibitem{charikar2002similarity}
M.~S. Charikar, ``Similarity estimation techniques from rounding algorithms,''
  in \emph{Proceedings of the thirty-fourth annual ACM symposium on Theory of
  computing}.\hskip 1em plus 0.5em minus 0.4em\relax ACM, pp. 380--388, 2002.

\bibitem{Motwani}
P.~Indyk and R.~Motwani, ``Approximate nearest neighbors: towards removing the
  curse of dimensionality,'' in \emph{Proceedings of the thirtieth annual ACM
  symposium on Theory of computing}, pp. 604--613, 1998.

\bibitem{Indyk_hd}
A.~Gionis, P.~Indyk, R.~Motwani \emph{et~al.}, ``Similarity search in high
  dimensions via hashing,'' in \emph{Vldb}, vol.~99, no.~6, pp. 518--529, 1999.

\bibitem{bawa2005lsh}
M.~Bawa, T.~Condie, and P.~Ganesan, ``{LSH} forest: self-tuning indexes for
  similarity search,'' in \emph{Proceedings of the 14th international
  conference on World Wide Web}, pp. 651--660, 2005.

\bibitem{andoni2017optimal}
A.~Andoni, T.~Laarhoven, I.~Razenshteyn, and E.~Waingarten, ``Optimal
  hashing-based time-space trade-offs for approximate near neighbors,'' in
  \emph{Proceedings of the Twenty-Eighth Annual ACM-SIAM Symposium on Discrete
  Algorithms}.\hskip 1em plus 0.5em minus 0.4em\relax Society for Industrial
  and Applied Mathematics, pp. 47--66, 2017.

\bibitem{rubinstein2018hardness}
A.~Rubinstein, ``Hardness of approximate nearest neighbor search,'' in
  \emph{Proceedings of the 50th Annual ACM SIGACT Symposium on Theory of
  Computing}.\hskip 1em plus 0.5em minus 0.4em\relax ACM, pp. 1260--1268, 2018.

\bibitem{shrivastava2014asymmetric}
A.~Shrivastava and P.~Li, ``Asymmetric lsh (alsh) for sublinear time maximum
  inner product search (mips),'' in \emph{Advances in Neural Information
  Processing Systems}, pp. 2321--2329, 2014.

\bibitem{ACM_andoni}
A.~Andoni and I.~Razenshteyn, ``Optimal data-dependent hashing for approximate
  near neighbors,'' in \emph{Proceedings of the forty-seventh annual ACM
  symposium on Theory of computing}.\hskip 1em plus 0.5em minus 0.4em\relax
  ACM, pp. 793--801, 2015.

\bibitem{nips_andoni}
A.~Andoni, P.~Indyk, T.~Laarhoven, I.~Razenshteyn, and L.~Schmidt, ``Practical
  and optimal lsh for angular distance,'' in \emph{Advances in Neural
  Information Processing Systems}, pp. 1225--1233, 2015.

\bibitem{Chakrabarti}
A.~Chakrabarti and O.~Regev, ``An optimal randomized cell probe lower bound for
  approximate nearest neighbor searching,'' \emph{SIAM Journal on Computing},
  vol.~39, no.~5, pp. 1919--1940, 2010.

\bibitem{Miltersen}
P.~B. Miltersen, ``Cell probe complexity-a survey,'' in \emph{Proceedings of
  the 19th Conference on the Foundations of Software Technology and Theoretical
  Computer Science, Advances in Data Structures Workshop}, p.~2, 1999.

\bibitem{noar}
A.~Andoni, A.~Naor, A.~Nikolov, I.~Razenshteyn, and E.~Waingarten,
  ``Data-dependent hashing via nonlinear spectral gaps,'' in \emph{Proceedings
  of the 50th Annual ACM SIGACT Symposium on Theory of Computing}.\hskip 1em
  plus 0.5em minus 0.4em\relax ACM, pp. 787--800, 2018.

\bibitem{Moshe_Heterogeneous}
M.~Dubiner, ``A heterogeneous high-dimensional approximate nearest neighbor
  algorithm,'' \emph{IEEE Transactions on Information Theory}, vol.~58, no.~10,
  pp. 6646--6658, 2012.

\bibitem{Moshe_Bucketing}
------, ``Bucketing coding and information theory for the statistical
  high-dimensional nearest-neighbor problem,'' \emph{IEEE Transactions on
  Information Theory}, vol.~56, no.~8, pp. 4166 -- 4179, Aug 2010.

\bibitem{mcdonald2018american}
D.~McDonald, E.~Hyde, J.~W. Debelius, J.~T. Morton, A.~Gonzalez, G.~Ackermann,
  A.~A. Aksenov, B.~Behsaz, C.~Brennan, Y.~Chen \emph{et~al.}, ``American gut:
  an open platform for citizen science microbiome research,'' \emph{MSystems},
  vol.~3, no.~3, pp. e00\,031--18, 2018.

\bibitem{krizhevsky2009learning}
A.~Krizhevsky, G.~Hinton \emph{et~al.}, ``Learning multiple layers of features
  from tiny images,'' 2009.

\end{thebibliography}
\end{document}